\documentclass[journal,comsoc]{IEEEtran}

\usepackage[T1]{fontenc}% optional T1 font encoding

\ifCLASSINFOpdf

\else

\fi

\usepackage{amsmath}

\usepackage[cmintegrals]{newtxmath}

\usepackage{amssymb,amsthm}
\usepackage{tabularx}
\usepackage{multirow}
\usepackage{booktabs}
\usepackage{graphicx}
\usepackage{algorithm}
\usepackage{algorithmic}
\usepackage{color}

\newtheorem{theorem}{Theorem}
\newtheorem{lemma}{Lemma}
\newtheorem{corollary}{Corollary}
\newtheorem{remark}{Remark}
\newtheorem{proof of theorem}{Proof of Theorem}
\usepackage{threeparttable}
\usepackage{subfigure}
\usepackage{caption}
\usepackage{color}

\begin{document}
\title{An Improved Frequent Directions Algorithm for Low-Rank Approximation via Block Krylov Iteration
}

\author{Chenhao Wang, Qianxin Yi, Xiuwu Liao, Yao Wang
\thanks{Chenhao Wang, Qianxin Yi, Xiuwu Liao and Yao Wang are with the Center for Intelligent Decision-making and Machine Learning, Xi'an Jiaotong University, Xi'an, 710049, China.}% <-this % stops a space
%\thanks{Qianxin Yi is with the School of Mathematics and Statistics, Xi'an Jiaotong University, Xi'an, 710049, China.}% <-this % stops a space
\thanks{Chenhao Wang and Qianxin Yi contribute equally to this work. Yao Wang is the corresponding author. Email: yao.s.wang@gmail.com.}}


\makeatletter 
\renewcommand{\@thesubfigure}{\hskip\subfiglabelskip}
\makeatother

%\markboth{IEEE TRANSACTIONS ON NEURAL NETWORKS AND LEARNING SYSTEMS}%
%{Shell \MakeLowercase{\textit{et al.}}: Bare Demo of IEEEtran.cls for IEEE Communications Society Journals}



\maketitle

\begin{abstract}
	Frequent Directions, as a deterministic matrix sketching technique, has been proposed for tackling low-rank approximation problems. This method has a high degree of accuracy and practicality, but experiences a lot of computational cost for large-scale data.  Several recent works on the randomized version of Frequent Directions greatly improve the computational efficiency, but unfortunately sacrifice some precision. To remedy such issue, this paper aims to find a more accurate projection subspace to further improve the efficiency and effectiveness of the existing Frequent Directions techniques. Specifically, by utilizing the power of Block Krylov Iteration and random projection technique, this paper presents a fast and accurate Frequent Directions algorithm named as r-BKIFD. The rigorous theoretical analysis shows that the proposed r-BKIFD has a comparable error bound with original Frequent Directions, and the approximation error can be arbitrarily small when the number of iterations is chosen appropriately. Extensive experimental results on both synthetic and real data further demonstrate the superiority of r-BKIFD over several popular Frequent Directions algorithms, both in terms of computational efficiency and accuracy. 
\end{abstract}

\begin{IEEEkeywords}
	Low-rank approximation, randomized sketching, Frequent Directions, Block Krylov Iteration, streaming algorithms.
\end{IEEEkeywords}

% \fi
\IEEEpeerreviewmaketitle

\section{Introduction}
   The objective of low-rank approximation is to approximate a given matrix by one with low-rank structure. It is an essential tool in many applications like computer vision \cite{turk1991eigenfaces}, signal processing \cite{parker2005signal}, recommender systems \cite{drineas2002competitive}, natural language processing \cite{ren2019label}, machine learning \cite{murphy2012machine}, principal component analysis (PCA) \cite{wold1987principal} and data mining \cite{skillicorn2007understanding}, to name a few examples. The reason for finding a low-rank approximation is that if we know in advance that the  given matrix possesses low rank structure, then doing a low-rank approximation is a neat way to strip off meaningless noise and obtain a more compact representation. The distance between original matrix and approximate matrix is usually measured by the Frobenius norm. On this basis, the optimal low-rank approximation can be obtained by truncated singular value decomposition (SVD). However, considering a squared matrix $A$ with dimension $n$, the computational complexity of SVD is up to $O(n^3)$. This is prohibitive for large-scale datasets, especially when data is collected sequentially or parallelly, for instance, various applications receive data on the fly by varying the time period including online advertising \cite{zhang2008detecting}, sensor network \cite{bonnet2001towards} and network traffic \cite{gilbert2001quicksand}.
   
   One popular solution to remedy the computational burden for processing large data matrices is the so-called matrix sketching technique. The main idea is to construct a sketch matrix $B$ which is much smaller than the original matrix $A$ but can retain most of the information of $A$, and then use $B$ instead of $A$ to do the subsequent operations, such as SVD. More precisely, given an $n \times d$ matrix $A$, the goal is to find an $\ell \times d$ matrix $B$ with $\ell \ll n$ such that $A^{T} A \approx B^{T} B$. The efficiency of doing operations on the concisely representable sketch matrix makes this technique widely used in various applications, including dimension reduction~\cite{drineas2006fast}, online learning~\cite{luo2016efficient}, clustering~\cite{yoo2016streaming}, among many others.
   
 For getting an approximate sketch matrix, many randomized techniques have drawn great attention, such as random sampling and random projection. The random sampling technique \cite{bhojanapalli2014tighter,mahoney2011randomized,bach2013sharp} obtains a precise representation of $A$ by sampling a small number of rows or columns and reweighting them. The most well-known random sampling technique is the leverage score sampling, in which the sampling probability is proportional to the leverage score of each column. This obviously poses the difficulty that the leverage score involves the calculation of the singular vectors of $A$, and thus hard to process streaming data and large-scale data. Therefore, one may pay more attention to the random projection technique \cite{halko2011finding} whose key is to find a random matrix $X$ used to project the original matrix $A$ to a much smaller matrix $B$. It is requested that the construction of $X$ should guarantee that $B$ captures the principal subspace of $A$. In addition, even a single pass over data is sufficient \cite{tropp2017practical}, this enables approximation for dense matrices that cannot be loaded completely in memory.
  \begin{figure*}[htp]
   	\centering
   	\subfigure[]{
   		\includegraphics[width=0.9\columnwidth]{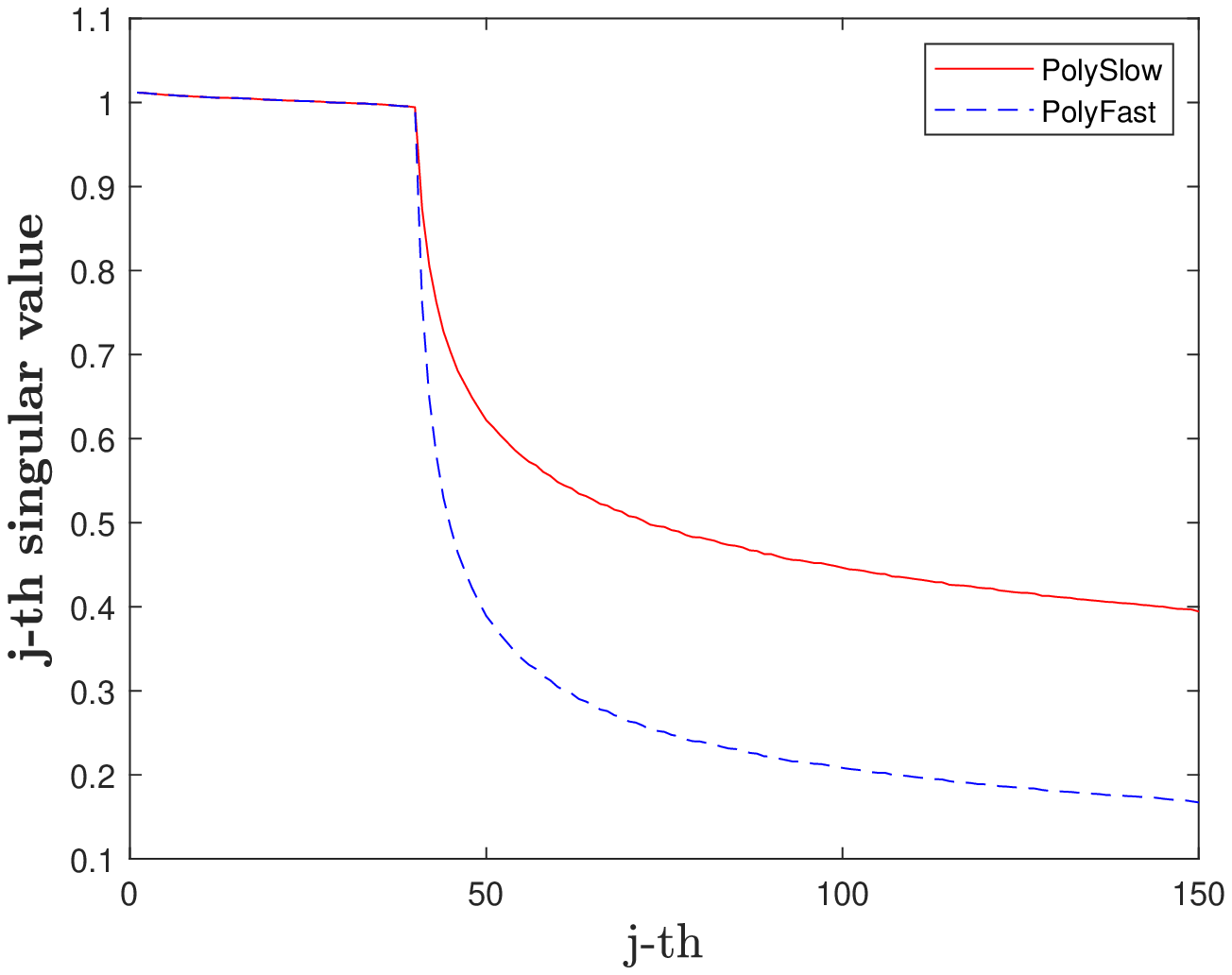}
   	}
   	\subfigure[]{
   		\includegraphics[width=0.9\columnwidth]{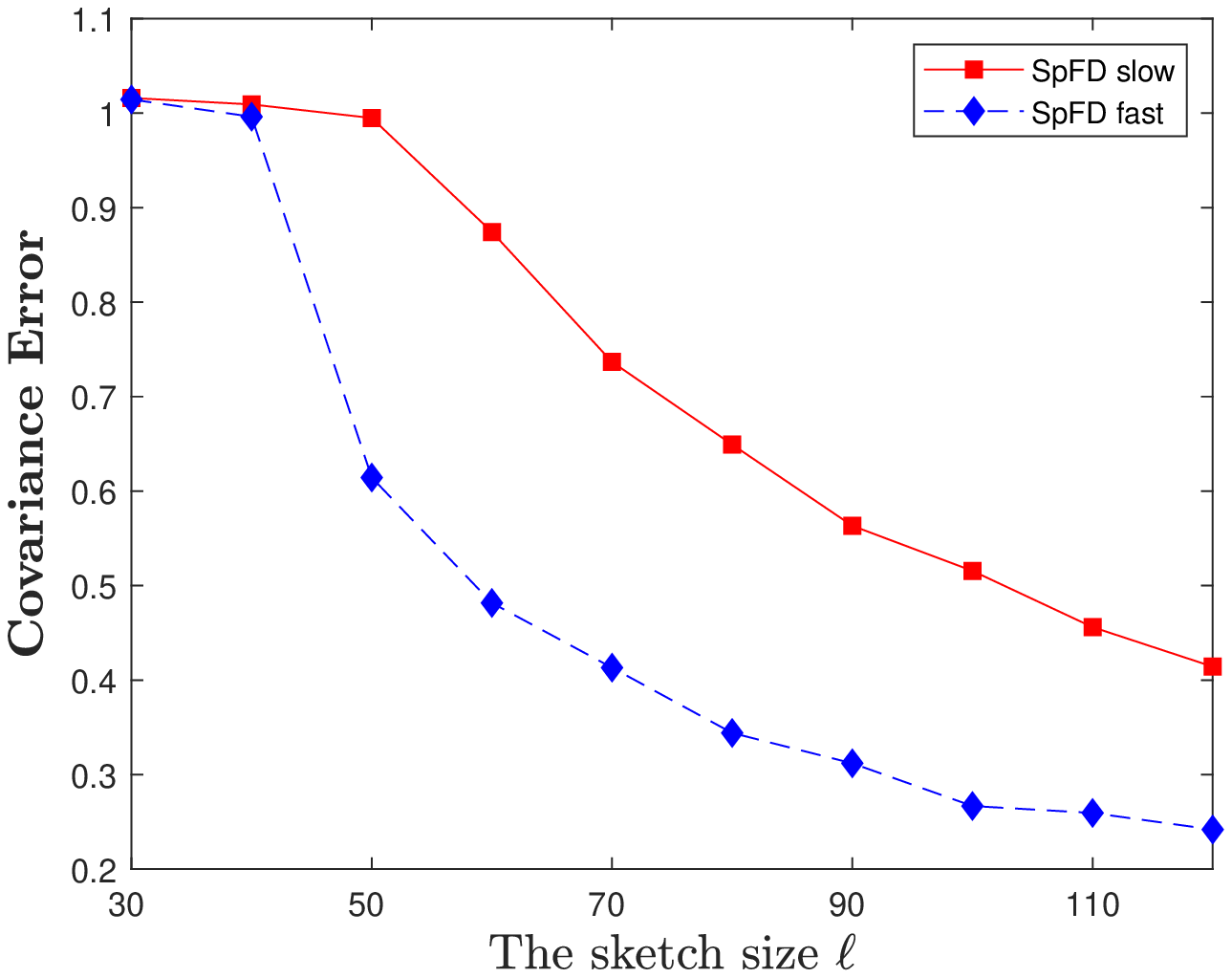}
   	}
   	\caption{The approximation on two synthetic data with different singular decaying spectrum: (a) the distribution of singular values; (b) the covariance error calculated as $\|A^T A - B^T B\|_2$ by SpFD10, a variant of SpFD method \cite{teng2018fast}.}
   	\label{fig:com}
   \end{figure*}
   
   Besides the aforementioned randomized method for constructing the sketch, a deterministic method named Frequent Directions (FD) \cite{liberty2013simple, ghashami2016frequent} was proposed recently.  The inspiration behind  this method is from estimating item frequency in a stream \cite{misra1982finding}. Precisely, for a given $n \times d$ matrix $A$, FD processes rows of $A$ in a stream and maintains an $\ell \times d$ sketch $B$. Then by setting $\ell = k + \frac{k}{\varepsilon}$, it achieves $(1+\varepsilon)$ best rank-$k$ approximation while costs $O(d\ell)$ space and runs in $O(nd\ell)$ time. Hence when facing massive data, such as surveillance video data, the efficiency is still limited due to the space restriction. Several attempts have been made to tackle this issue by combining the idea from random projection \cite{teng2018fast, chen2017frosh, chen2019making}. For example, Teng and Chu \cite{teng2018fast} proposed a method named SpFD that uses the CountSketch matrix to capture more information from the original matrix, thus can accelerate the computation significantly.  However, there still  have room for further improvement, considering that the projection procedure cannot be often capable of compressing the information accurately, leading to deteriorate the performance. As a typical case shown in Fig. 1, when the singular values decay rapidly, SpFD can obtain a good approximation when the sketch size is over 50. However, when the decrease is slow, SpFD could not provide a satisfactory result even for a larger sketch size. This mainly because that the efficiency of random projection relies on the gap between $k$-th singular value and $(k+1)$-th singular value~\cite{musco2015randomized}. Therefore, to improve the accuracy  of  the randomized variants of FD, there is an urgent need for finding a more accurate projection subspace.

   The Krylov subspace method was firstly introduced in solving a system of linear equations \cite{lanczos1950iteration}, then generalized by Golub et al. \cite{golub1977block}, and more recently, Musco et al.~\cite{musco2015randomized} brought it into the randomized SVD and established the first gap-free theoretical analysis for the low-rank approximation. Compared with another popular technique, i.e., subspace iteration~\cite{Gu2014SubspaceIR, halko2011finding}, for improving the precision in low-rank approximation, Krylov subspace method utilizes less iterations for achieving the same precision theoretically and experimentally.  This motivates us to apply the Krylov subspace method to accelerate the FD algorithm, while could maintain its precision. 
  
To sum up, we propose in this work a fast and accurate Frequent Directions algorithm by incorporating the power of Krylov subspace and random projection techniques. Primarily, each block of matrix $A$ is compressed by a non-oblivious projection matrix constructed from Krylov subspace and then embedded into the FD procedure to update the sketch matrix $B$. Our main contributions are summarized as follows:
   \begin{itemize}
   	\item[1)] The newly proposed FD algorithm named r-BKIFD that can skillfully integrate Block Krylov Iteration into randomized FD, so that we can obtain a more accurate subspace during projection procedure. Taking the dense Gaussian random matrix and the sparse CountSketch matrix as examples, we demonstrate the merits of the proposed r-BKIFD in terms of the approximation error and computational speed.
   	\item[2)] The theoretical analysis shows that our method has the comparable theoretical guarantees with original FD. Specifically, we derive the error bounds in terms of  both the covariance and projection errors. It is also shown that such error bounds would be arbitrarily small if we choose an appropriate number of iterations.
   	\item[3)] Extensive experiments are carried out on both synthetic and real data to show that the proposed r-BKIFD outperforms over traditional FD and its variants in most cases.
   \end{itemize} 
\noindent \textbf{Notations.} For an $n \times d$ matrix $A$, $a_{i}$ denotes its $i$-th row,  $a_{ij}$ denotes its $(i, j)$-th element. The matrices $I_n$ represents the $n$-dimensional identity matrix and $0^{n \times d}$ is the all-zero valued matrix with dimension $n \times d$. The Frobenius norm of $A$ is defined as $\|A\|_{F}=\sqrt{\sum_{i}\sum_{j}a_{ij}^2}$, and the spectral norm of $A$ is $\|A\|_{2}=\sup _{x \neq 0}\|A x\| /\|x\|$. The rank-$k$ approximation of $A$ is expressed as $A_{k}=U_{k}\Sigma_{k}V_{k}^{T}$, where $A=U\Sigma V^T$ represents the SVD of $A$. Let $\sigma_{i}$ denote the $i$-th singular value of $A$. The notation $nnz(A)$ denotes the number of non-zero entries of $A$. And we use $\widetilde{O}(n)$ to hide the logarithmic factor on $n$.
\section{Related work}
	 As a basic dimension reduction method, low rank approximation of large-scale matrices is an ubiquitous tool in scientific computing, machine learning, numerical analysis, among a number of other areas \cite{drineas2006fast, markovsky2012low, ye2005generalized}. It can be mainly formulated as the problem that for a given matrix $M$ and an input parameter $k$, one would like to find a matrix $M^{\prime}$ of rank at most $k$ to minimize the Frobenius norm of the discrepancy between $M$ and $M^{\prime}$, i.e.,
	$$
	\min _{\operatorname{rank}(M^{\prime}) \leq k}\|M-M^{\prime}\|_{F}.
	$$
	The classic Eckart-Young-Mirsky theorem shows the best low-rank matrix approximation can be obtained from the truncated SVD. However, for a matrix $M \in \mathbb{R}^{n \times d}$, the computational complexity of calculating the truncated SVD is $O (nd^{2})$, which is unacceptable for large-scale matrix data. Sketching algorithms have been proposed to alleviate the heavy computational cost, by mapping the input matrix to a smaller surrogate matrix called sketch, thus one  can perform the low-rank approximation on such sketch as an alternative.

\subsection{Randomized Sketching Techniques}
   Popular matrix sketching techniques include many randomized algorithms, such as random sampling \cite{boutsidis2014near} and random projection \cite{bingham2001random}. Random sampling forms a sketch by finding the small subset of rows or columns based on a pre-defined probability distribution. Random projection allows the original matrix to be efficiently processed in a lower dimensional space by using a random matrix, such as Gaussian \cite{dasgupta1999elementary}, CountSketch \cite{clarkson2017low} and subsampled randomized Hadamard transform (SRHT) \cite{tropp2011improved}. The Johnson-Lindenstrauss (JL) Lemma \cite{lindenstrauss1984extensions} shows that such random  matrices can preserve the pairwise distance between any two data points. It is known that the $d\times m$ Gaussian random matrix $S$ is defined in the form of $S=\frac{1}{\sqrt{m}}G$, where each entry of $G$ is sampled i.i.d. from $\mathcal{N}(0,1)$. And the CountSketch matrix stems from estimating the most frequent items in a data stream \cite{charikar2002finding}, further applied in performing low-rank approximation \cite{clarkson2017low}. Mathematically, it is constructed as $X =  D \Phi^{T}\in \mathbb{R}^{d \times m}$, where

\begin{itemize}
	\item $\Phi \in \mathbb{R}^{m \times d}$ is a binary matrix with $\Phi_{h(i), i } = 1$ and $\Phi_{j, i } = 0$ for all $j \ne h(i)$. Here $h$ is a uniformly random map from $[d] \rightarrow [m]$;
	\item $D$ is a $d\times d$ diagonal matrix with each diagonal element chosen from $\{-1,1\}$ with equal probability.  
\end{itemize}
   Note that CountSketch matrix is an extremely sparse matrix due to the structure of one non-zero element per row. Thus, given the input matrix $A\in\mathbb{R}^{n \times d}$ and CountSketch matrix $X\in\mathbb{R}^{d \times m}$, the computation cost to get $AX$ is $O(nnz(A))$, which is superior to the costs of $O(ndm)$ for Gaussian and $O(ndlog(m))$ for SRHT \cite{tropp2011improved, ailon2009fast}. Clarkson $\&$ Woodruff \cite{clarkson2017low} further illustrated CountSketch is the fastest known procedure for low rank approximation. Thus it is well suitable for sparse data and has been frequently applied to various applications, including differential privacy \cite{balu2016differentially}, deep learning \cite{cui2017kernel}, among others.

\subsection{Frequent Directions}
Different from the above randomized techniques, Frequent Directions (FD) algorithm is firstly proposed by Liberty as a deterministic matrix sketching technique in \cite{liberty2013simple}. Instead of projecting or sampling the whole matrix at once, it processes the matrix in a row-update approach. That is, given an input matrix $A \in \mathbb{R}^{n \times d}$, the goal is to construct a sketch matrix $B \in \mathbb{R}^{(\ell-1) \times d}$ which is much smaller than $A$ but is still a good estimation. Precisely, the matrix $B$ is initialized to an all-zero valued matrix. We insert the rows of $A$ into $B$ until it is fullfilled. Then a shrinkage procedure is conducted by computing the SVD and subtracting the squared $\ell$-th singular value from all squared singular values. Considering that the last row of $B$ is always all-zero valued after the shrinkage procedure, we could insert continually until all rows are processed. The running time is dominated by the SVD which takes $O(d \ell^2)$ time, so the total time cost is $O(n d \ell^2)$. Later Ghashami et al. \cite{ghashami2016frequent} modify the original FD by doubling the space of $B$ to reduce the running time to $O(n d \ell)$. See more details in Algorithm $\ref{Fast-FD}$.  It has been showed that FD has the following error bound holds for any $k< \ell$,
\begin{align} \label{FD-property}
\left\|A^{ T} A-B^{T} B\right\|_{2}
\leq  \frac{   \left\|A-A_{k}\right\|_{F}^{2}}{\ell-k}.
\end{align}
Noting that setting\begin{small} $\ell=\lceil k+1/\varepsilon\rceil$\end{small} yields error of \begin{small}$\varepsilon \left\|A-A_{k}\right\|_{F}^{2}$\end{small} , that is, the sketch matrix $B$ is a good low-rank approximation. 
\begin{algorithm}[!htb]
	%\small
	\caption{Fast Frequent Directions (Fast-FD) \cite{ghashami2016frequent}}
	\label{Fast-FD}
	\begin{algorithmic}[1]
		\REQUIRE $\mathrm{A} \in \mathbb{R}^{n \times d}, \text { sketch size } \ell$\\
		\ENSURE $\mathrm{B} \in \mathbb{R}^{(\ell-1) \times d}$\\
		\STATE $B \leftarrow 0^{2\ell \times d}$
		\FOR{$i\in1, \ldots, n$} 
		\STATE Insert $a_{i}$ into a zero valued row of $B$\\
		\IF{$B$ has no zero valued rows}
		\STATE $\left[U, \Sigma, V\right] \leftarrow \operatorname{svd}(B)$\\
		%\STATE $C \leftarrow \Sigma V^{T}$\\
		\STATE $\delta \leftarrow \sigma_{\ell}^{2}$\\
		\STATE $B \leftarrow \sqrt{\max(\Sigma^{2}-\delta I_{2\ell},0)} \cdot V^{T}$
		\ENDIF
		\ENDFOR	
		\STATE return $B\leftarrow B(1 :\ell-1, :)$
	\end{algorithmic}
\end{algorithm}

	Since FD is of high accuracy guarantee and well suitable for the streaming settings, several studies have embedded it into online learning. Boutsidis et al. \cite{boutsidis2014online} proposed the online version of PCA (OPCA) and embedded FD in it to reduce both the time and space complexities. Leng et al. \cite{leng2015online} utilized FD to learn hash functions in the online fashion with low computational complexity and storage space. Recently, many improvements have been considered in improving the accuracy and efficiency.  Luo et al. \cite{luo2019robust} proposed Robust Frequent Directions (RFD) by introducing an adaptive regularizer to improve the approximation error bound by a factor 1/2. Not only sketching the primary matrix $B$, Huang \cite{huang2019near} considered to use the random sampling technique to sketch the part removed in the singular values shrinkage procedure, and proved that it is a space-optimal algorithm with faster running time. Besides, several studies considered the random projection techniques to improve the efficiency. Ghashami et al. \cite{ghashami2016efficient} considered the sparsity of original matrix and combined the randomized SVD to accelerate FD. Chen et al. \cite{chen2017frosh} proposed the so-called Faster Frequent Directions by utilizing subsampled randomized Hadamard transform (SRHT) on each data batch and then performing FD on the compact small matrix. Teng et al. \cite{teng2018fast} combined FD with CountSketch matrix to achieve comparable accuracy with low computational cost.  Considering that the integration of random projection techniques would deteriorate the accuracy, this work aims at finding a more precise projection method without lossing much accuracy while maintaining the algorithm running efficiently.

\subsection{Block Krylov Iteration}
   Historically, the classical Lanczos algorithm was first proposed by Lanczos \cite{lanczos1950iteration} to compute the extremal eigenvalues and corresponding eigenvectors of symmetric matrices, and then generalized by Golub and Kahan \cite{golub1965calculating} to solve the singular value pairs of $m\times n$ non-symmetric matrices. The basic idea is to construct Krylov subspace for initial vector $v$, and then project the original matrix onto this subspace. Thus, by using the eigenvalue pairs of projection matrix to approximate the counterpart of original matrix, we get the  Krylov subspace  described as 
   $$K :=\left[v, A v,A^{2}v, \ldots,A^{q-1}v\right].$$

   Note that the invovled $v$ is a single vector, the works \cite{golub1977block, cullum1974block}  modified the Lanczos algorithm from single vector $v$ to a block of $b$ vectors $V =\left[v_1,  \ldots,v_b\right]$ and builded the Krylov subspace as
   $$K :=\left[V, A V,A^{2}V, \ldots,A^{q-1}V\right].$$

 Compared to classical Lanczos, block Lanczos is more efficient in terms of memory and cache. Recently, with the explosive development of randomized algorithm, randomized block Krylov algorithm has emerged, which could be seen as an integration of classical block Lanczos algorithm with randomized starting matrix $AX$, where $X$ is a random matrix that chosen as {Gaussian} random matrix. The detailed procedure is presented in Algorithm $\ref{BKI}$. 

%Specially, the Krylov subspace is constructed as below. 
%   $$K :=\left[A X,\left(A A^{T}\right) A X, \ldots,\left(AA^{T}\right)^{q} A X\right].$$

   The key idea is to take the random projection $V=A X$ as the initial matrix, instead of the arbitrary set of vectors $V$ that may result in poor convergence. The convergence analysis proposed by Musco and Musco \cite{musco2015randomized} reveals that this method has faster convergence rate with respect to number of iterations, and can capture a more accurate range space, as compared with the popular simultaneous iteration (also known as power iteration) technique, which is defined as
   	$$ K :=\left(AA^{T}\right)^{q} A X.$$ 
   The requirement of $q$ is just $\Theta(\frac{\log d}{\sqrt{\varsigma}})$ for getting the $(1+\varsigma)$  relative-error bound for Block Krylov Iteration, instead of $\Theta(\frac{\log d}{{\varsigma}})$ for power iteration. This shows that Block Krylov Iteration can get the accuracy guarantee with fewer iterations. A more detailed theoretical analysis could be found in \cite{drineas2018structural, yuan2018superlinear}. 
\begin{algorithm}[!htb]
	%\small
	\caption{Block Krylov Iteration \cite{musco2015randomized}}
	\label{BKI}
	\begin{algorithmic}[1]
		\REQUIRE $A \in \mathbb{R}^{n \times d} ,\text { error } \varsigma \in(0,1), \text { rank } k \leq n$\\
		\ENSURE $Z \in \mathbb{R}^{n \times k}$\\
		\STATE $q :=\Theta\left(\frac{\log d}{\sqrt{\varsigma}}\right), X \sim \mathcal{N}(0,1)^{d \times k}$
		\STATE $K :=\left[A X,\left(A A^{T}\right) A X, \ldots,\left(AA^{T}\right)^{q} A X\right]$
		\STATE Orthonormalize the columns of $K $ to obtain $Q$
		\STATE Compute $ M :=Q^{T} A A^{T} Q $
		\STATE Set $ \overline{U}_{k} $ to the top $ k$ singular vectors of $M$
		\STATE return $Z=Q \overline{U}_{k}$
	\end{algorithmic}
\end{algorithm}	
%\begin{figure*}[htp]	
%	\centering	
%	\includegraphics[width=1.6\columnwidth]{pic//krylov.pdf}	
%	\caption{\textcolor{blue}{Comparison between block Krylov iteration and simultaneous iteration.}}
%	\label{fig:illu2}
%\end{figure*}

\section{The proposed algorithm}

   For universal large-scale streaming datasets, although many randomized FD variants achieve low computational cost, the algorithmic accuracy is sacrificed to a certain extent for getting the low-rank approximation.	Considering that Block Krylov Iteration gives nearly optimal low-rank approximation with the fastest known theoretical runtime, we present a new algorithm named r-BKIFD, which incorporates the Block Krylov Iteration technique into FD to reduce the computational cost with better accuracy guaranteed.
  
   The classical Block Krylov Iteration is limited to Gaussian random matrix for starting guess, and we can extend it to CountSketch random matrix by observing that some of the real-world datasets are extremely sparse, such as hyperspectral data \cite{ul2011fast}, recommendation data \cite{tang2012dynamic}, speech spectrograms \cite{kameoka2009robust} and so on. For computing $AX$, when $X$ is the CountSketch matrix, the computation complexity is only $O(nnz(A))$ instead of $O(ndk)$ for Gaussian matrix. We thus could sequentially transform the input matrix without explicitly generating CountSketch matrix for the case that the input matrix couldn't fit in memory. The above excellent properties make CountSketch perform well when constructing the Krylov subspace especially when the data matrices are sparse. Therefore, our subsequent analysis is based on both Gaussian random and CountSketch matrices.

   We now give a detailed description of the proposed r-BKI algorithm. For getting a more accurate approximation, we set $m\ge\ell$. At first, we apply $\left(AA^{T}\right)^{i} A\ (i=0,1,\ldots,q)$ to form the Krylov matrix $K$ which contains all the information accumulated along the projection process, then we employ an orthonormal procedure to obtain the newly compressed matrix $Z$ and $P$. See Algorithm $\ref{r-BKI}$ for detailed description of r-BKI.

\begin{algorithm}[!htb] 
	%\small
	\caption{r-BKI}
	\label{r-BKI}
	\begin{algorithmic}[1]
		\REQUIRE $A \in \mathbb{R}^{n \times d} ,\text { error } \varsigma \in(0,1), \text { integers } m,  \ell$\\
		\ENSURE $Z \in \mathbb{R}^{n \times \ell},P \in \mathbb{R}^{\ell \times d}$\\
		\STATE $q :=\Theta\left(\frac{\log d}{\sqrt{\varsigma}}\right),$ let $ X \in \mathbb{R}^{d \times m}$ be a randomized matrix
		\STATE $ K :=\left[A X,\left(A A^{T}\right) A X, \ldots,\left(AA^{T}\right)^{q} A X\right]$
		\STATE Orthonormalize the columns of $ K $ to obtain $Q \in \mathbb{R}^{n \times (q+1) m} $
		\STATE Compute $ M :=Q^{T} A A^{T} Q \in \mathbb{R}^{(q+1) m \times (q+1) m}	$
		\STATE Set $ \overline{U}_{\ell} $ to the top $ \ell$ singular vectors of $M$
		\STATE $Z=Q\overline{U}_{\ell}$
		\STATE return $P=Z^{T}A$
	\end{algorithmic}
\end{algorithm}	

Then we illustrate how to integrate r-BKI into FD. Given the streaming data $A =\left[A_{(1)} ; \cdots ; A_{(s)}\right]\in \mathbb{R}^{n \times d}$ with each $A_{(i)}\in \mathbb{R}^{\frac{n}{s} \times d}\ (i=1,2,\ldots,s)$, our goal is to obtain a small sketch $B\in \mathbb{R}^{(\ell-1) \times d}$ that offers a good performance in preserving crucial information of the original matrix $A$. We assume that $n/s$ is an integer, otherwise we can change it into an integer by appending zero rows to $A$. Compared with the classical FD which directly performs singular values shrinkage procedure on each $\ell$ rows of $A$, we mainly embed r-BKI technique for each batch $A_{(i)} (i=1,2,\ldots,s)$ to obtain an intermediate sketch matrix with more compact representation but preserves accuracy, and then perform Fast-FD on the intermediate sketch matrix. Here r-BKI is used to find a more accurate subspace representation with less computational complexity during the projection procedure. 

\begin{figure*}[htp]	
	\centering	
	\includegraphics[width=2\columnwidth]{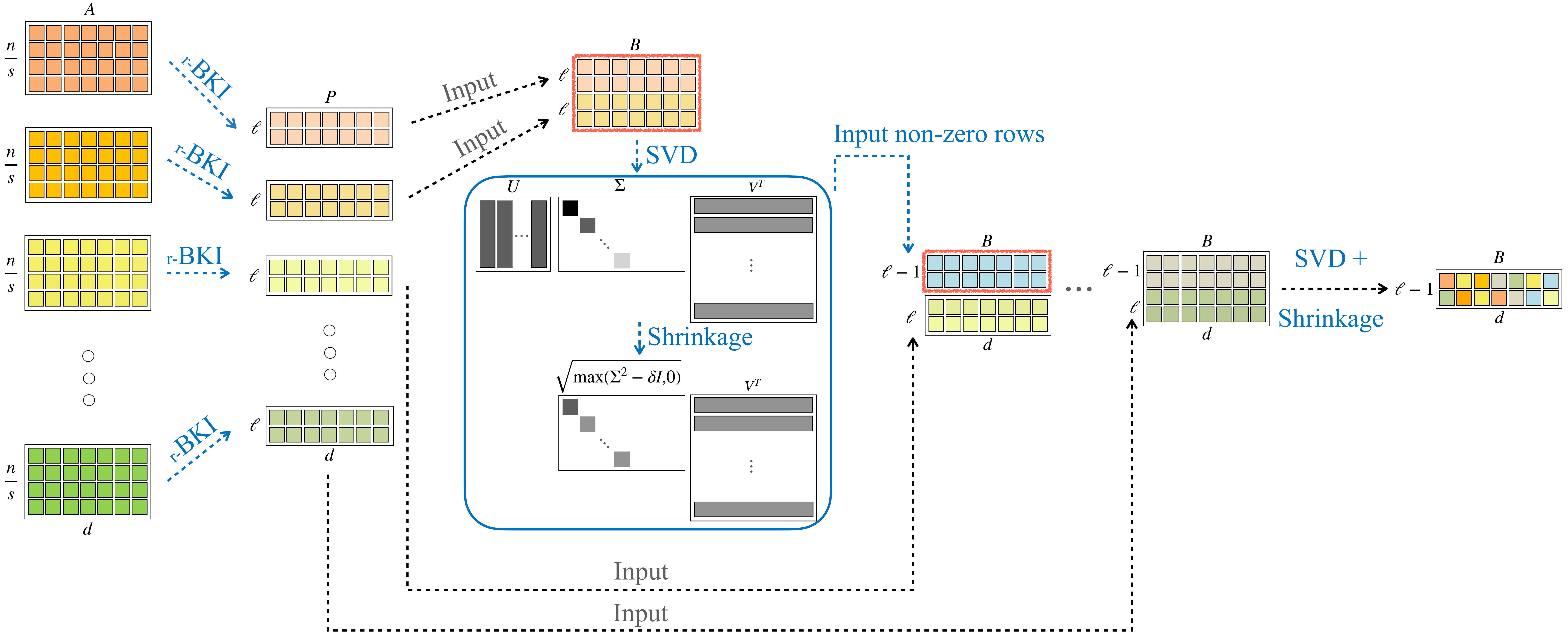}	
	\caption{Illustration of r-BKIFD.}
	\label{fig:illu}
\end{figure*}

\begin{algorithm}[!htb]  
	\caption{r-BKIFD}
	\label{r-BKIFD}
	\begin{algorithmic}[1]
		\REQUIRE $A \in \mathbb{R}^{n \times d} ,\text { error } \varsigma \in(0,1), \text { integers } m,  \ell$\\
		\ENSURE $\mathrm{B} \in \mathbb{R}^{(\ell-1) \times d}$\\
%		\STATE  Set $B \in 0 ^{\ell \times d}$\\
		\STATE  $B  \leftarrow \text{r-BKI}\left( A\left(1 : \frac{n}{s}, :\right),\varsigma, m, \ell \right)$\\
%		\STATE $B \leftarrow P$\\
		\FOR{$i\in1, \ldots, s-1$} 
		\STATE $P \leftarrow \text{r-BKI}\left( A\left(i \frac{n}{s}+1 :(i+1) \frac{n}{s}, :\right),\varsigma, m, \ell \right)$\\
		\STATE $B=\left[B;P\right]$
		\STATE $\left[U, \Sigma, V\right] \leftarrow \operatorname{svd}(B)$\\
		%\STATE $C \leftarrow \Sigma V^{T}$\\
		\STATE $\delta \leftarrow \sigma_{\ell}^{2}$\\
		\STATE $B \leftarrow \sqrt{\max(\Sigma^{2}-\delta I,0)} \cdot V^{T}$\\
		\STATE $B \leftarrow B(1:\ell-1,:)$
		\ENDFOR	
		\STATE return $B$
	\end{algorithmic}
\end{algorithm}	

The detailed procedure is listed in Algorithm 4. Precisely, for each batch $A_{(i)}$, we apply the r-BKI algorithm to compress it into a relatively small intermediate sketch matrix $P$. The sketch matrix $B$  is initialized as the first  intermediate  matrix $P$.  Then for the rest of data, each time we append the sketch $P$ into $B$. Similar to the traditional FD, we perform  the singular values shrinkage procedure and maintain the first $\ell-1$ rows of $B$, and as a result, the remaining rows of B are set to be zeros and replaced by the next intermediate sketch matrix P. This iterative process continues until all batches are processed. The illustration is shown in Fig. \ref{fig:illu}. 

%The detailed procedure is listed in Algorithm 4. Precisely, we first initialize the sketch matrix $B$ as an all-zero valued matrix. Then at each iteration, for each batch $A_{(i)}$, we apply the r-BKI algorithm to compress it into a relatively small intermediate sketch matrix $P$, and input rows of $P$ generated in the previous step into the all zero rows of matrix $B$. Similar to traditional FD,  we finally perform the singular values shrinking procedure  to make the last $\ell+1$ rows of $B$ reset to be zero. This iterative process continues until all batches are processed. The illustration is shown in Fig. \ref{fig:illu}. 
%\textcolor{blue}{It is worth mentioning that on each loop, the $(\ell-1)\times d$-dimensional matrix $B$ was appended to $2\ell-1$ rows by r-BKI, and then its last $\ell$ rows were reset to 0 by an SVD+shrinking process. Finally, we retained the first $\ell-1$ non-zero rows. Notice that in the first loop, the initial matrix $B$ was $\ell$ rows, which were appended to $2\ell$ rows through r-BKI, and then its $\ell+1$ rows were reset to 0 through the SVD+shrinking process. Finally, we retained the former $\ell-1$ non-zero rows. That is to say,} the sketch matrix $B$ maintains the last $\ell+1$ rows is always all-zero valued at each iteration, which makes it possible to implement the algorithm for the continuous influx of data streams. 

\section{Error bounds}
In this section, we theoretically analyze the accuracy of the proposed algorithm r-BKIFD. To this end, we shall first introduce some useful lemmas.

\begin{lemma}[Theorem 2.3 of \cite{drineas2018structural}]\label{lemma1}
	Given data matrix $A\in \mathbb{R}^{n \times d}$ and the
	random matrix $X\in \mathbb{R}^{d \times m}$, let the sketch
	$Z\in \mathbb{R}^{n \times \ell}$ be constructed by Algorithm $\ref{r-BKI}$.
	The best rank-$\ell$ approximation to A can be written as $A_{\ell} = U_{\ell} \Sigma_{\ell} V_{\ell}^{T}$, and let $\ell< rank(A)$.
	If rank $\left(V_{\ell}^{T} X\right)=\ell$, then 
	\begin{small}
		$$ \label{choose}
		\left\|A-Z Z^{T} A\right\|_{2} \leq\left\|A-A_{\ell}\right\|_{2}+\left\|\phi\left(\Sigma_{\ell, \perp}\right)\right\|_{2}\left\|V_{\ell, \perp}^{T} X\left(V_{\ell}^{T} X\right)^{\dagger}\right\|_{F}.
		$$
	\end{small}
\end{lemma}
Lemma \ref{lemma1} characterizes the distance between input matrix $A$ and projection matrix $P$ in Block Krylov Iteration step. Note that the upper bound of $\left\|A-Z Z^{T} A\right\|_{2}$ is closely related to the properties of random matrix $X$. That is to say, this lemma is helpful for choosing appropriate random matrix to estimate the original matrix accurately. A well-behaved random matrix $X$ could tighten the second part $\left\|V_{\ell, \perp}^{T} X\left(V_{\ell}^{T} X\right)^{\dagger}\right\|_{F}$. In the following lemma, we will try to bound $\left\|\phi\left(\Sigma_{\ell, \perp}\right)\right\|_{2}$. 
\begin{lemma}[Lemma 2.4 of \cite{drineas2018structural}]\label{lemma2}
	If $\frac{\sigma_{\ell}-\sigma_{\ell+1}}{\sigma_{\ell+1}} \geq \gamma>0$ holds, then there exists a polynomial $\phi(x)$ of degree $2 q+1$
	with odd powers only, such that $\phi\left(\sigma_{i}\right) \geq \sigma_{i}>0$ for $1 \leq i \leq \ell,$ and
	%\small
	\begin{small}	
		$$
		\left|\phi\left(\sigma_{i}\right)\right| \leq \frac{4 \sigma_{\ell+1}}{2^{(2 q+1) \min \{\sqrt{\gamma}, 1\}}}, \quad i \geq \ell+1
		$$
	\end{small}Hence
	%\small
	\begin{small}	
		$$
		\left\|\phi\left(\Sigma_{\ell}\right)^{-1}\right\|_{2} \leq \sigma_{\ell}^{-1} \quad \text { and } \quad\left\|\phi\left(\Sigma_{\ell, \perp}\right)\right\|_{2} \leq \frac{4 \sigma_{\ell+1}}{2^{(2 q+1) \min \{\sqrt{\gamma}, 1\}}}.
		$$
	\end{small}
\end{lemma}
It is not hard to see that as the number of iterations $q$ and singular value gap $\gamma$ {increase}, $\left\|\phi\left(\Sigma_{\ell, \perp}\right)\right\|_{2}$ will exponentially decay. Moreover, when the sketch size $\ell$ increases, $\sigma_{\ell+1}$ gets smaller, which also makes the bound actually tighter.
\begin{remark}
	As $q$ increases, the error bound is drastically reduced. The essential reason is the introduction of the Krylov subspace. Unlike the power method that aims at computing the dominant eigenspace, Krylov subspace contains the information accumulated along the way is used. This construction of the Krylov matrix makes full use of the orignal matrix, so that less information is lost during projection.
\end{remark}
\begin{lemma}[Matrix Bernstein inequality, \cite{tropp2015introduction}] \label{Matrix-Bernstein-inequality}
	Let $\left\{A_{i}\right\}_{i=1}^{s} \in \mathbb{R}^{n \times d}$ be independent random matrices with $\mathbb{E}\left[A_{i}\right]=0^{n \times d}$ and $\left\|A_{i}\right\|_{2} \leq$
	$R$ for all $i \in[s] .$ Define a variance parameter as $\sigma^{2}=$ $\max \left\{\left\|\sum_{i=1}^{s} \mathbb{E}\left[A_{i} A_{i}^{T}\right]\right\|_{2},\left\|\sum_{i=1}^{s} \mathbb{E}\left[A_{i}^{T} A_{i}\right]\right\|_{2}\right\}$. Then, for all $\epsilon \geq 0$ we have
	$$
	\mathbb{P}\left(\left\|\sum_{i=1}^{s} A_{i}\right\|_{2} \geq \epsilon\right) \leq(d+n) \exp \left(\frac{-\epsilon^{2} / 2}{\sigma^{2}+R \epsilon / 3}\right).
	$$
\end{lemma}

\begin{lemma}[Courant-Fischer min-max theorem, \cite{van1996matrix}]\label{minmax}
	Let $A$ be an $n \times n$ Hermitian matrix with eigenvalues $\lambda_{1} \leq \ldots \leq \lambda_{k} \leq \ldots \leq \lambda_{n}$, then $$\lambda_{k}=\min _{U}\left\{\max _{x}\left\{R_{A}(x) \mid x \in U \text{ and } x \neq 0\right\} \mid \operatorname{dim}(U)=k\right\}$$
	where the Rayleigh-Ritz quotient $R_{A}: \mathbb{C}^{n} \backslash\{0\} \rightarrow \mathbb{R}$ defined by
	$$
	R_{A}(x)=\frac{(A x, x)}{(x, x)}
	$$
	and $(\cdot, \cdot)$ denotes the Euclidean inner product on $\mathbb{C}^{n} .$
\end{lemma}

%\begin{lemma}[Markov's inequality, \cite{vershynin2018high}]\label{Markov}
%	For any non-negative random variable $A$ and $\epsilon\ge0$, we have
%	$$
%	\mathbb{P}(A\ge \epsilon)\le \frac{\mathbb{E}A}{\epsilon}.
%	$$
%\end{lemma}
%Equipped with the lemmas above, we can present the  error bounds of the proposed r-BKIFD in terms of both covariance and projection errors. To this end, we adopt two widely used random matrices in the Block Krylov Iteration step, including Gaussian matrix and CountSketch matrix. 
\subsection{Error Bounds for GA-BKIFD}
It is known that Gaussian random matrix has high quality sketch accuracy and is easy to implement \cite{wang2015practical}. In this subsection, we apply it to the r-BKIFD algorithm and call the algorithm as GA-BKIFD. The theoretical performance is guaranteed by the following theorem.
\begin{theorem}[Covariance error of GA-BKIFD]\label{thm2}
	Given data $A =\left[A_{(1)} ; \cdots ; A_{(s)}\right]\in \mathbb{R}^{n \times d}$, where each $A_{(i)}\in \mathbb{R}^{\frac{n}{s} \times d}$, let the small sketch
	$B \in \mathbb{R}^{(\ell-1) \times d}$ be constructed by Algorithm $\ref{r-BKIFD}$, where $X$ is a Gaussian random matrix. For any $\eta \in(0,1)$ and $\varepsilon\ge0$, if $\frac{\sigma_{\ell}-\sigma_{\ell+1}}{\sigma_{\ell+1}} \geq \gamma>0$, then with probability at least $1-2s\exp(-\varepsilon^{2}/2)-\eta$, we have
	\begin{small}
		\begin{align}
			%	\label{theo:11}
			&\left\|A^{ T} A-B^{T} B\right\|_{2} \notag\\
			\le&\left(s\left(1+\delta\right)+\log \left(\frac{2 d}{\eta}\right) \frac{4 \left(1+\delta\right)}{3}+\sqrt{2 s \left(1+\delta\right)^2 \log \left(\frac{2 d}{\eta}\right)}\right)\notag\\
			&\times\left\|A-A_{\ell}\right\|_{2}^{2}+\frac{   \left\|A-A_{k}\right\|_{F}^{2}}{\ell-k}, \label{snega}
		\end{align}
	\end{small}where $1+\delta=\left(1+\frac{4 }{2^{(2 q+1) \min \{\sqrt{\gamma}, 1\}}}\frac{\sqrt{d-\ell}(\sqrt{d}+\sqrt{m}+\varepsilon)}{\sqrt{m}-\sqrt{\ell}-\varepsilon}\right)^{2}$, $\sigma_{i}$ is the singular values of $A$ in descending order.
\end{theorem}

To explore the trend of the error bound with each variable more conveniently and clearly, we hide the logarithmic factor on $(d, \eta)$. Note that $\frac{   \left\|A-A_{k}\right\|_{F}^{2}}{\ell-k}$ decreases obviously as $\ell$ increases. Thus we focus on analyzing the effect of $\widetilde{O}\left(s(1+\delta)\right)\left\|A-A_{\ell}\right\|_{2}^2$. Firstly, a small increase in $q$ and $\gamma$ can lead to an exponential decay in $\delta$, therefore, for the fixed singular value gap $\gamma$, $\delta$ can become arbitrarily small if an appropriate $q$ is chosen; Secondly, with the increase of sketch size $\ell$, $(\ell+1)$-th singular value (i.e., $\left\|A-A_{\ell}\right\|_{2}$) becomes smaller, we stress that this advantage is even more significant when the singular value gap $\gamma$ is large.

Note that the above analysis is based on covariance error, now we introduce a key lemma which illustrates the relationship between covariance  error and projection error.
\begin{lemma}[covariance error to projection error \cite{huang2019near}] \label{lemma4}
	\begin{align}\label{sne-fne}	
		\left\|A-\pi^k_{B}(A)\right\|_{F}^{2} \leq\left\|A-A_{k}\right\|_{F}^{2}+2 k \cdot\left\|A^{T} A-B^{T} B\right\|_{2} 
	\end{align}
	where $\pi^k_{B}(A)$ is the projection of $A$
	onto the top-$k$ singular vectors of $B$. 
\end{lemma}

This lemma shows that as long as we obtain the error bound of the covariance error, we can also get the error bound of the projection error. This property is very important in the low rank approximation. Many researches focus on the covariance error, because it can reveal the difference between two matrices more substantially.

The following corollary shows the projection  error of GA-BKIFD, which follows by combining (\ref{snega}) and (\ref{sne-fne}).
\begin{corollary}
	Given data $A =\left[A_{(1)} ; \cdots ; A_{(s)}\right]\in \mathbb{R}^{n \times d}$, where each $A_{(i)}\in \mathbb{R}^{\frac{n}{s} \times d}$, let the small sketch
	$B \in \mathbb{R}^{(\ell-1) \times d}$ be constructed by Algorithm $\ref{r-BKIFD}$, where $X$ is a Gaussian random matrix. For any $\eta \in(0,1)$ and $\varepsilon\ge0$, if $\frac{\sigma_{\ell}-\sigma_{\ell+1}}{\sigma_{\ell+1}} \geq \gamma>0$, then with probability at least $1-2s\exp(-\varepsilon^{2}/2)-\eta$, we have
	%\begin{scriptsize}
	\begin{align}
		&\left\|A-\pi^k_{B}(A)\right\|_{F}^{2}   
		\notag\\
		\le&2k\left(s\left(1+\delta\right)+\log \left(\frac{2 d}{\eta}\right) \frac{4 \left(1+\delta\right)}{3}+\sqrt{2 s \left(1+\delta\right)^2 \log \left(\frac{2 d}{\eta}\right)}\right)\notag\\
		&\times\left\|A-A_{\ell}\right\|_{2}^{2}+\frac{\ell+k  }{\ell-k} \left\|A-A_{k}\right\|_{F}^{2}
		, \notag
	\end{align}
	%\end{scriptsize}
	where $1+\delta=\left(1+\frac{4 }{2^{(2 q+1) \min \{\sqrt{\gamma}, 1\}}}\frac{\sqrt{d-\ell}(\sqrt{d}+\sqrt{m}+\varepsilon)}{\sqrt{m}-\sqrt{\ell}-\varepsilon}\right)^{2}$, and $\sigma_{i}$ is the singular values of $A$ in descending order.
\end{corollary}
%\textcolor{blue}{Same as the analysis of covariance  error, the error bound decreases as $\gamma$, $q$ and $\ell$ increase, and the subsequent experiments will further verify our conclusion.}
\subsection{Error Bounds for CS-BKIFD}
As mentioned before, Gaussian random matrix has been well applied to the proposed r-BKIFD algorithm. However, when the dimension of input matrix reaches a larger scale, the time complexity to perform matrix multiplication is too high. This is because it destroys the sparse nature of original matrix if the input matrix is sparse. To address this issue, we introduce the CountSketch matrix with sparse structure to r-BKIFD in this subsection, which is called  as CS-BKIFD. Its error bound in terms of covariance error is listed in the following theorem.
\begin{theorem}[Covariance error of CS-BKIFD]\label{thm5}
	Given data $A =\left[A_{(1)} ; \cdots ; A_{(s)}\right]\in \mathbb{R}^{n \times d}$, where each $A_{(i)}\in \mathbb{R}^{\frac{n}{s} \times d}$, let the small sketch
	$B \in \mathbb{R}^{(\ell-1) \times d}$ be constructed by Algorithm $\ref{r-BKIFD}$, where $X$ is a CountSketch matrix. For any $\varepsilon, p,\eta \in(0,1)$, if $\frac{\sigma_{\ell}-\sigma_{\ell+1}}{\sigma_{\ell+1}} \geq \gamma>0$ and 
	$m \geq \frac{\ell^{2}+\ell}{\varepsilon^{2} p},$
	then with probability at least $1-sp-\eta$, we have
	%\scriptsize
	\begin{small}
		\begin{align}
			&\left\|A^{ T} A-B^{T} B\right\|_{2} \notag \\
			\le&\left(s\left(1+\delta\right)+\log \left(\frac{2 d}{\eta}\right) \frac{4 \left(1+\delta\right)}{3}+\sqrt{2 s \left(1+\delta\right)^2 \log \left(\frac{2 d}{\eta}\right)}\right)\notag\\
			&\times\left\|A-A_{\ell}\right\|_{2}^{2}+\frac{   \left\|A-A_{k}\right\|_{F}^{2}}{\ell-k}, \label{snecs}
		\end{align}
	\end{small}where $1+\delta=\left(1+\frac{4 }{2^{(2 q+1) \min \{\sqrt{\gamma}, 1\}}}\sqrt{\frac{d(d-\ell)}{1-\varepsilon}}\right)^{2}$, $\sigma_{i}$ is the singular values of $A$ in descending order.
\end{theorem}
{
	Similar to the above analysis, the projection error  bound of CS-BKIFD can be obtained immediately by combining (\ref{sne-fne}) and (\ref{snecs}).
}

{\begin{remark}
		The core analysis of the error bound is consistent with the algorithm GA-BKIFD. In addition, as the sketch size $\ell$ increases, the algorithmic accuracy is improved, which will be verified in the experimental study.
\end{remark}}

%\begin{remark}
%	We now argue that the theorem holds with high probability. Taking $\delta=\delta_{0}/s$ where $\delta_{0} \in(0,1)$, then the successful probability can be rewritten as $1-\delta_{0}$ and $m \geq \frac{s(\ell^{2}+\ell)}{\varepsilon^{2} \delta_{0}}$. Note that a larger sketch size $\ell$ can make the approximation error small, that is, $s$ does not need to be too large.
%\end{remark}
\begin{corollary}
	Given data $A =\left[A_{(1)} ; \cdots ; A_{(s)}\right]\in \mathbb{R}^{n \times d}$, where each $A_{(i)}\in \mathbb{R}^{\frac{n}{s} \times d}$, let the small sketch
	$B \in \mathbb{R}^{(\ell-1) \times d}$ be constructed by Algorithm $\ref{r-BKIFD}$, where $X$ is a CountSketch matrix. For any $\varepsilon, p,\eta \in(0,1)$, if $\frac{\sigma_{\ell}-\sigma_{\ell+1}}{\sigma_{\ell+1}} \geq \gamma>0$ and 
	$m \geq \frac{\ell^{2}+\ell}{\varepsilon^{2} p},$
	then with probability at least $1-sp-\eta$, we have
	%\scriptsize 
	\begin{align}
		&\left\|A-\pi^k_{B}(A)\right\|_{F}^{2}  
		\notag\\
		\le&2k\left(s\left(1+\delta\right)+\log \left(\frac{2 d}{\eta}\right) \frac{4 \left(1+\delta\right)}{3}+\sqrt{2 s \left(1+\delta\right)^2 \log \left(\frac{2 d}{\eta}\right)}\right)\notag\\
		&\times\left\|A-A_{\ell}\right\|_{2}^{2}+\frac{\ell+k  }{\ell-k} \left\|A-A_{k}\right\|_{F}^{2}
		, \notag
	\end{align}
	where $1+\delta=\left(1+\frac{4 }{2^{(2 q+1) \min \{\sqrt{\gamma}, 1\}}}\sqrt{\frac{d(d-\ell)}{1-\varepsilon}}\right)^{2}$, and $\sigma_{i}$ is the singular values of $A$ in descending order.
\end{corollary}

\subsection{Comparison of GA-BKIFD and CS-BKIFD}
{
	We shall make a comparison of GA-BKIFD and CS-BKIFD in terms of accuracy and running time. For the algorithmic accuracy, we observe that the covariance error bound can achieve $\widetilde{O}\left(s(1+\delta)\right)\left\|A-A_{\ell}\right\|_{2}^2+\frac{   \left\|A-A_{k}\right\|_{F}^{2}}{\ell-k}$ when $q$ is large, according to Theorems \ref{thm2} and \ref{thm5}. However, the random size $m$ should satisfy $m\ge\ell$ for GA-BKIFD while $m \geq \frac{\ell^{2}+\ell}{\varepsilon^{2} p}$ for CS-BKIFD. Therefore, GA-BKIFD achieves almost the same accuracy guarantees with less sampling numbers. For the algorithmic running time, the matrix multiplication operation generated in the construction of Krylov subspace takes up a lot of time. Fortunately, because of the sparse structure of CountSketch matrix, CS-BKIFD could run faster in this step, that is to say, it has a lower computational complexity just as the following subsection shown, and thus works well in some practical situations. 
}
\subsection{Comparison with Existing Algorithms}
TABLE \ref{comparison} shows detaild comparison in terms of the projection and covariance errors. For easy and intuitive comparison, we rewrite the original error bounds and use $\widetilde{O}$ to hide the logarithmic.\\
\indent First of all,  we can observe that the error bounds of the traditional deterministic algorithm FD are sharper than all these randomized FD variants. This is mainly because that the techniques one use to derive randomized FD variants' error bounds still rely on the properties of FD. Besides, it can be emphasized here that the error bounds of the proposed r-BKIFD are superior to FFD and SpFD, according to the following detailed comparative analysis. \\
\indent In terms of covariance error, our bound is tighter than FFD from two aspects. Firstly, the first term of our bound is related to the $(\ell+1)$-th singular value of $A$ rather than the largest singular value in FFD, which is more advantageous when sketch size $\ell$ is large. Secondly, a small increase in $q$ and $\gamma$ can lead to an exponential decay in $\delta$, and as a result, for the fixed singular value gap $\gamma$, if we choose an appropriate $q$, the $\delta$ in our bound can be arbitrarily small, whereas the $\Delta_t$ in FFD cannot be. The reason is that $\Delta_t=\Theta\left(\sqrt{\frac{\min(\frac{n}{s},d) \log \left(2 \min(\frac{n}{s},d)/ \delta\right)}{\ell/2}}\right)$, and one can observe that $\Delta_t$ can not decay exponentially with one of the variables. We stress that such two advantages come from the incorporation of Block Krylov Iteration technique.\\
\indent In terms of projection error, our bound is much tighter than SpFD in most cases. That is, it is easy to check that when $\delta_1\le \frac{1}{2}$ and $\ell \le \frac{k}{8s\delta_1  (1+\delta)\log\left(\frac{2 d}{\eta}\right)}+k$, r-BKIFD can achieve a tighter upper bound than SpFD. And such two assumptions can be satisfied most often, because the failure probability $\delta_1$ should be small. By the way, although \cite{teng2018fast} aims to analyze the error bound $\left\|A-[AV]_kV^T\right\|_F^2$ of low-rank approximation, it is essentially analyzing the projection error due to the use of inequality $\left\|A-[AV]_kV^T\right\|_F^2\le \left\|A-AV_kV_k^T\right\|_F^2$ in our analysis.

\begin{table*}[!htbp]
		\begin{center}
	\caption{Comparison with FD and its randomized variants}\label{comparison}
	\begin{threeparttable}
		\begin{tabular}{ccc}
			\toprule
			Algorithm  & Covariance error & Projection error \\
			\midrule
			FD \cite{ghashami2016frequent}	&$\frac{1  }{\ell-k} \left\|A-A_{k}\right\|_{F}^{2}$	&$\left(1+\frac{k  }{\ell-k}\right) \left\|A-A_{k}\right\|_{F}^{2}$	 \\
			FFD \cite{chen2017frosh}	&$\max\limits_i\widetilde{O}\left(\sqrt{s}(1+\Delta_t)\left\|A_{(i)}\right\|_{2}^2+\varepsilon \left\|A-A_{k}\right\|_{F}^{2}\right)$ &N/A	 \\
			SpFD \cite{teng2018fast}	&N/A	&$\frac{k}{(\ell-k)\delta_1}\left\|A_k\right\|_{F}^{2}+(1+\frac{k}{(\ell-k)\delta_1})\left\|A-A_{k}\right\|_{F}^{2}$	 \\
			r-BKIFD (this work)	&$\max\limits_i\widetilde{O}\left(s(1+\delta)\left\|A_{(i)}-[A_{(i)}]_{\ell}\right\|_{2}^2+\varepsilon \left\|A-A_{k}\right\|_{F}^{2}\right)$	&$8ks\left(1+\delta\right)\log \left(\frac{2 d}{\eta}\right) 
			\left\|A-A_{\ell}\right\|_{2}^{2}+\frac{\ell+k  }{\ell-k} \left\|A-A_{k}\right\|_{F}^{2}$	 \\
			\bottomrule
		\end{tabular}
	\end{threeparttable}
		\end{center}
\end{table*}

\subsection{Complexity Analysis}
The running time of r-BKIFD is dominated by the step 4 of performing r-BKI, that is, the procedure for computing the sketch matrix $P$. Thus, if  the submatrix $A_i \in \mathbb{R}^{b \times d}$ is chosen as the  Graussian random matrix, it needs the time of $O(bdmq)$ to construct the Krylov space $K$; while if  $A_i$ is chosen as CountSketch matrix, it only needs $O(nnz(A_i)q)$ time to get $K$. Then the QR decomposition is further needed to obtain $Q$, which requires the time of $O(b (mq)^2)$. And the truncated SVD could cost the time of $O((mq)^2 \ell)$ to get $\overline{U}_{\ell}$. After that, the main cost lies in the step 6 of performing SVD on $B$, which costs $O(d \ell^2)$. Summarizing all these calculations, the computational cost of each iteration is about $O(bdmq + bm^2q^2+ m^2q^2 \ell +  d \ell^2)$. Considering we only should proceed $\frac{n}{s}$ rows, the total cost for GA-BKIFD is $O(ndmq + nm^2 q^2 + sm^2q^2 \ell + s d \ell^2)$. Further noting that $\sum_i nnz(A_i) = nnz(A)$, it is easy to conclude that the total cost for CS-BKIFD is $O(nnz(A)q + nm^2 q^2 + sm^2q^2 \ell+ s d \ell^2)$. Therefore,  CS-BKIFD would be a better choice for large-scale sparse datasets.

\section{Experiments}
In this section, the proposed r-BKIFD is compared with three popular algorithms, namely FD \cite{ghashami2016frequent}, SFD \cite{ghashami2016efficient}  and SpFD10 \cite{teng2018fast}, through a series of synthetic and real data experiments. All such algorithms are implemented in MATLAB 2018a on a 56-core CPU (2.20 GHz) with 128 GB of RAM,  and we run each method 30 times and take the average result. Their detailed information is listed as follows:
\begin{itemize}
	\item[1] FD: Algorithm 1.
	\item[2] SFD \cite{ghashami2016efficient}: It is a randomized FD algorithm which bases on Gaussian random matrix and power iteration.
	\item[3] SpFD10: It is a randomized FD algorithm by utilizing sparse subspace embedding method, which chooses $q=10$ to balance the precision and computational time of the SpFDq algorithm proposed by \cite{teng2018fast}. 
	\item[4] GA-BKIFD: Following Algorithm 4, here we choose to use standard Gaussian random matrix in the Block Krylov Iteration step.
	\item[5] CS-BKIFD: Following Algorithm 4, here we choose to use standard CountSketch matrix in the Block Krylov Iteration step.
	
\end{itemize}

For measuring the accuracy of these computing algorithms, we consider both the covariance and projection errors.  The covariance error is defined as $\|A^TA - B^TB\|_2/\|A\|_F^2$, which measures the difference in singular values. And the projection error is defined by projecting $A$ onto the top-$k$ singular vectors of  $B$, i.e., $\|A - \pi^k_{B}(A)\|_F/\|A - A_k\|_F^2$. Moreover, we also measure the computational cost by changing the sketch size $\ell$.

\begin{figure*}[htp]

	\centering
	\subfigure{
		\includegraphics[width=0.42\columnwidth]{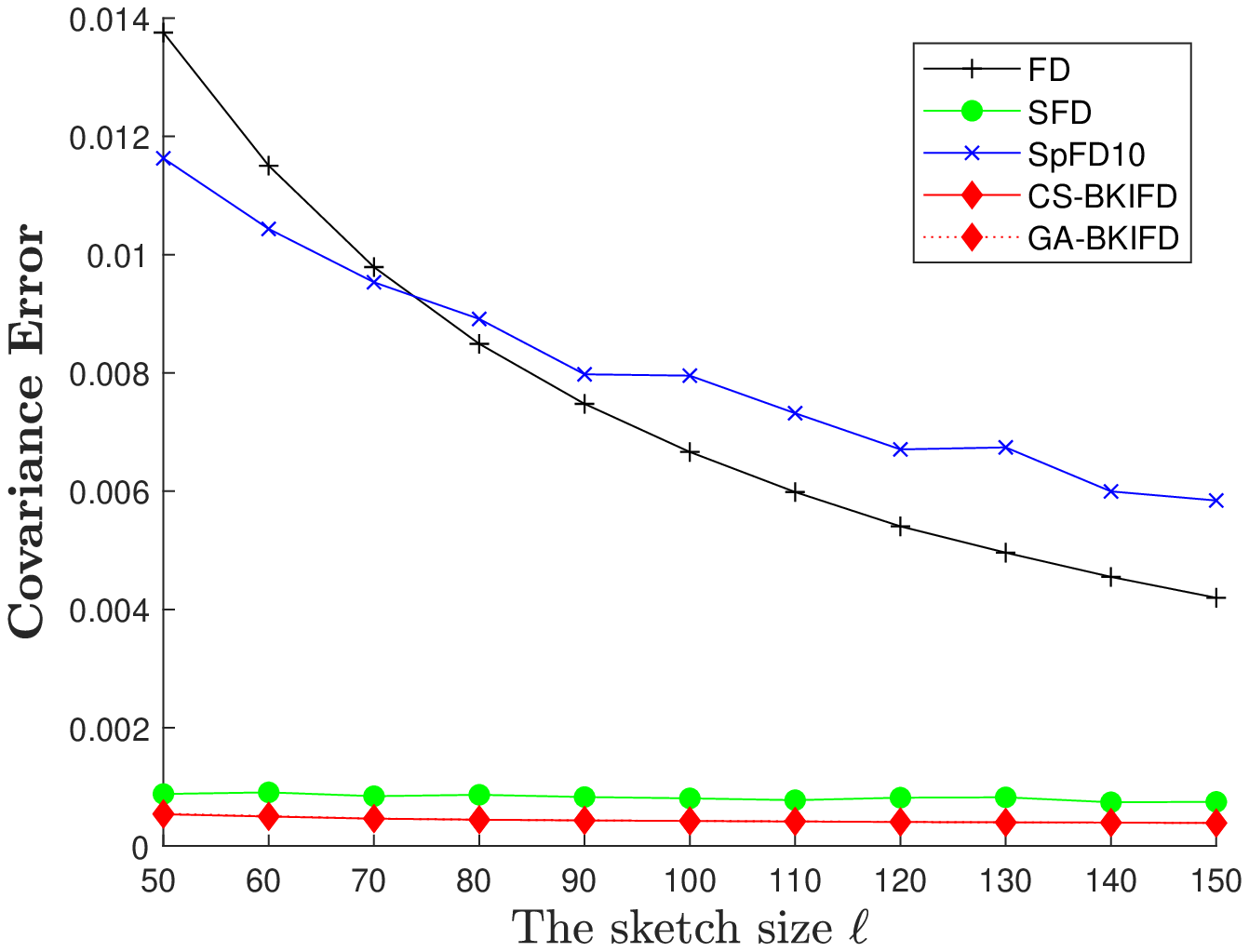}
	}
	\subfigure{
		\includegraphics[width=0.42\columnwidth]{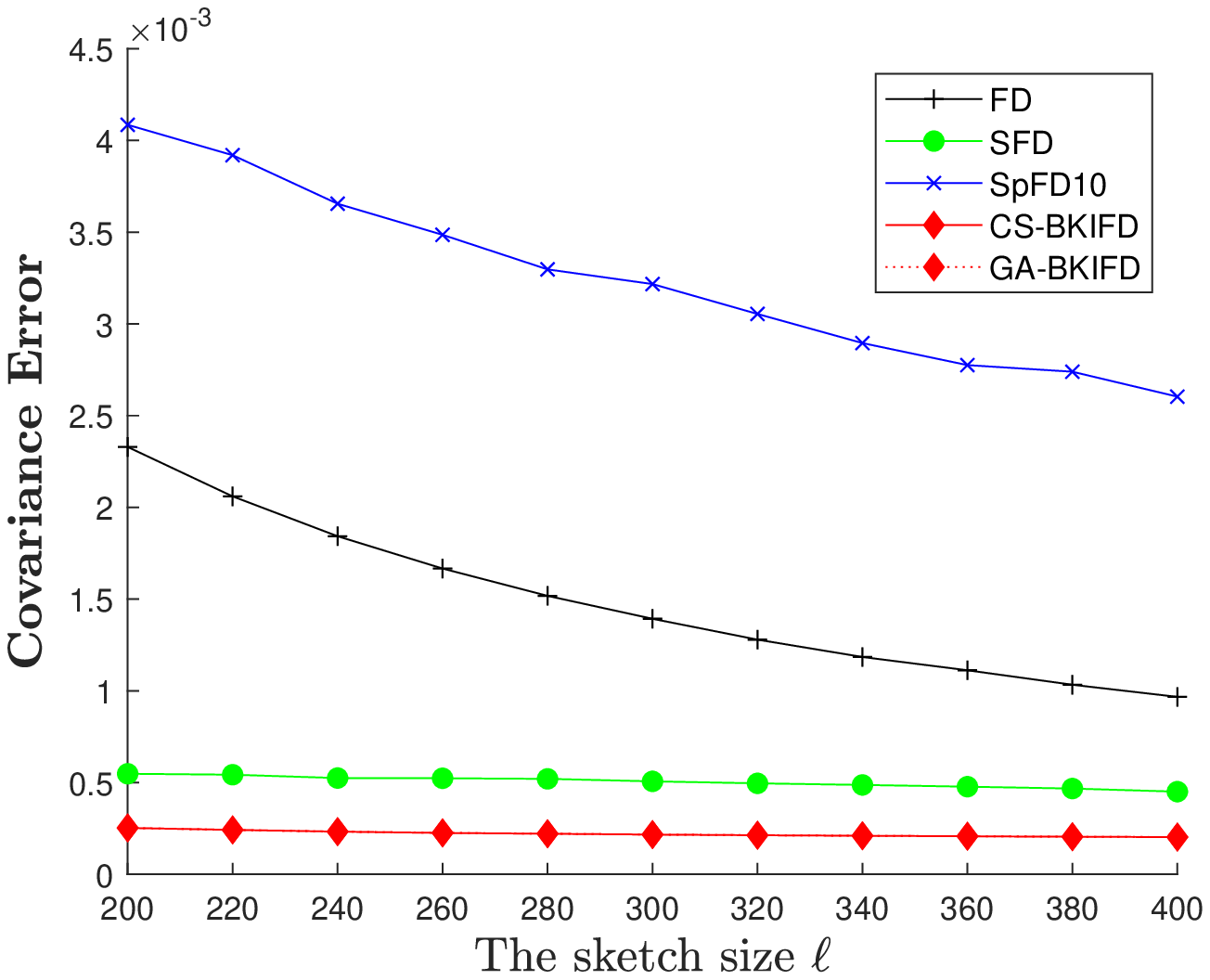}
	}
	\subfigure{
		\includegraphics[width=0.42\columnwidth]{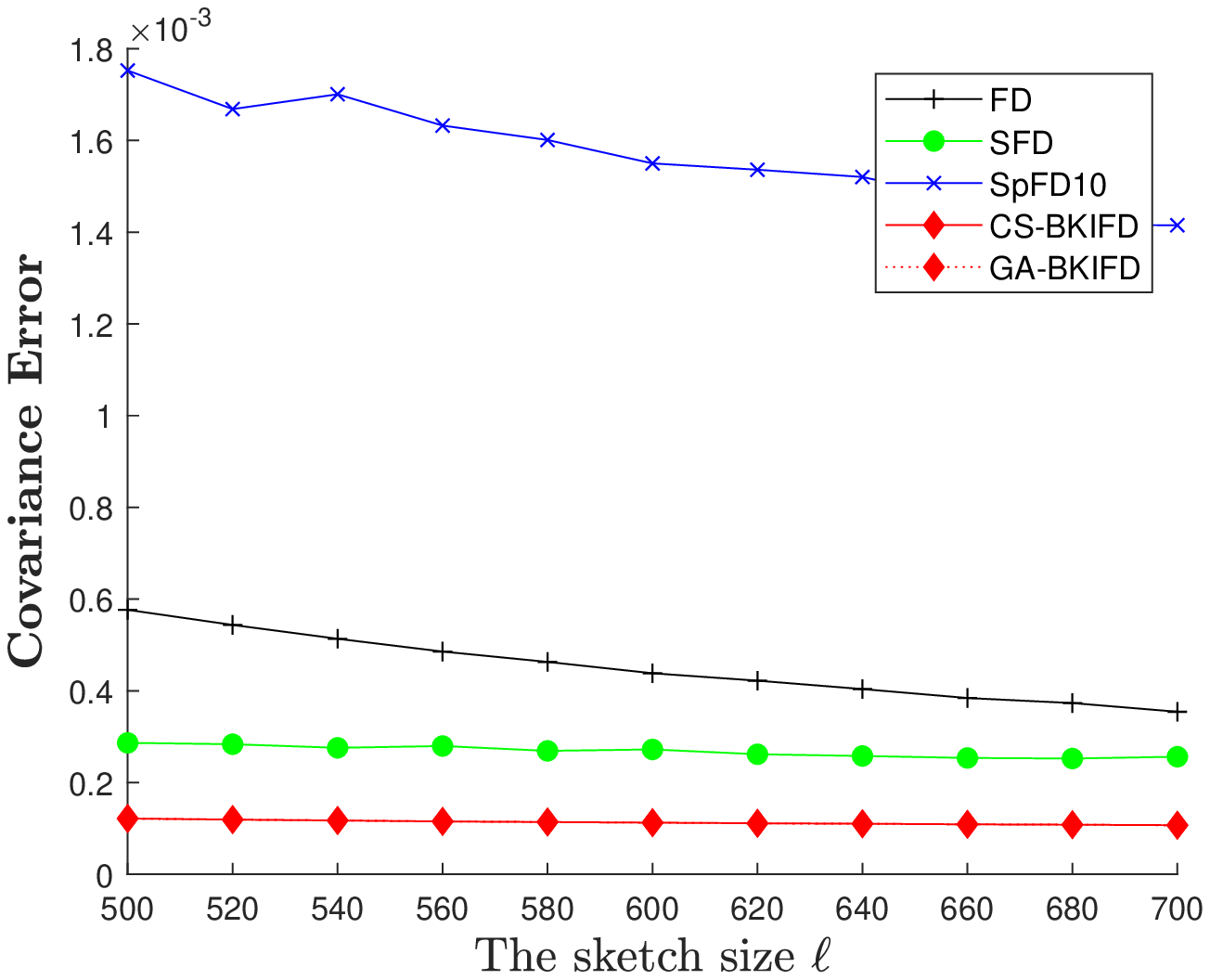}
	}
	\subfigure{
		\includegraphics[width=0.42\columnwidth]{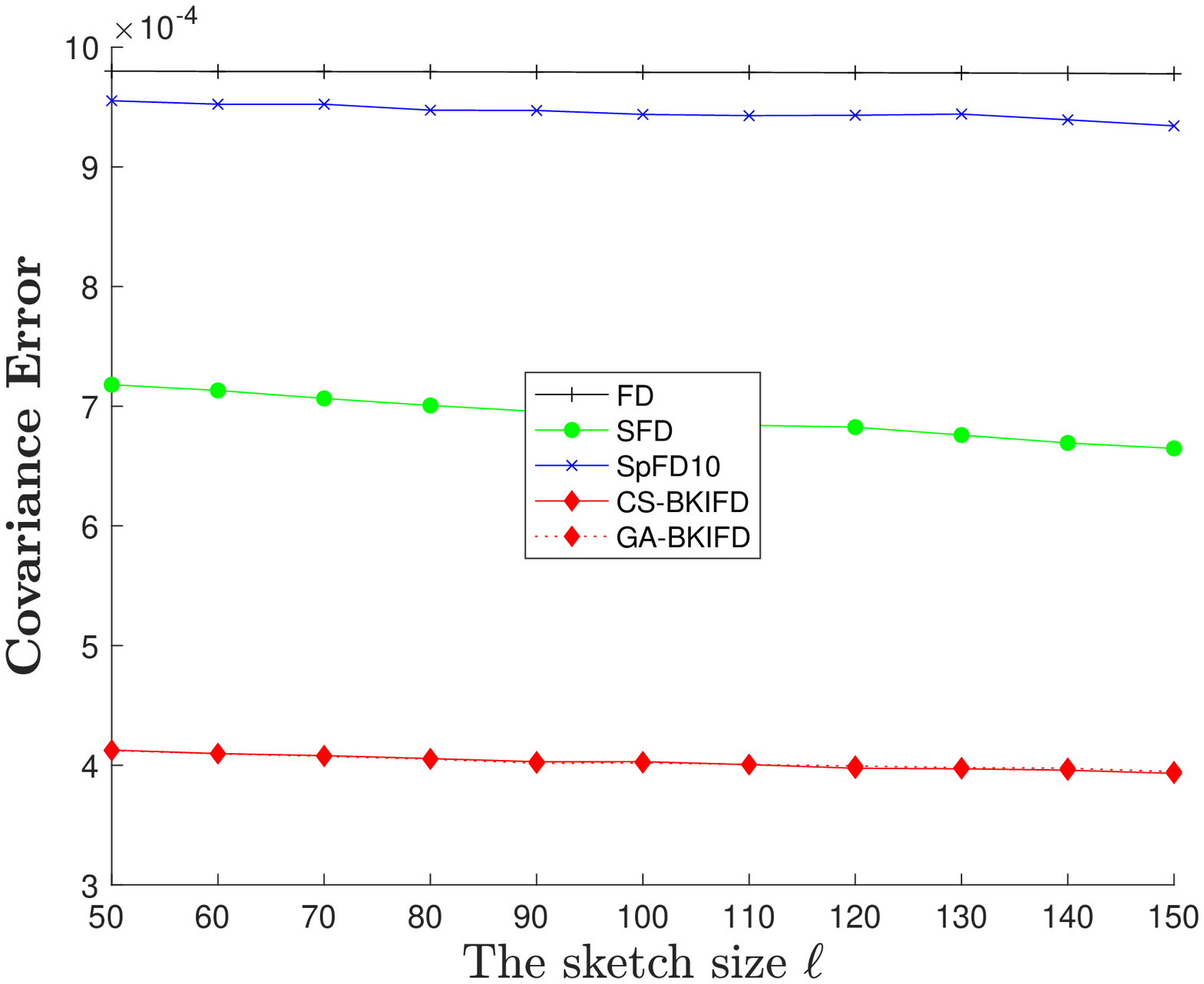}
	}
	\\
	\subfigure{
		\includegraphics[width=0.42\columnwidth]{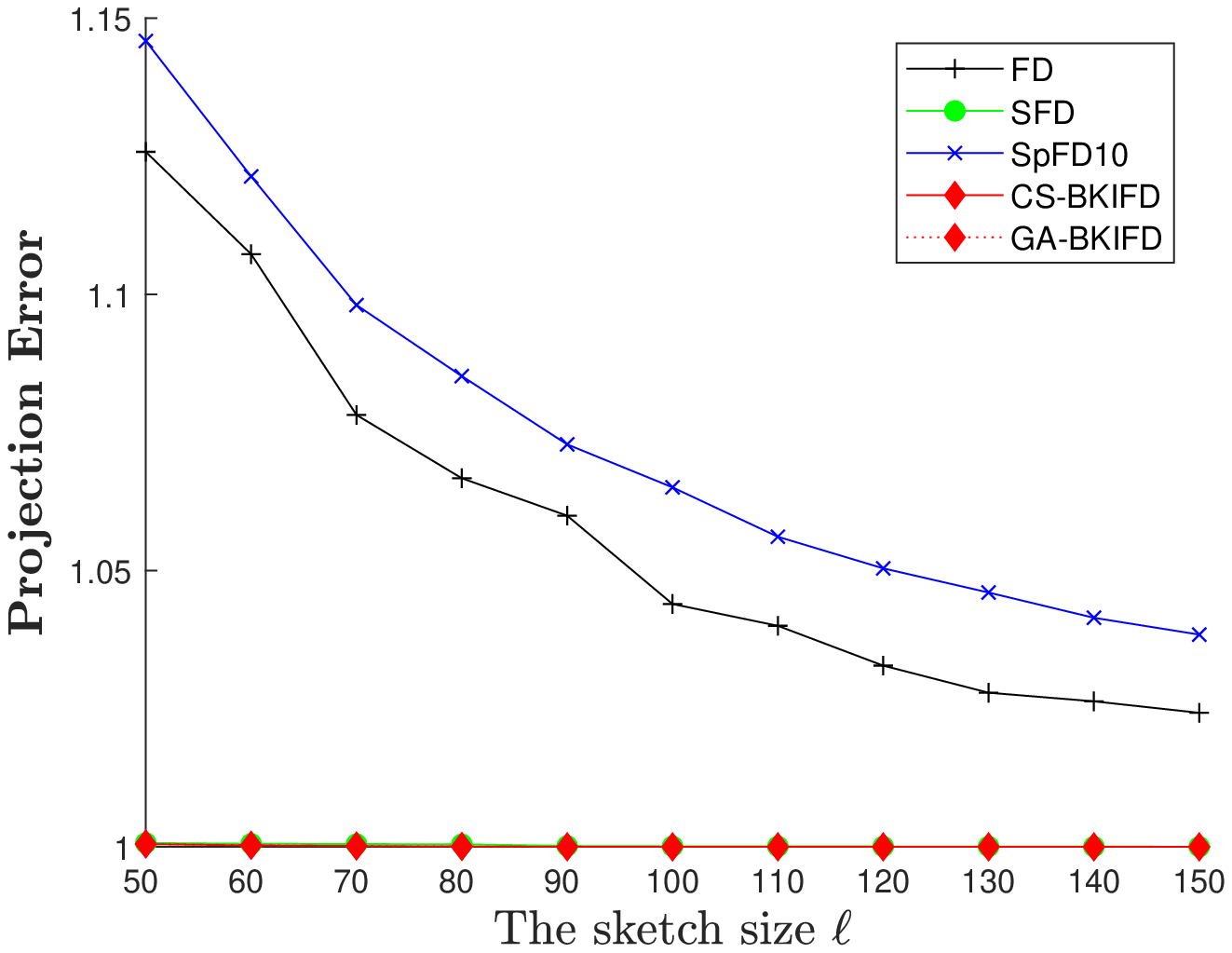}
	}
	\subfigure{
		\includegraphics[width=0.42\columnwidth]{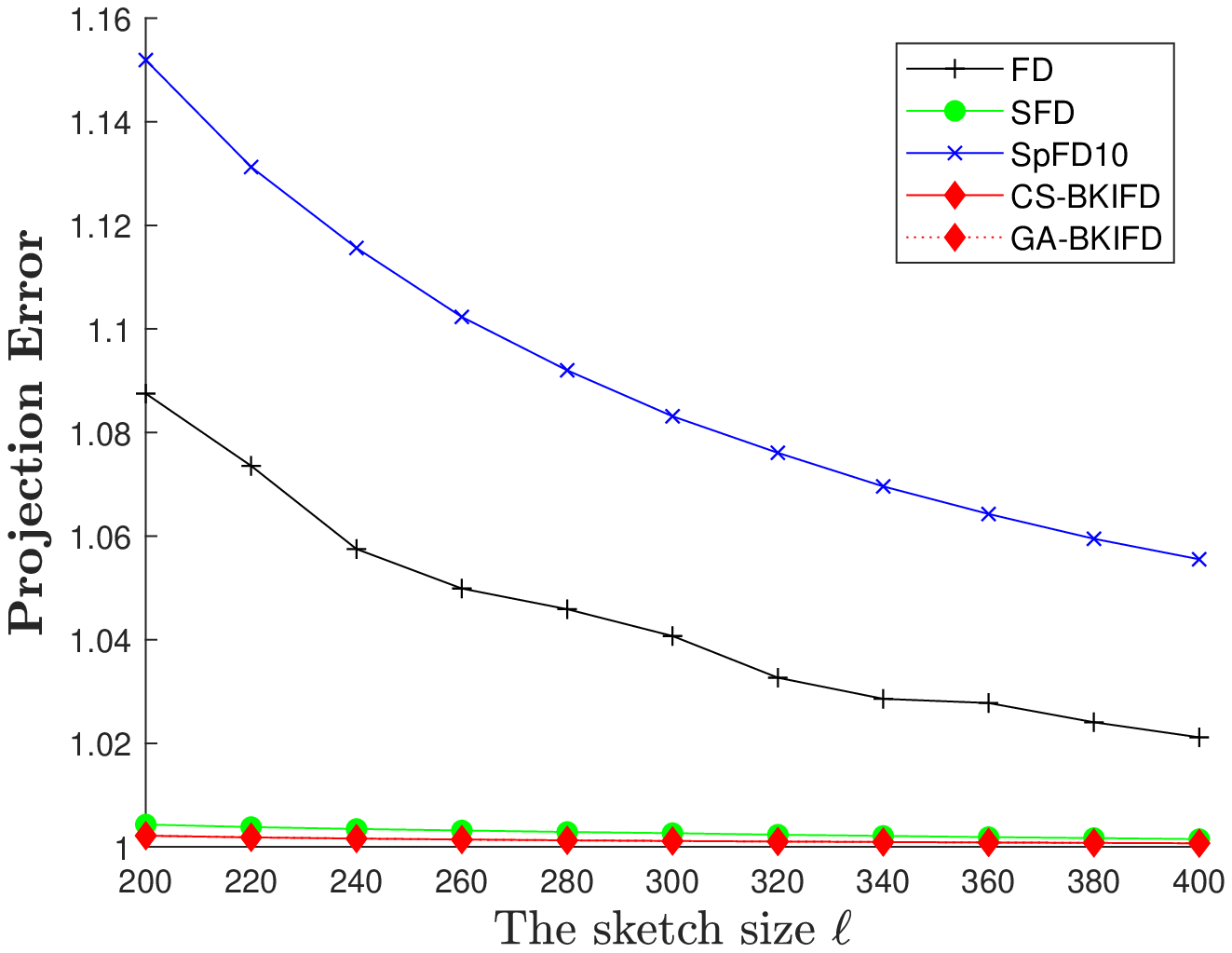}
	}
	\subfigure{
		\includegraphics[width=0.42\columnwidth]{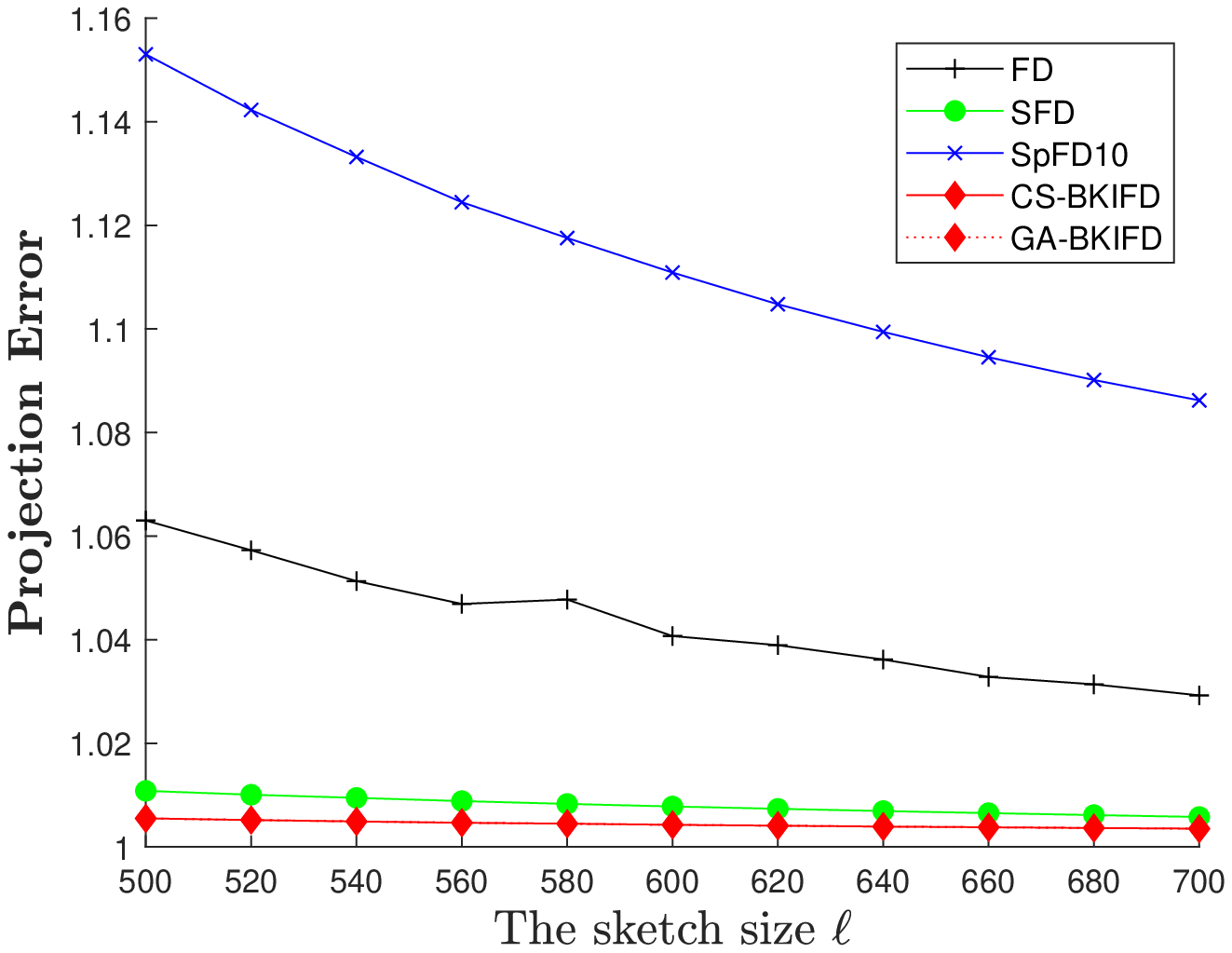}
	}
	\subfigure{
		\includegraphics[width=0.42\columnwidth]{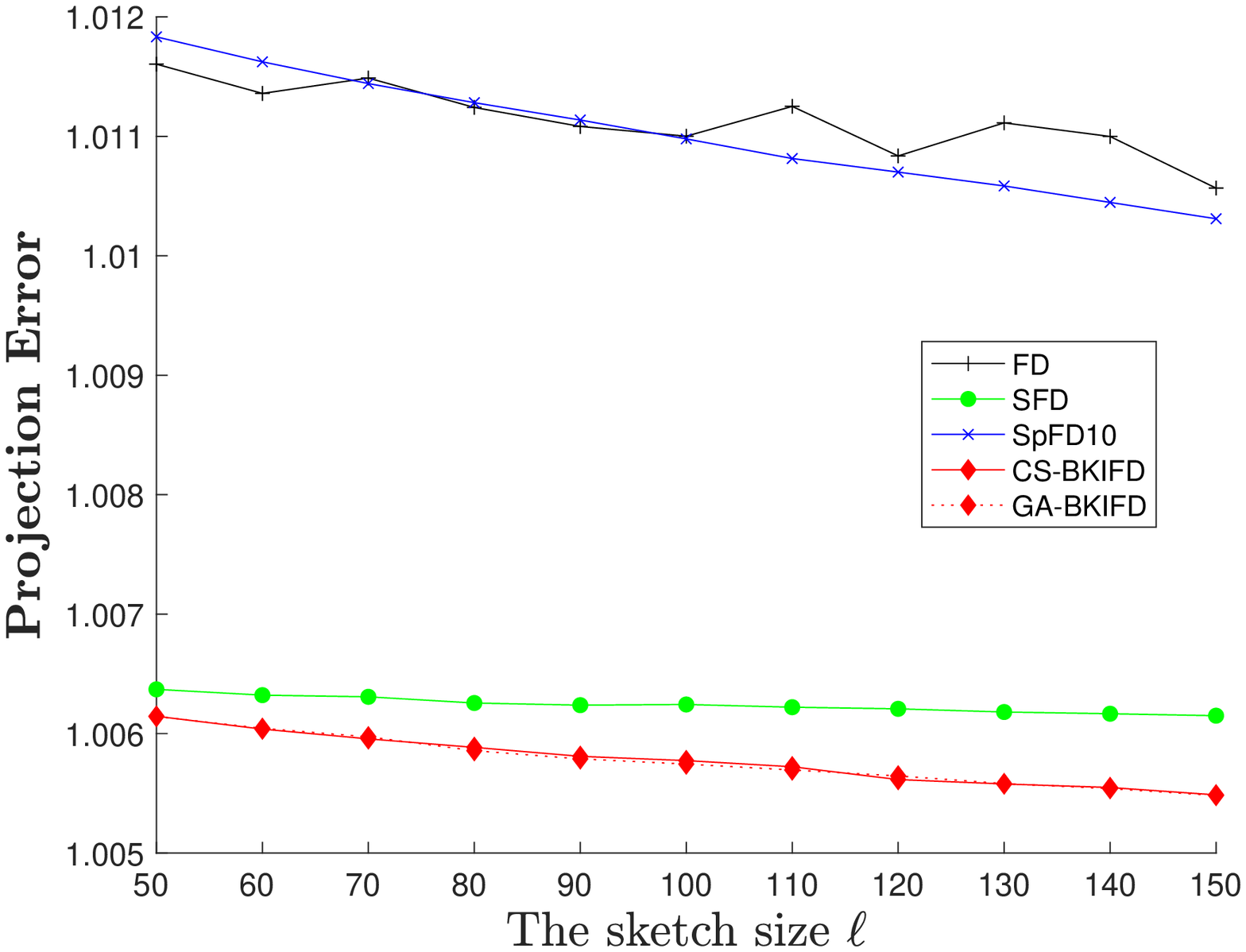}
	}
	\\
	\subfigure[\small{ k = 50 (dense)}]{
		\includegraphics[width=0.42\columnwidth]{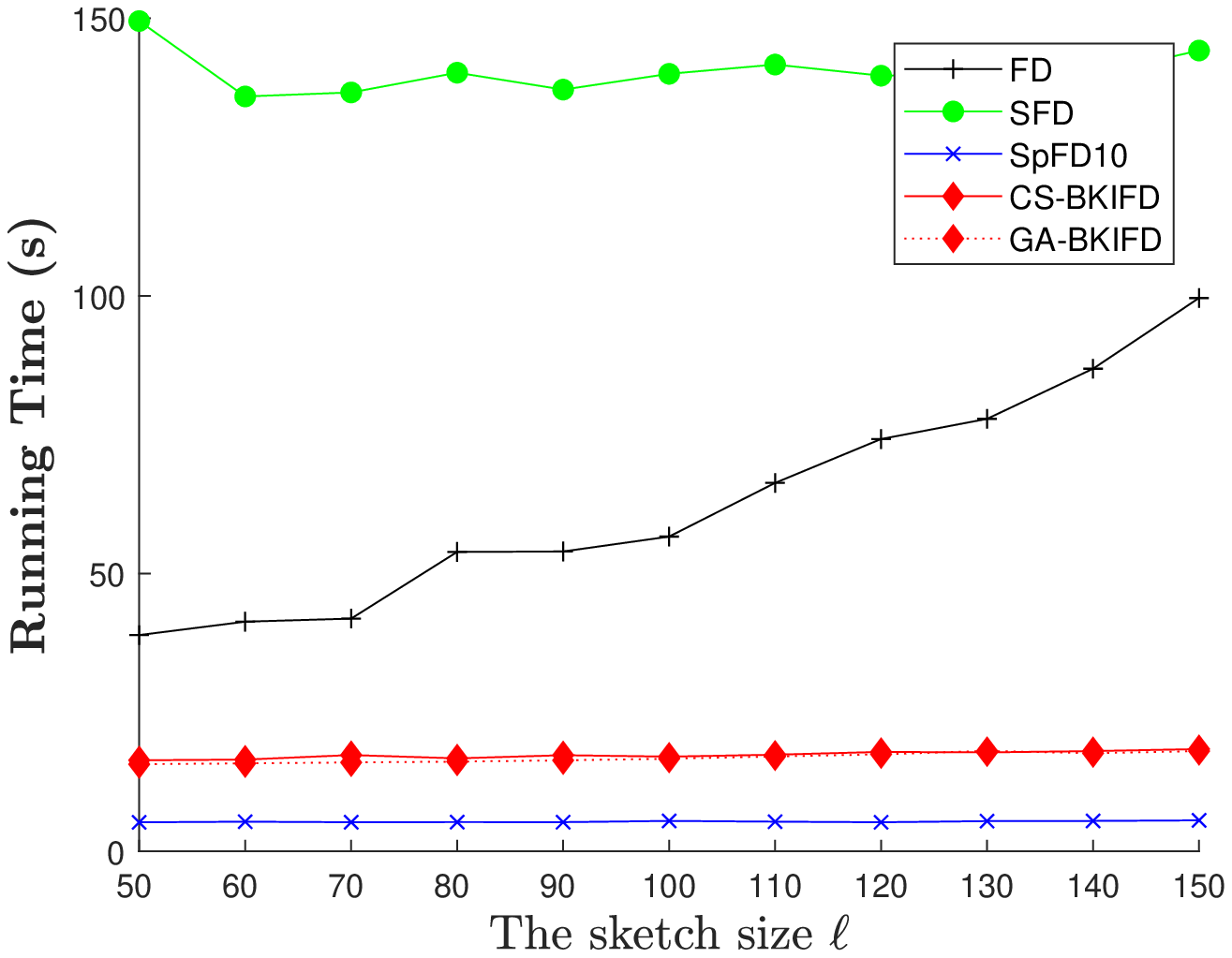}	
	}
	\subfigure[\small{k = 200 (dense)}]{
		\includegraphics[width=0.42\columnwidth]{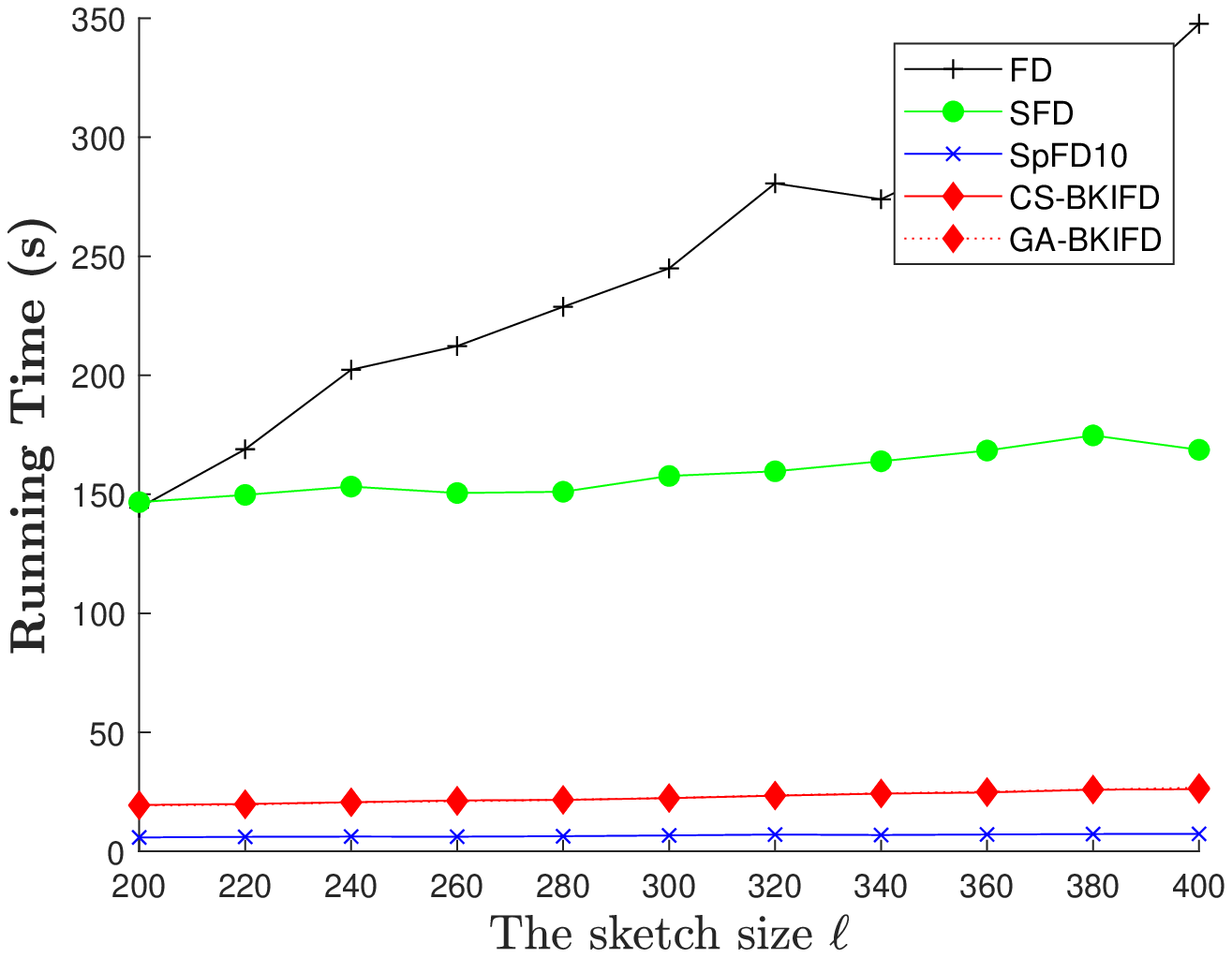}
	}	
	\subfigure[\small{k = 500 (dense)}]{
		\includegraphics[width=0.42\columnwidth]{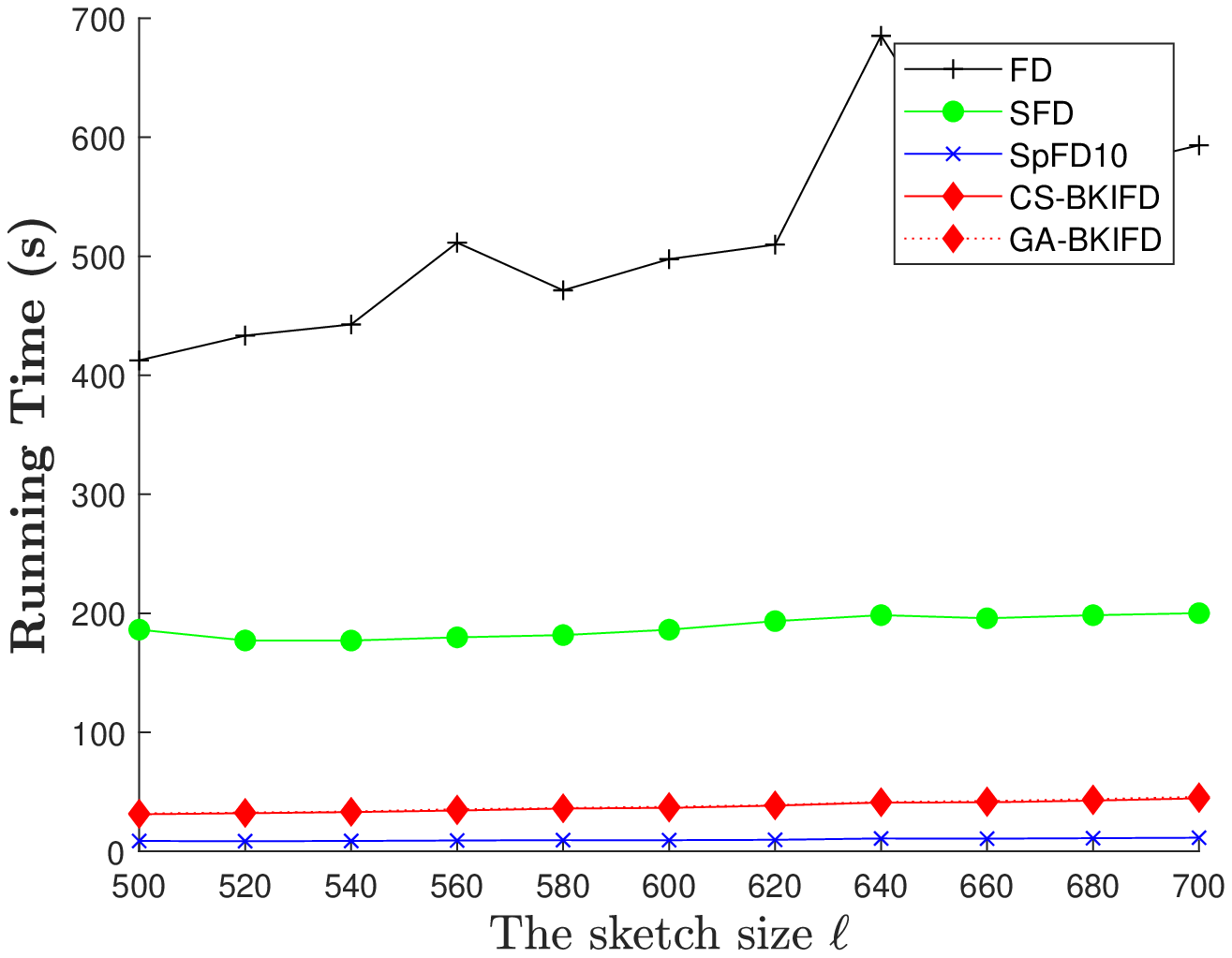}
	}
	\subfigure[\small{k = 50 (sparse)}]{
		\includegraphics[width=0.42\columnwidth]{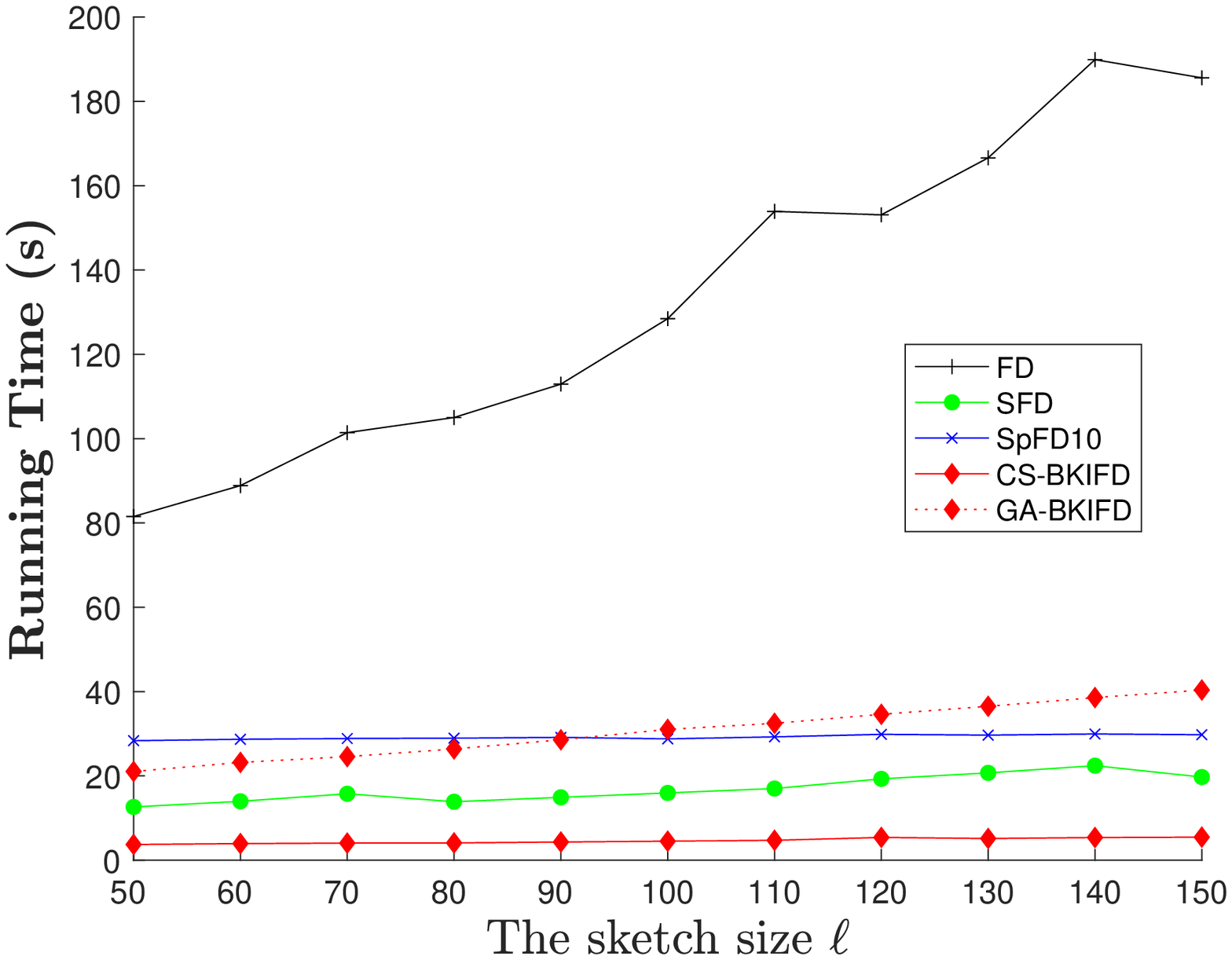}	
	}
	\\
	\centering
	\subfigure{
		\includegraphics[width=0.42\columnwidth]{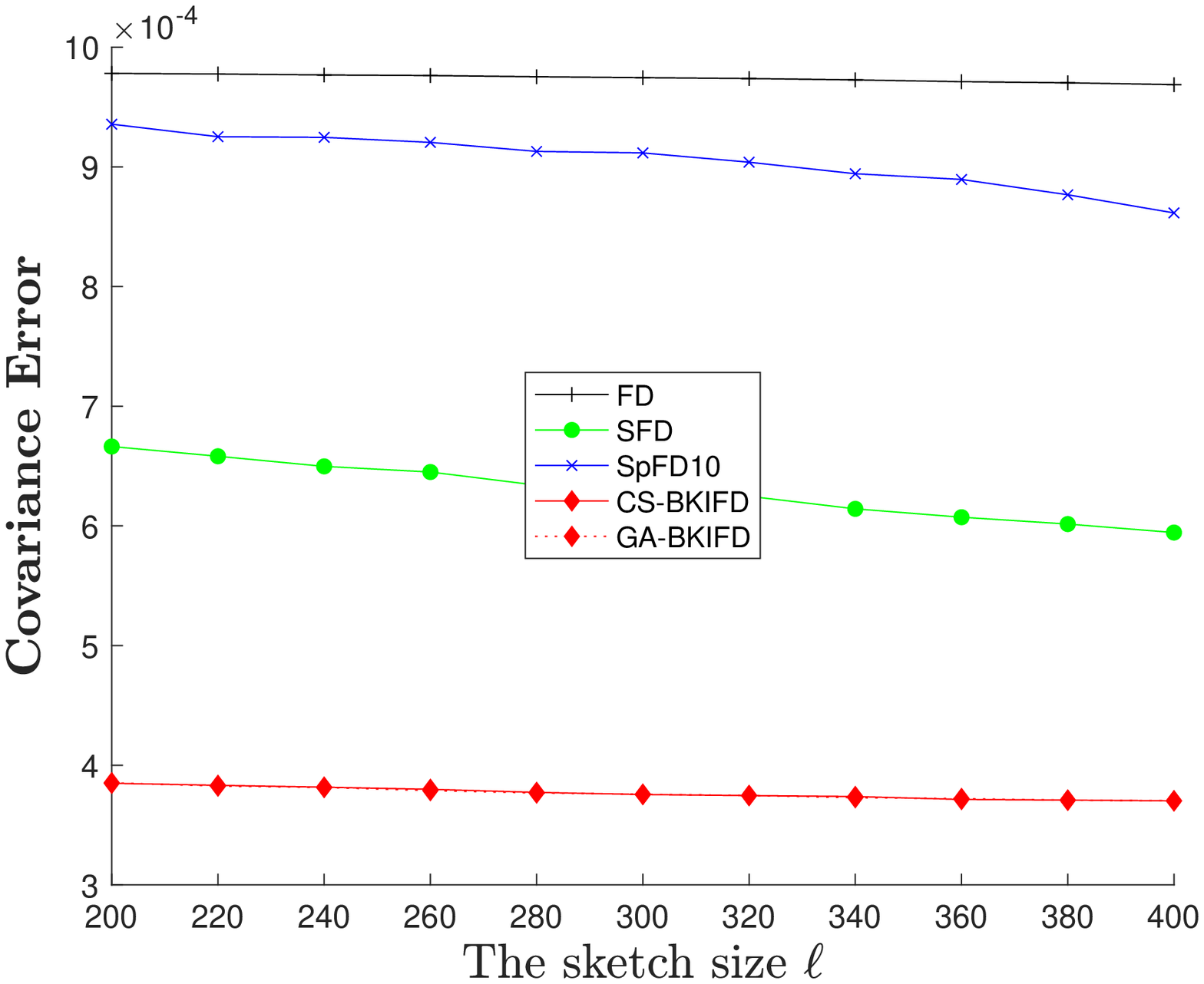}
	}
	\subfigure{
		\includegraphics[width=0.42\columnwidth]{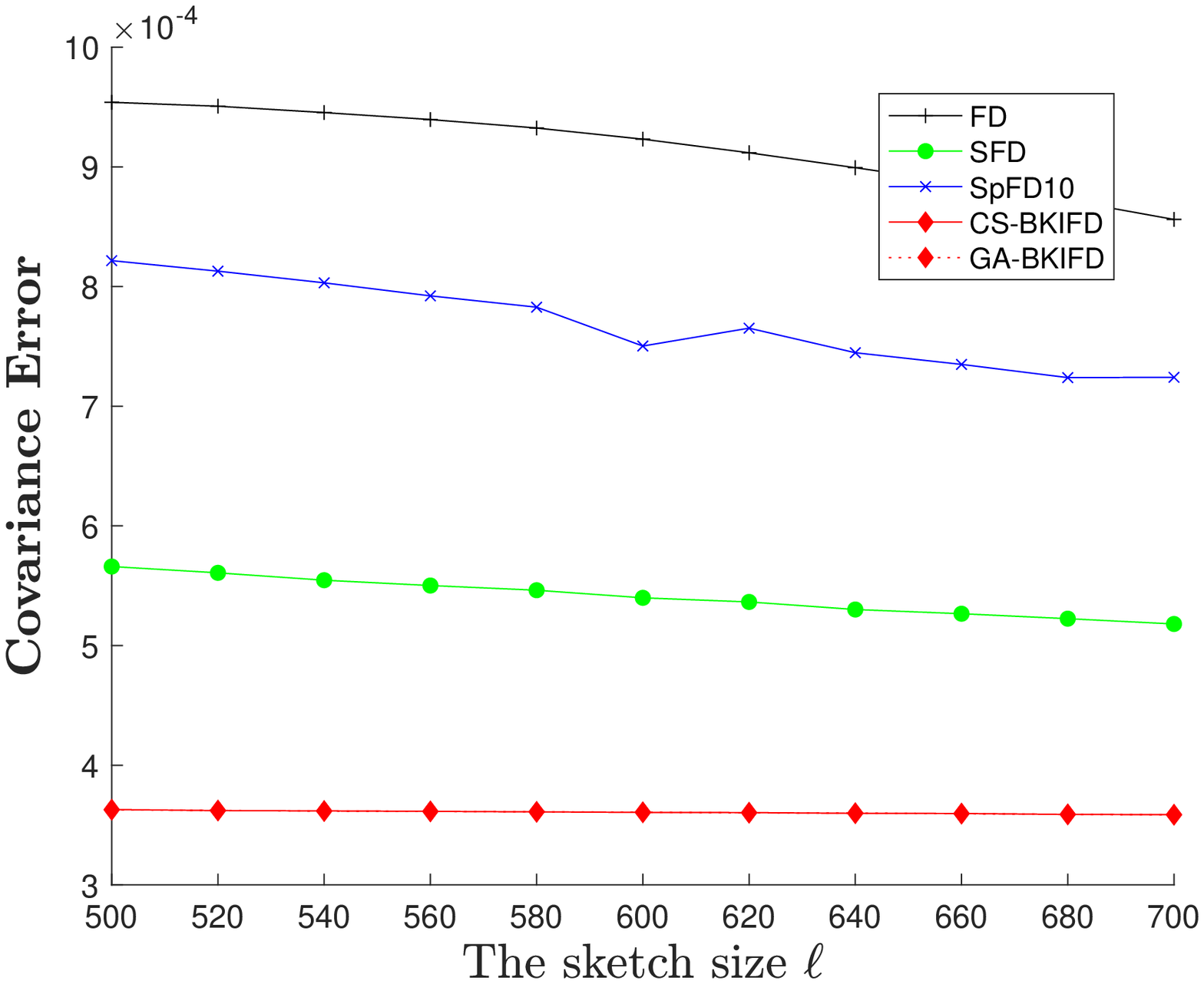}
	}
	\subfigure{
		\includegraphics[width=0.42\columnwidth]{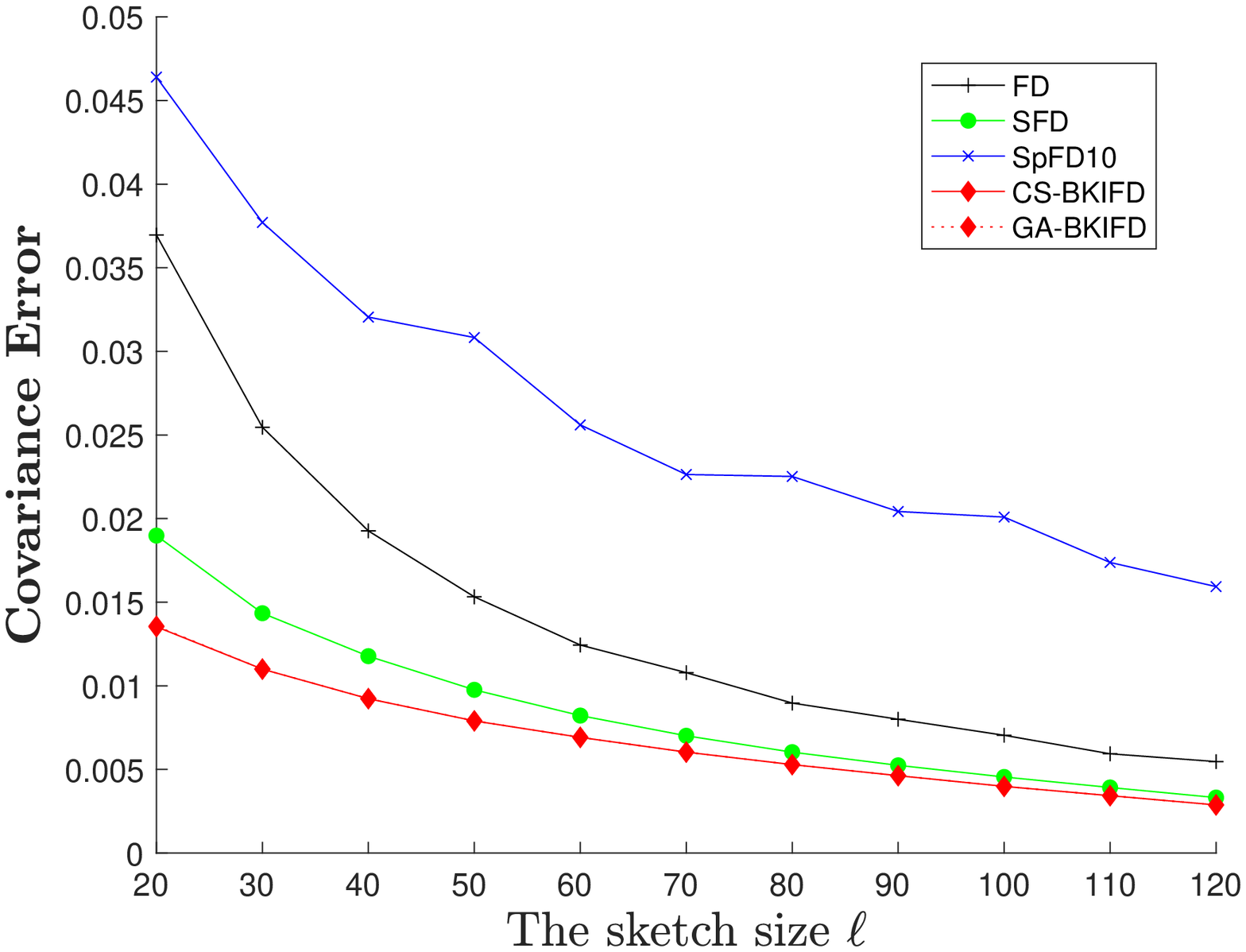}
	}
	\subfigure{
		\includegraphics[width=0.42\columnwidth]{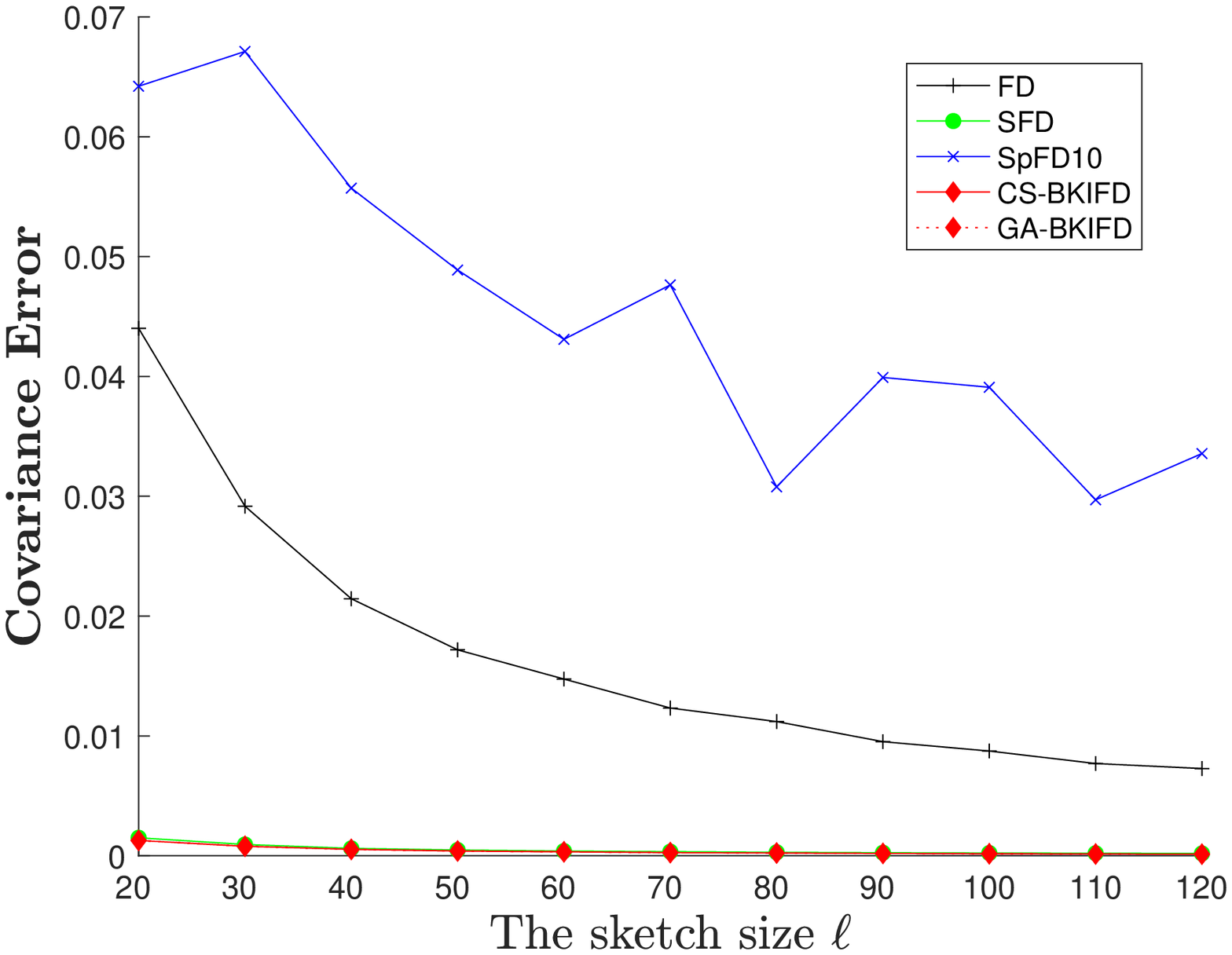}
	}
	\\
	\subfigure{
		\includegraphics[width=0.42\columnwidth]{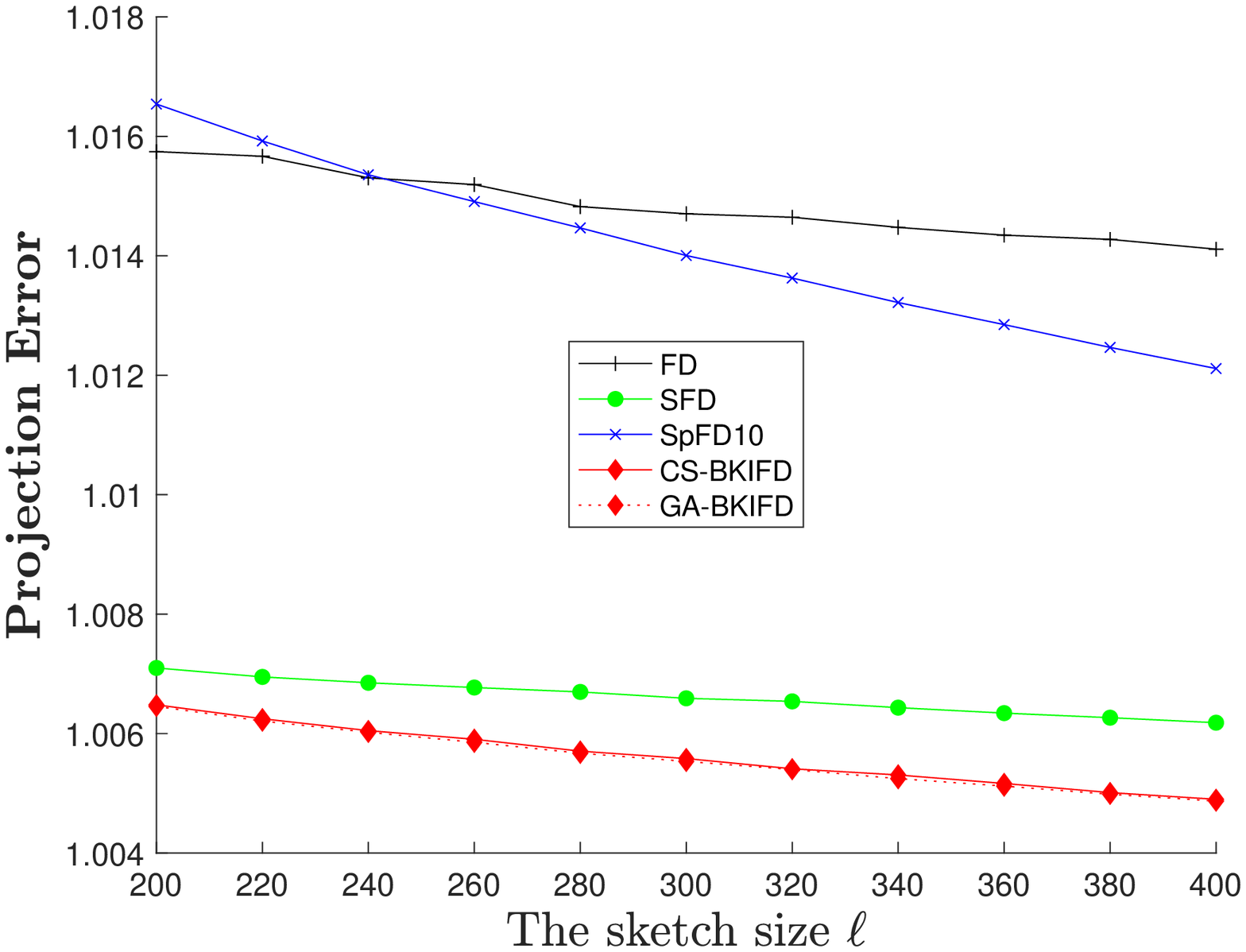}
	}
	\subfigure{
		\includegraphics[width=0.42\columnwidth]{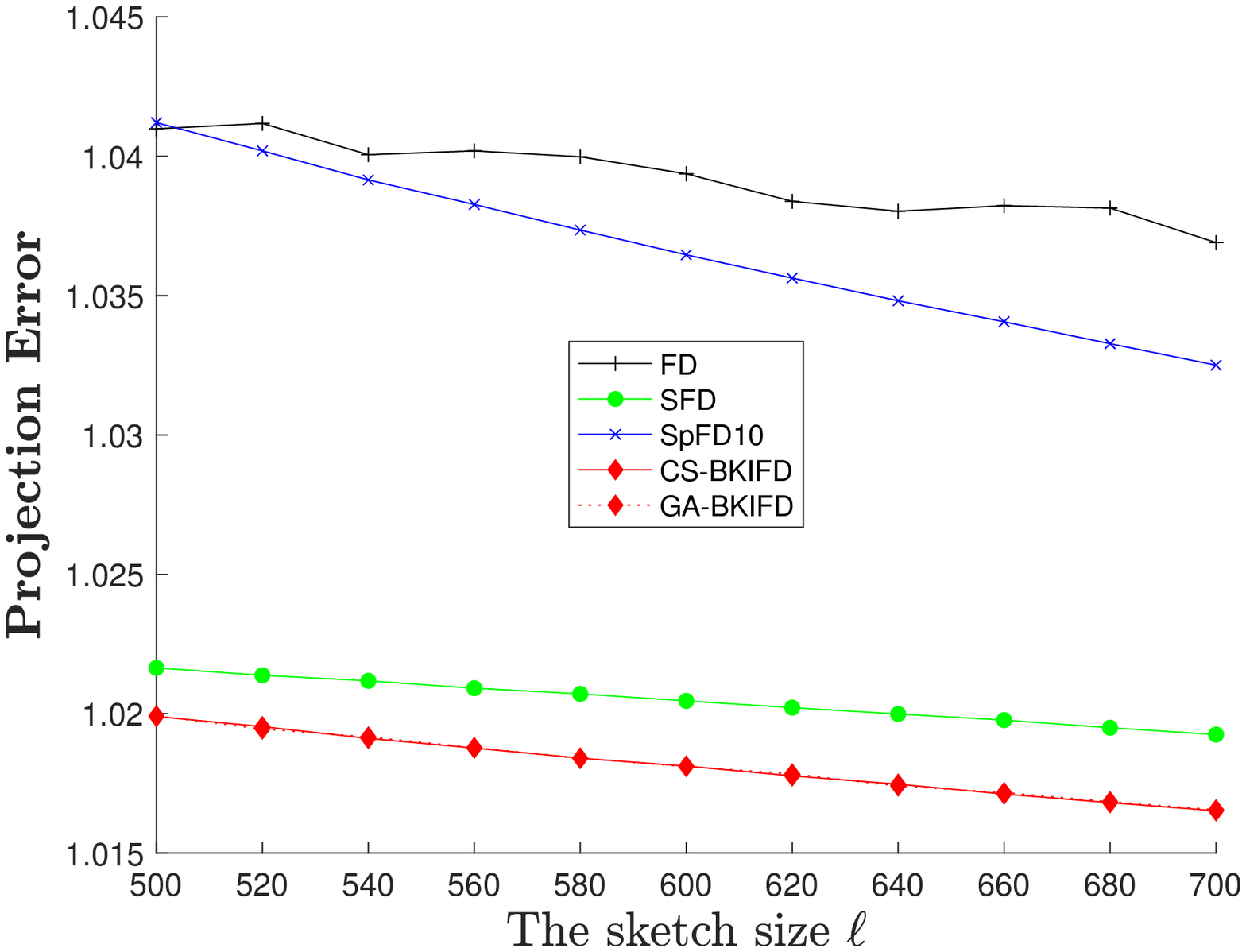}
	}
	\subfigure{
		\includegraphics[width=0.42\columnwidth]{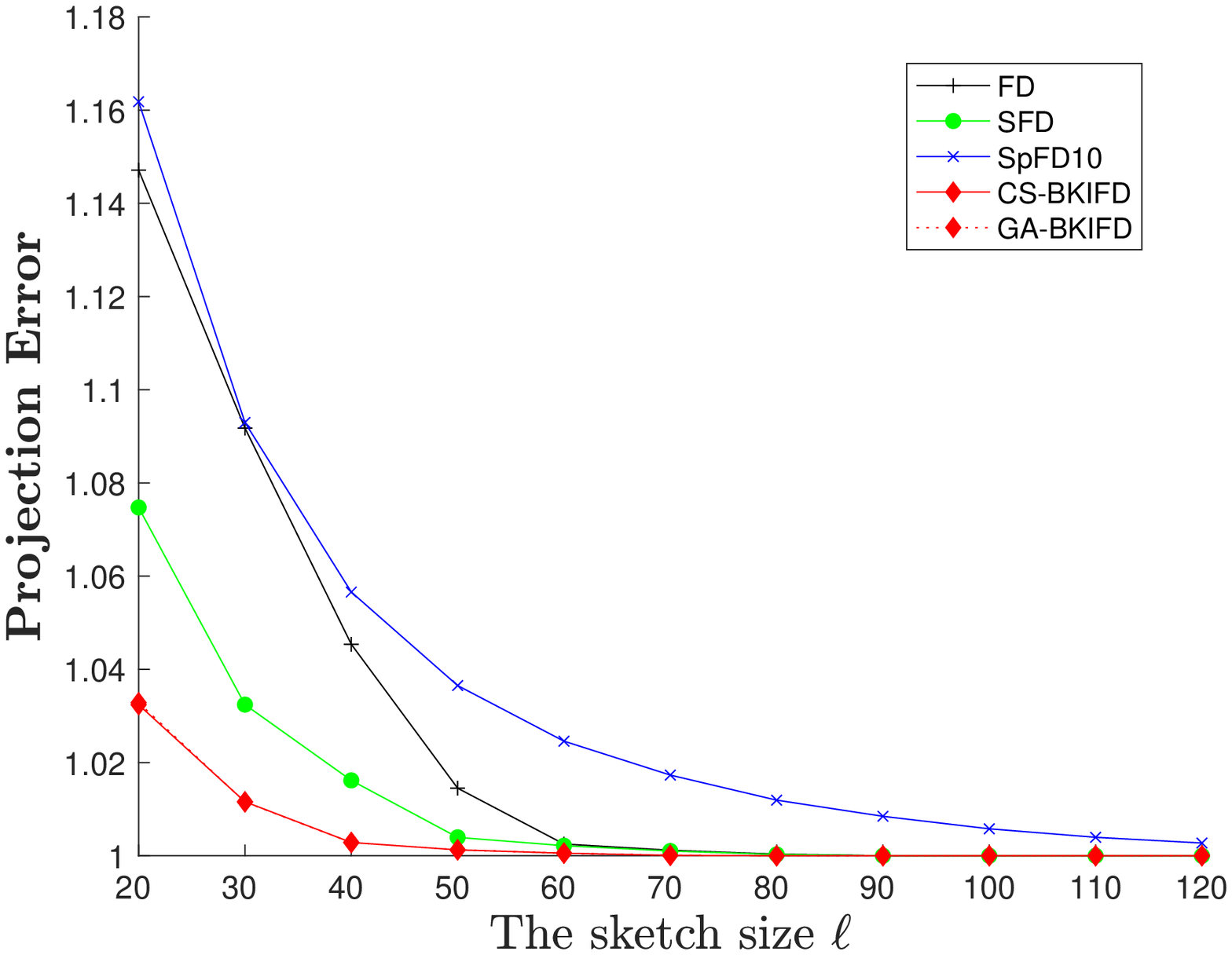}
	}
	\subfigure{
		\includegraphics[width=0.42\columnwidth]{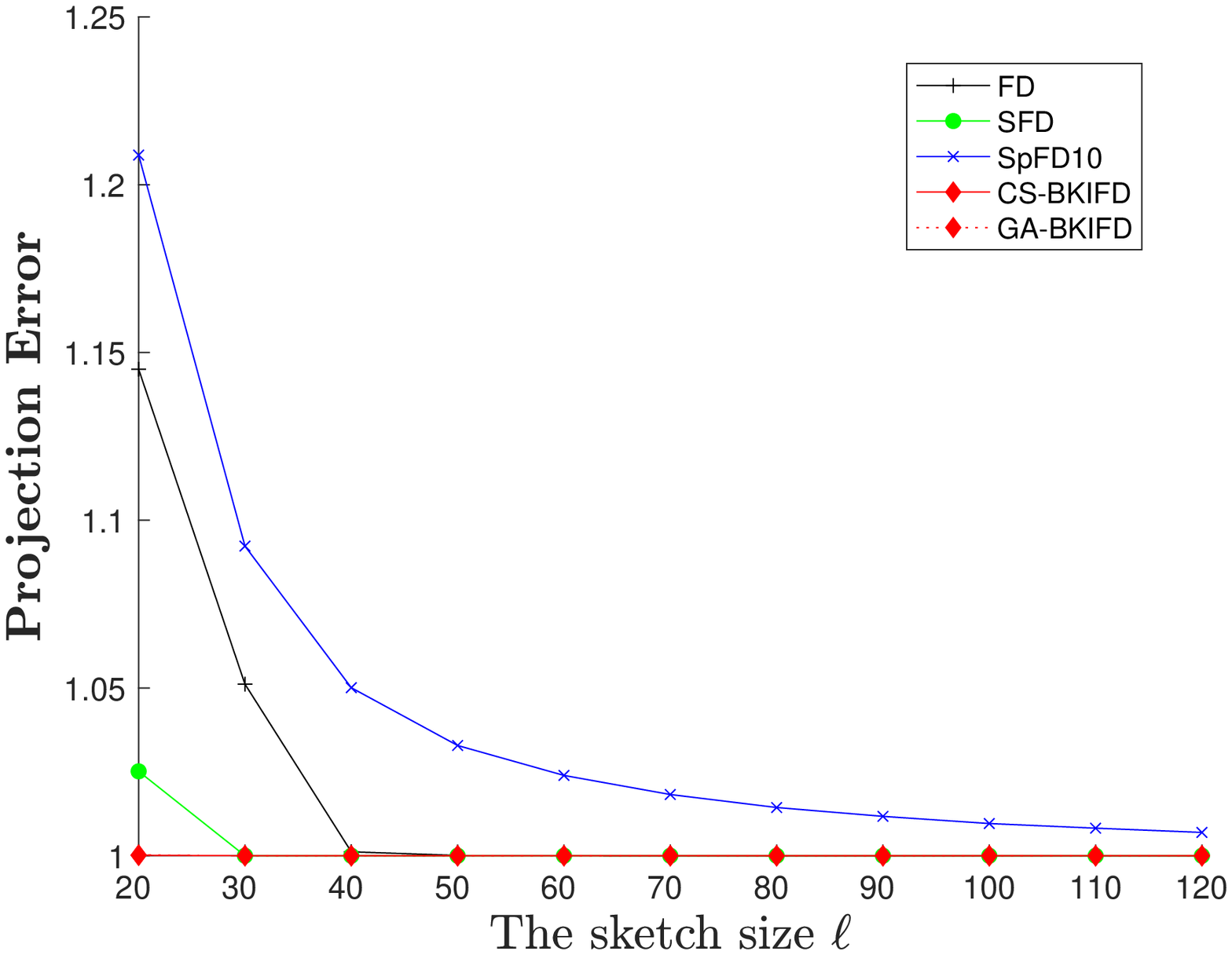}
	}
	\\
	\subfigure[\small{k = 200 (sparse)}]{
		\includegraphics[width=0.42\columnwidth]{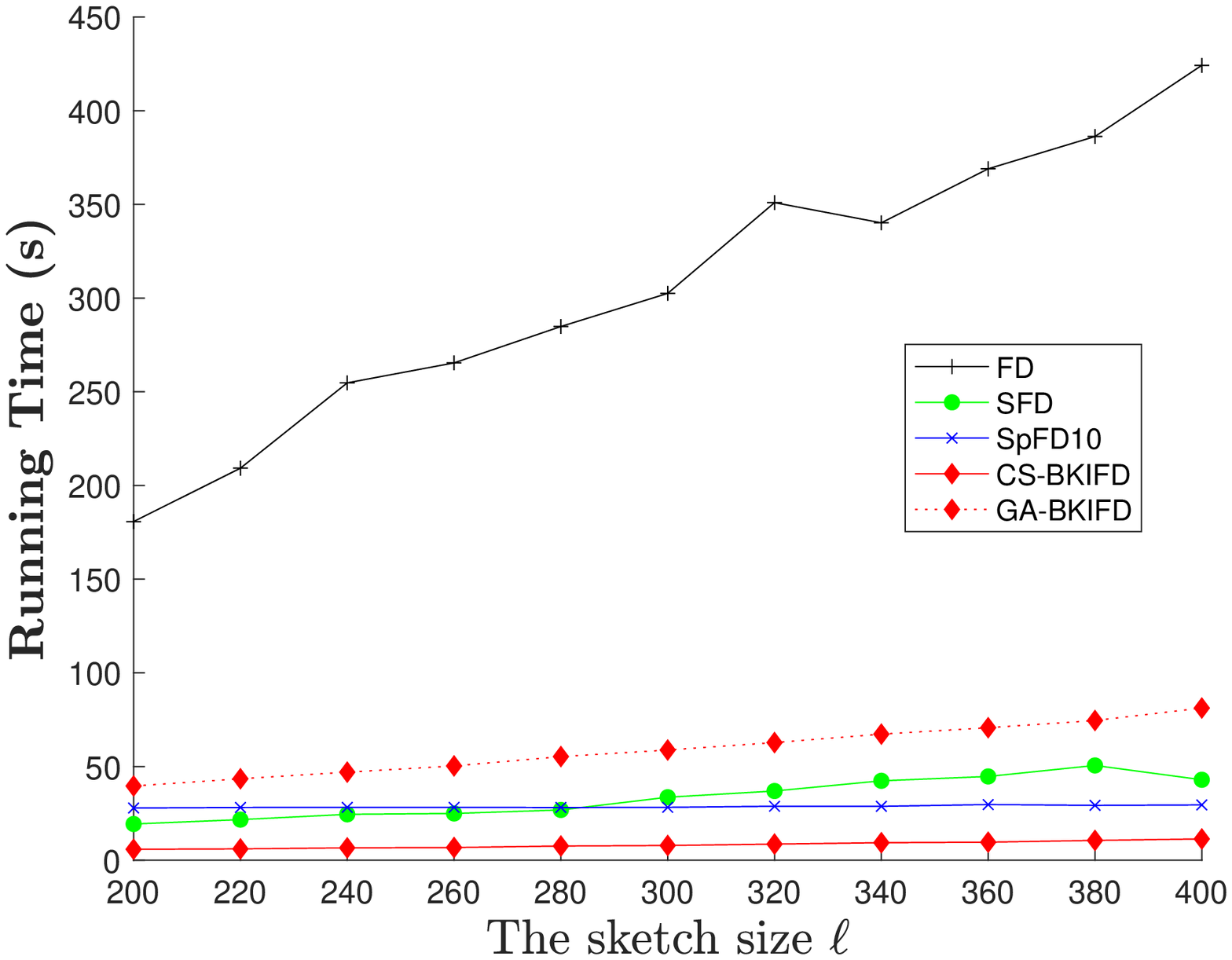}	
	}
	\subfigure[\small{k = 500 (sparse)}]{
		\includegraphics[width=0.42\columnwidth]{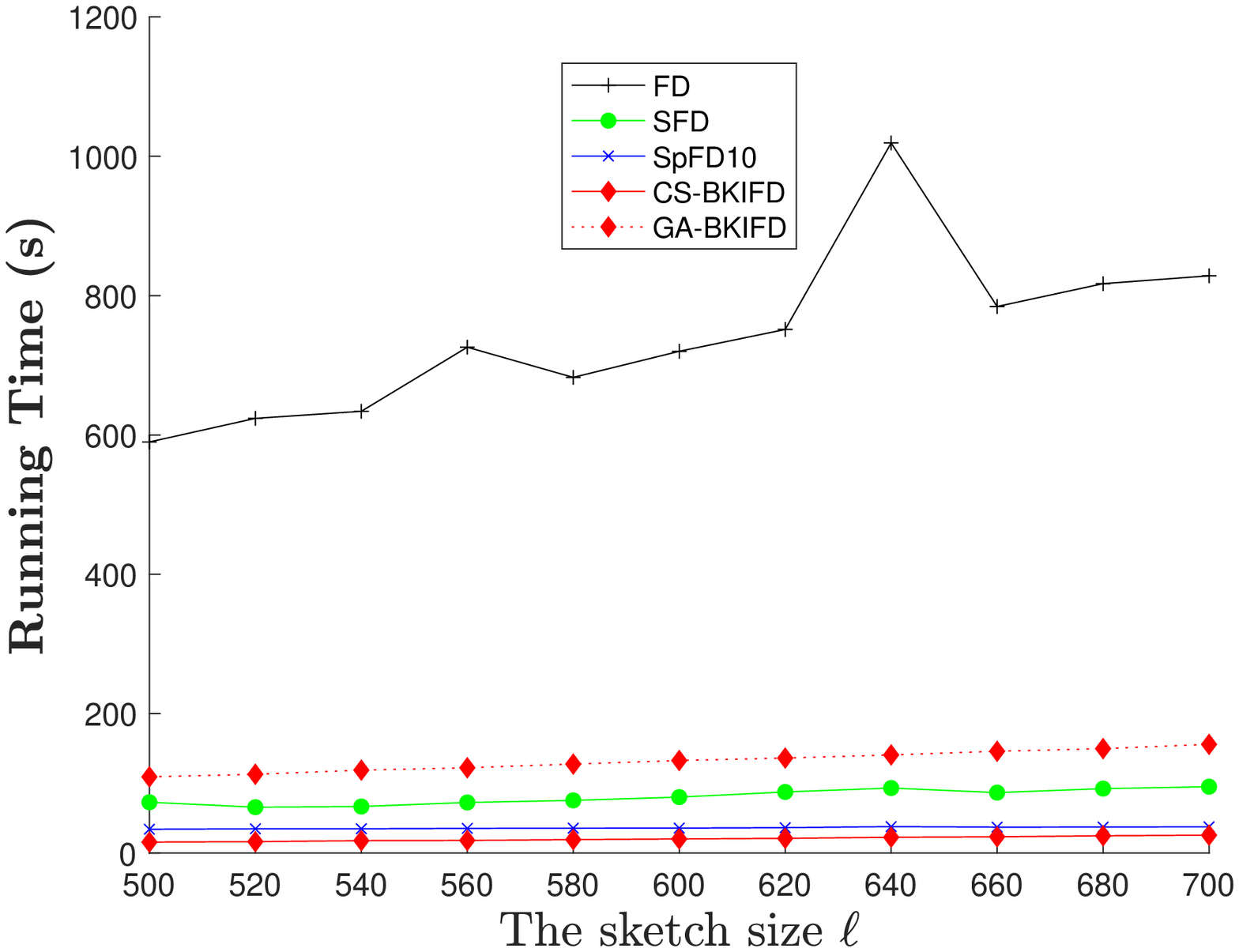}
	}	
	\subfigure[\small{w8a}]{
		\includegraphics[width=0.42\columnwidth]{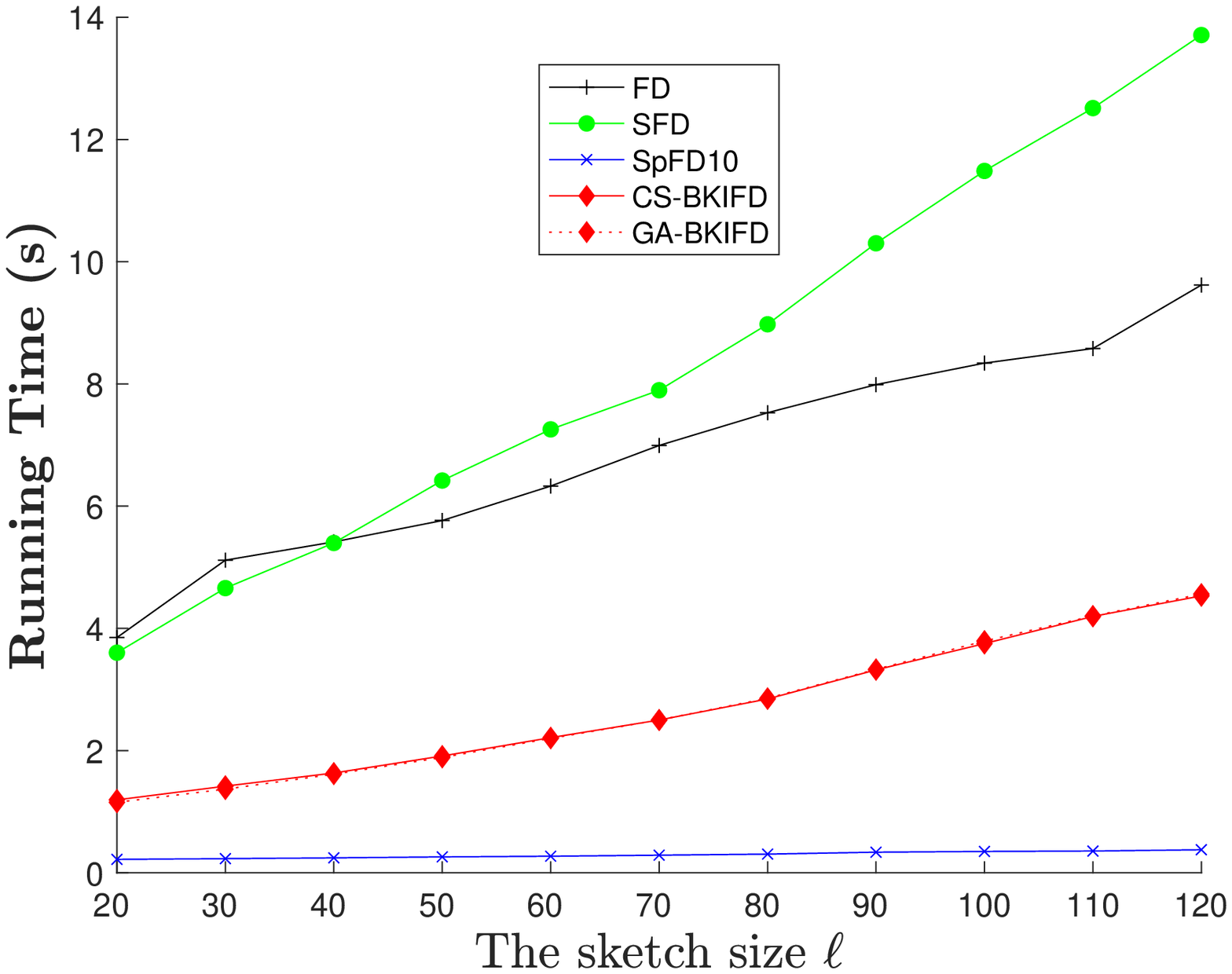}
	}
	\subfigure[\small{CIFAR-10}]{
		\includegraphics[width=0.42\columnwidth]{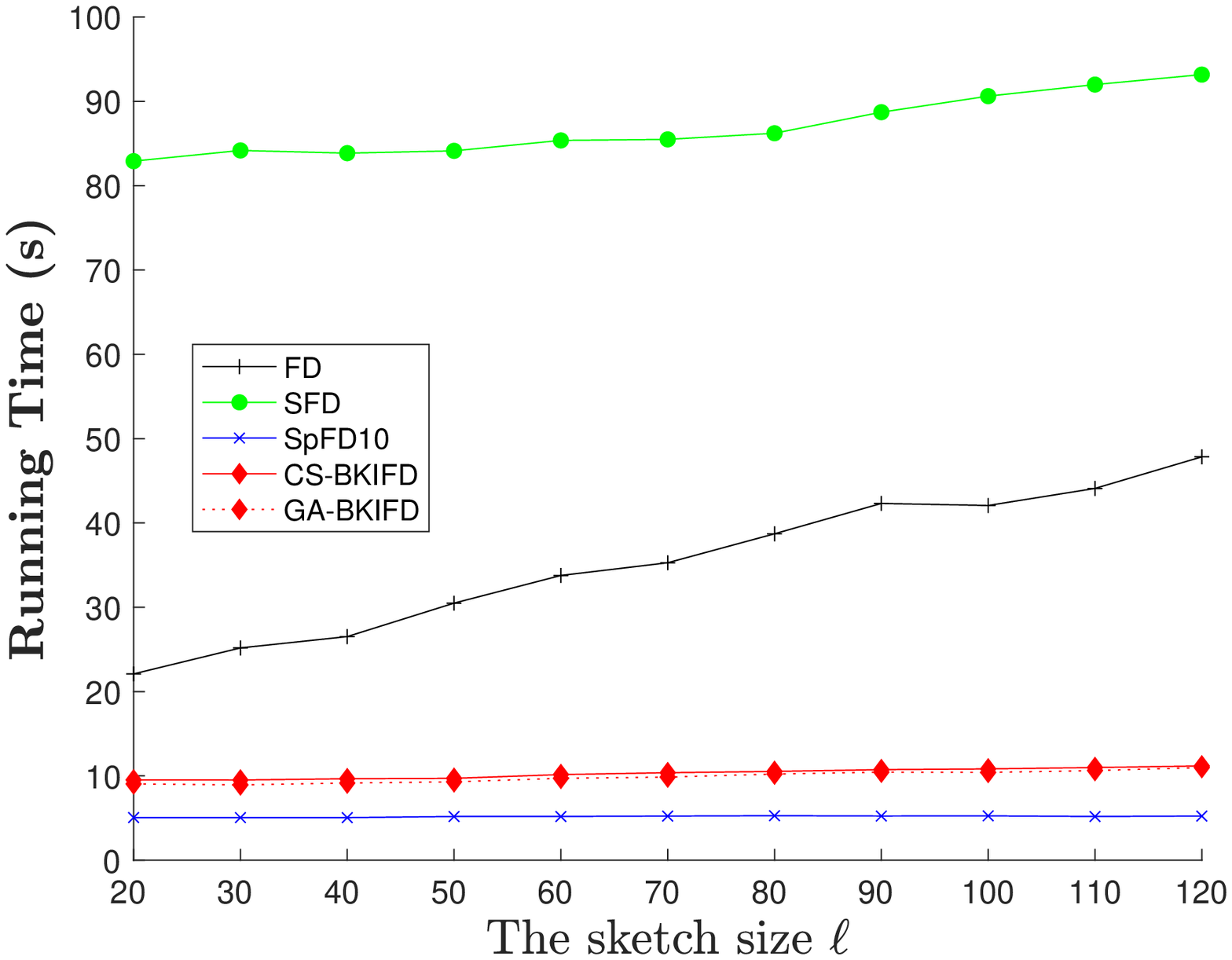}	
	}
	\caption{Comparison results on synthetic datasets and real datasets: w8a and CIFAR-10}
	\label{fig:exp}
\end{figure*}

\begin{figure*}[htbp]
	\centering
	\subfigure{
		\includegraphics[width=0.42\columnwidth]{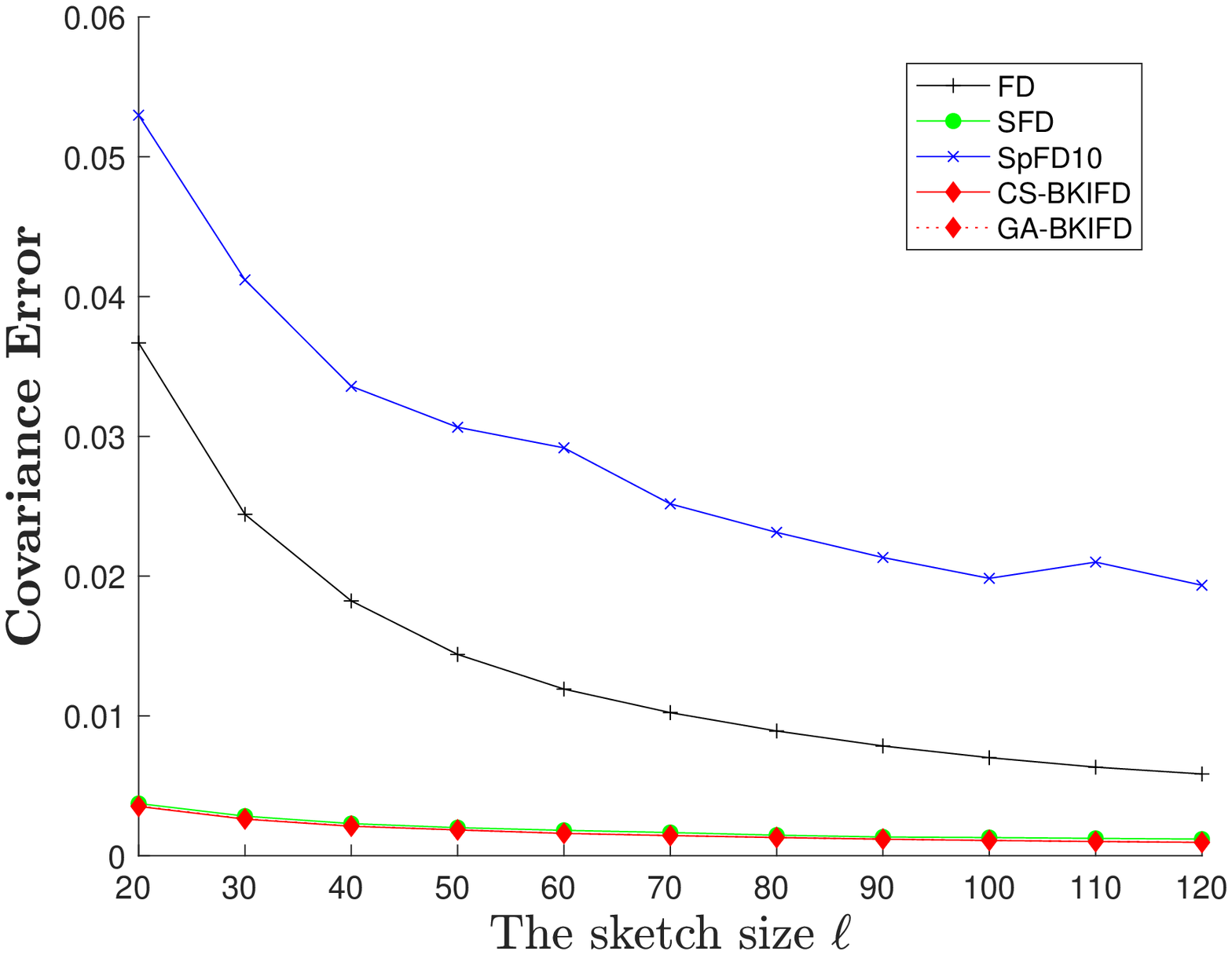}
	}
	\subfigure{
		\includegraphics[width=0.42\columnwidth]{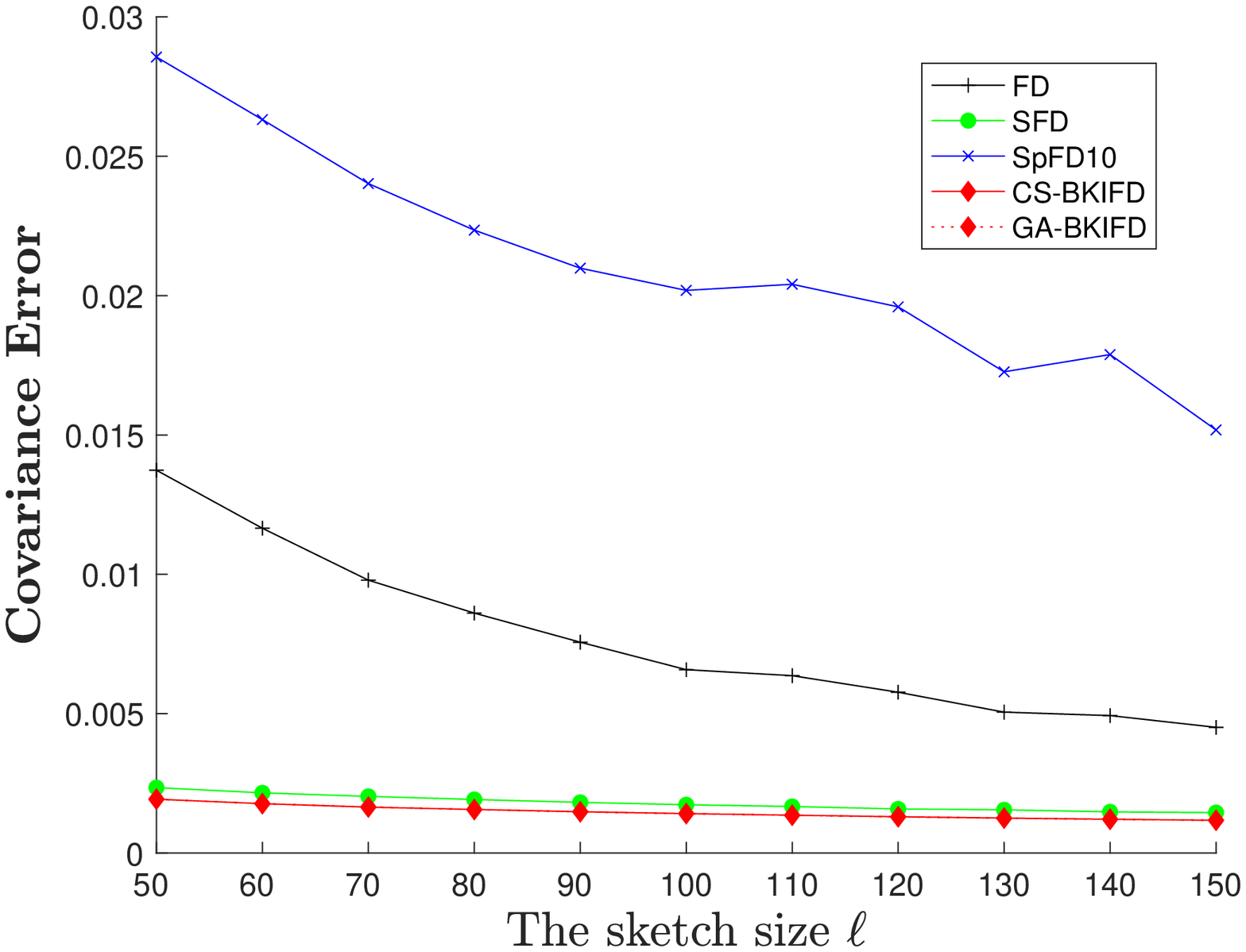}
	}
	\subfigure{
		\includegraphics[width=0.42\columnwidth]{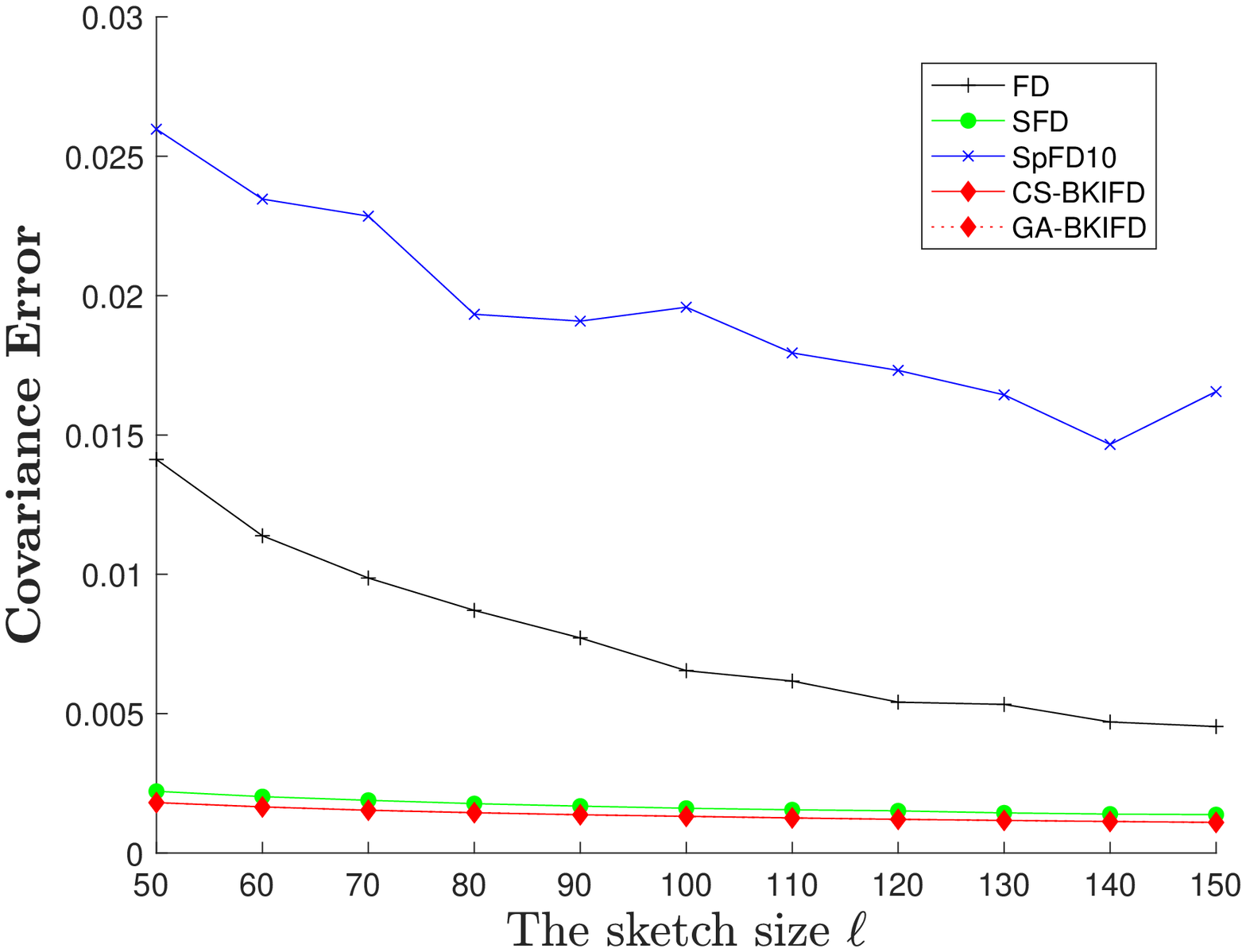}
	}
	\subfigure{
		\includegraphics[width=0.42\columnwidth]{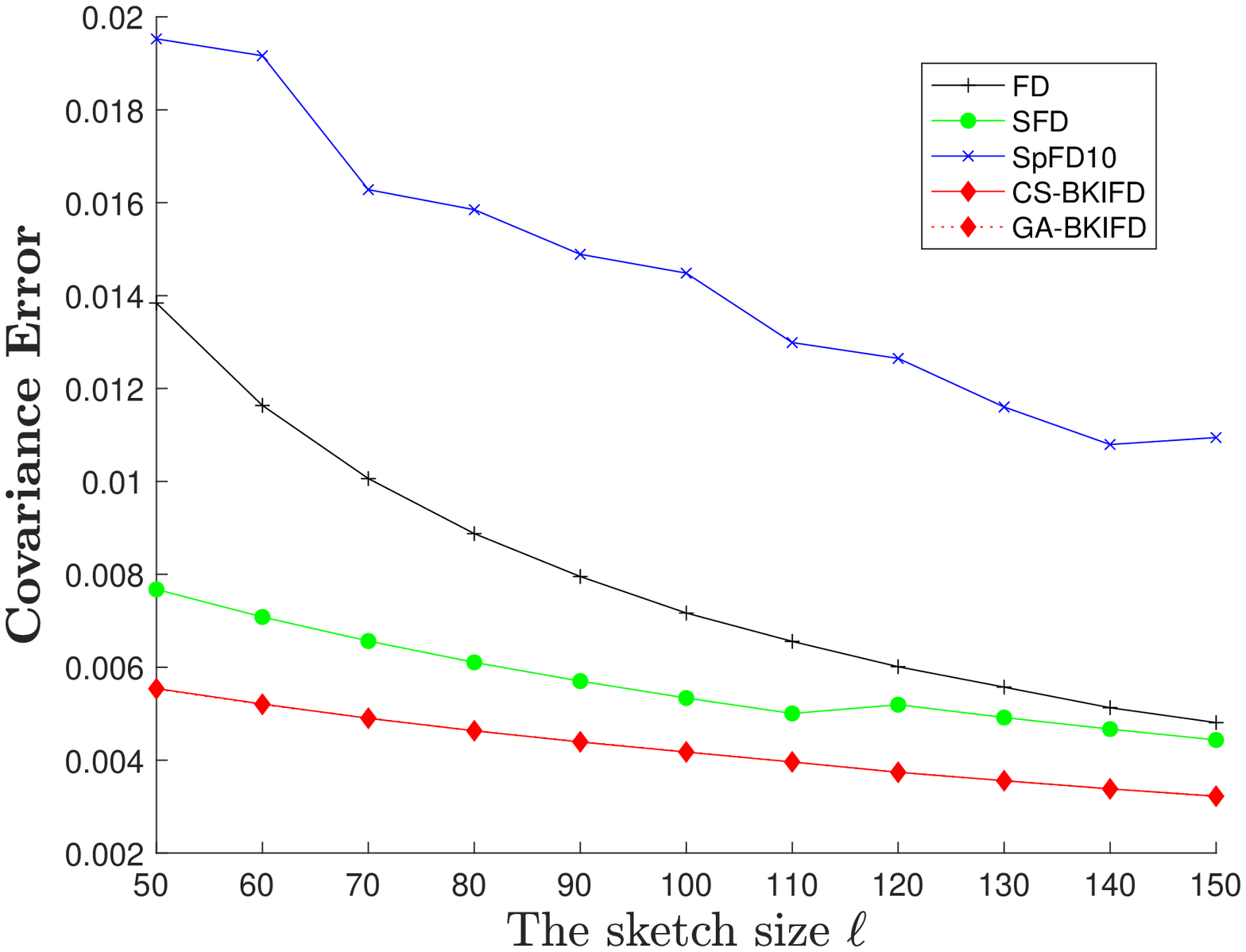}
	}
	\\
	\subfigure{
		\includegraphics[width=0.42\columnwidth]{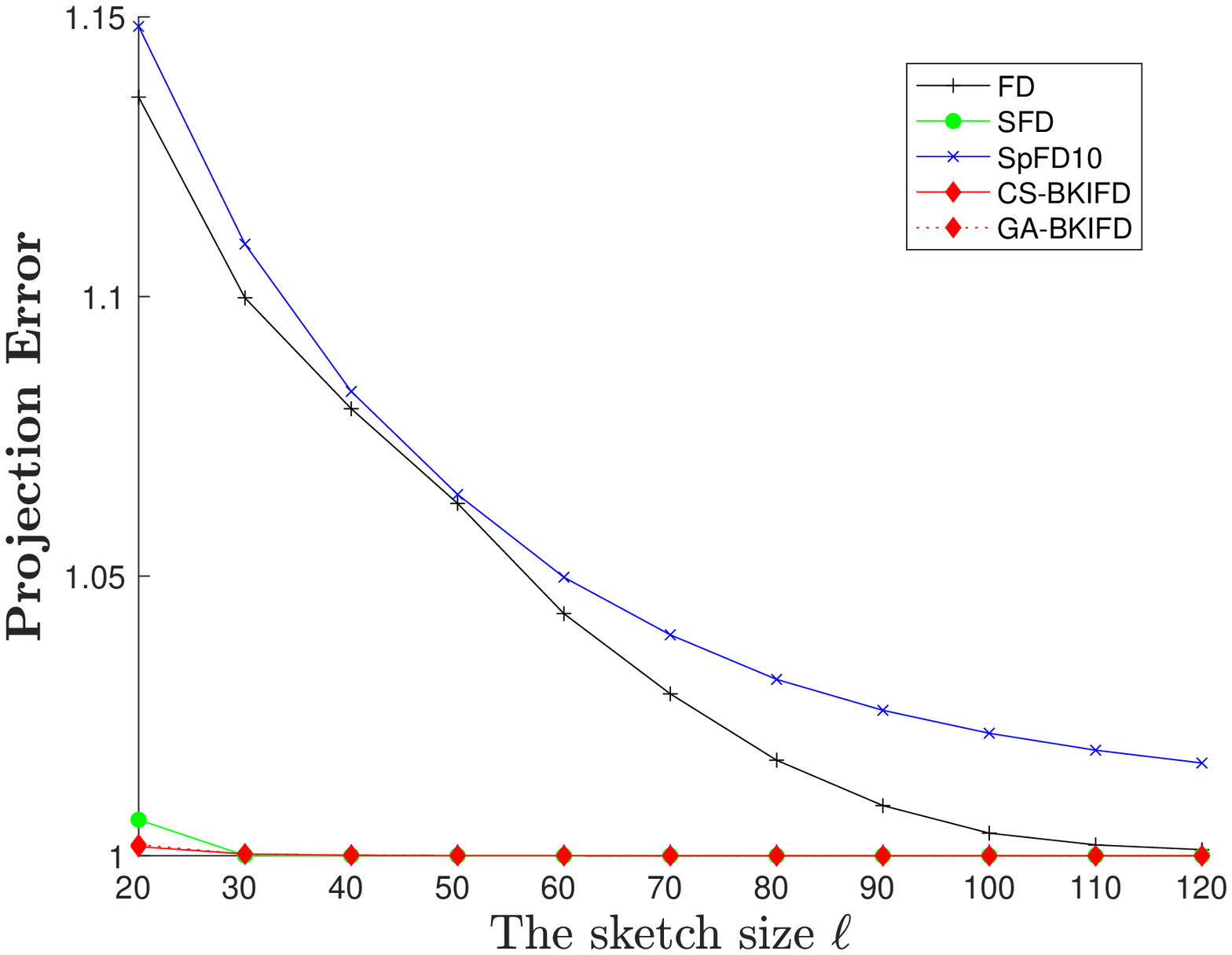}
	}
	\subfigure{
		\includegraphics[width=0.42\columnwidth]{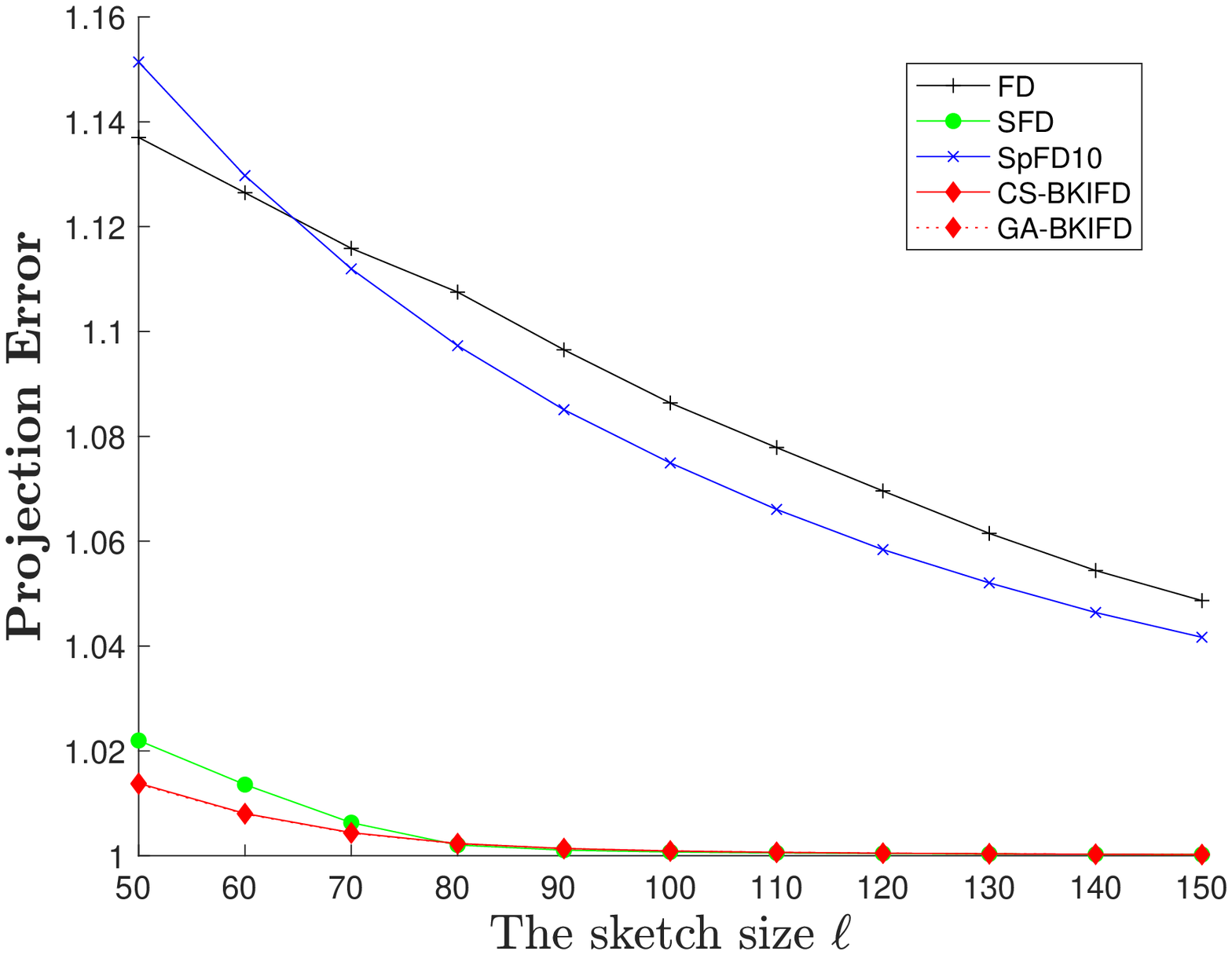}
	}
	\subfigure{
		\includegraphics[width=0.42\columnwidth]{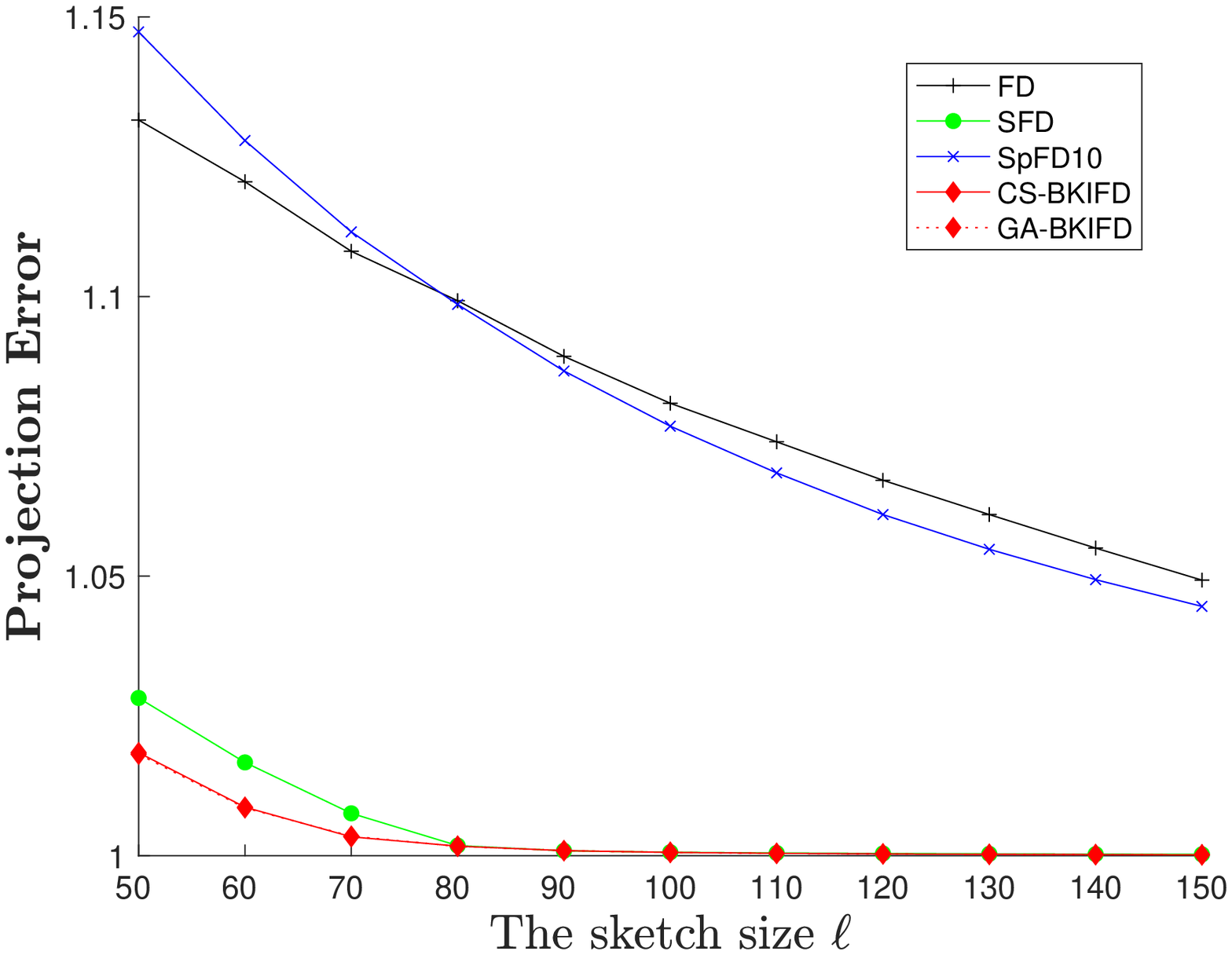}
	}
	\subfigure{
		\includegraphics[width=0.42\columnwidth]{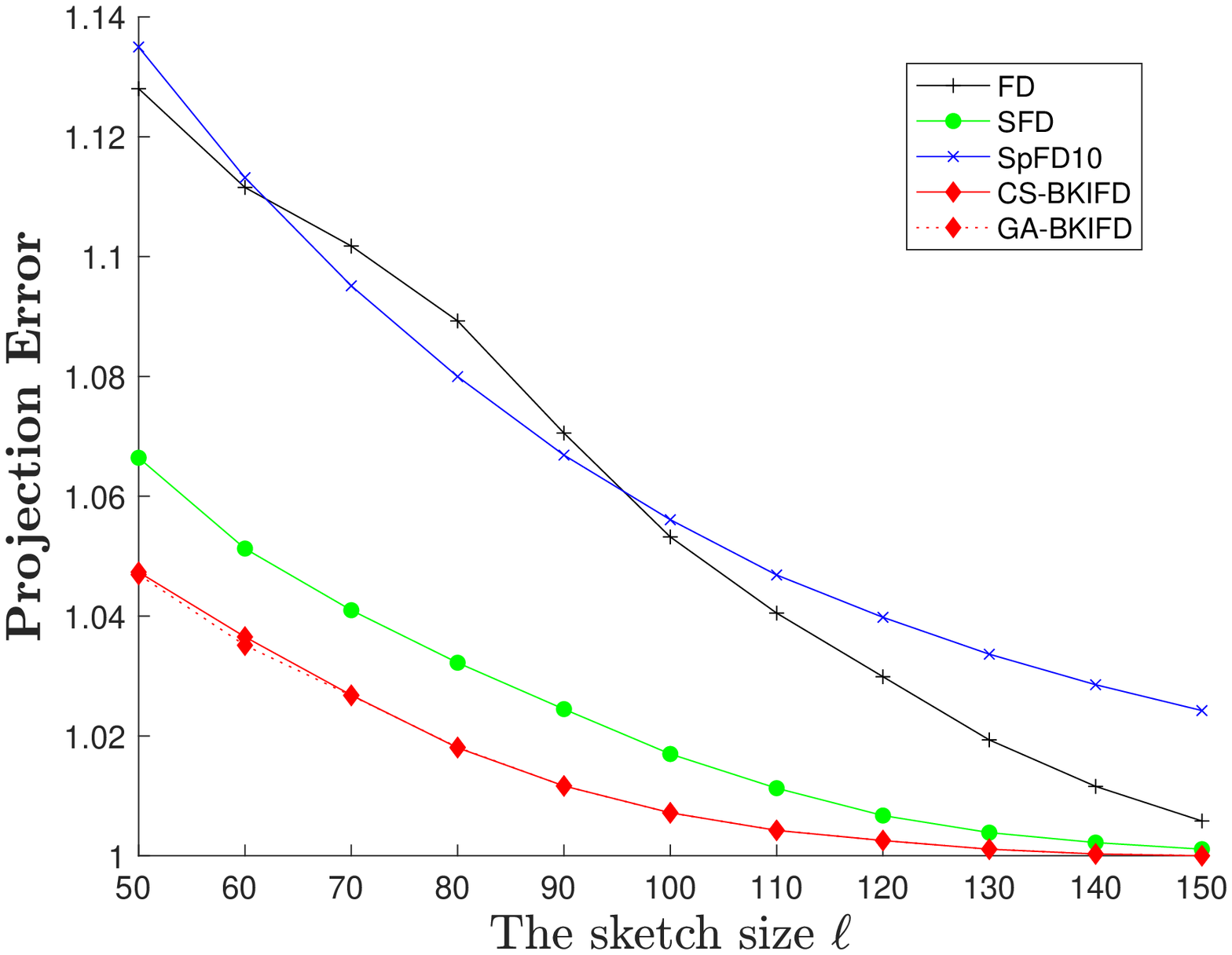}
	}
	\\
	\subfigure[\small{Sido0}]{
		\includegraphics[width=0.42\columnwidth]{time_42.eps}
	}	
	\subfigure[\small{MovieLens-10M}]{
		\includegraphics[width=0.42\columnwidth]{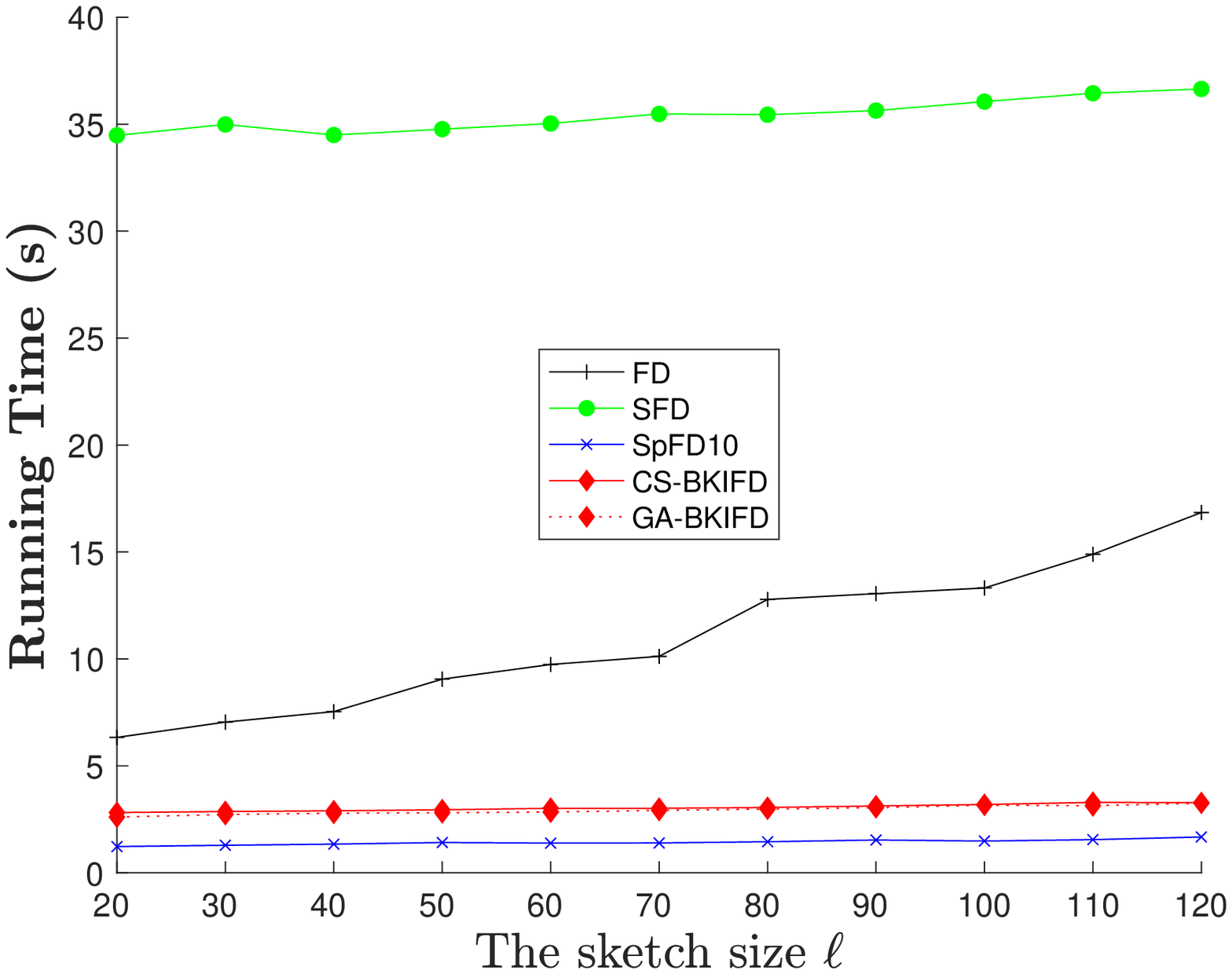}
	}
	\subfigure[\small{MovieLens-20M}]{
		\includegraphics[width=0.42\columnwidth]{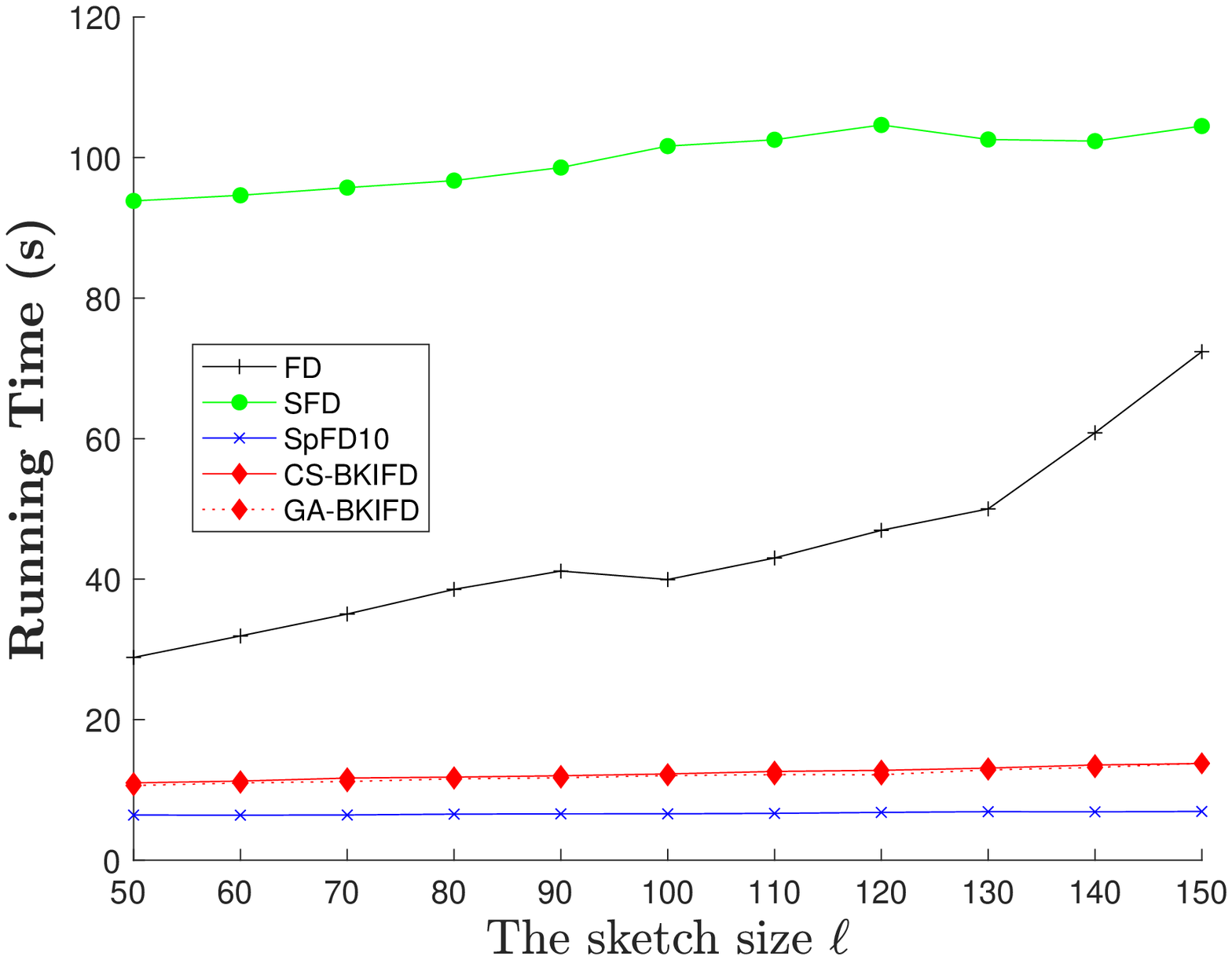}	
	}
	\subfigure[\small{Protein}]{
		\includegraphics[width=0.42\columnwidth]{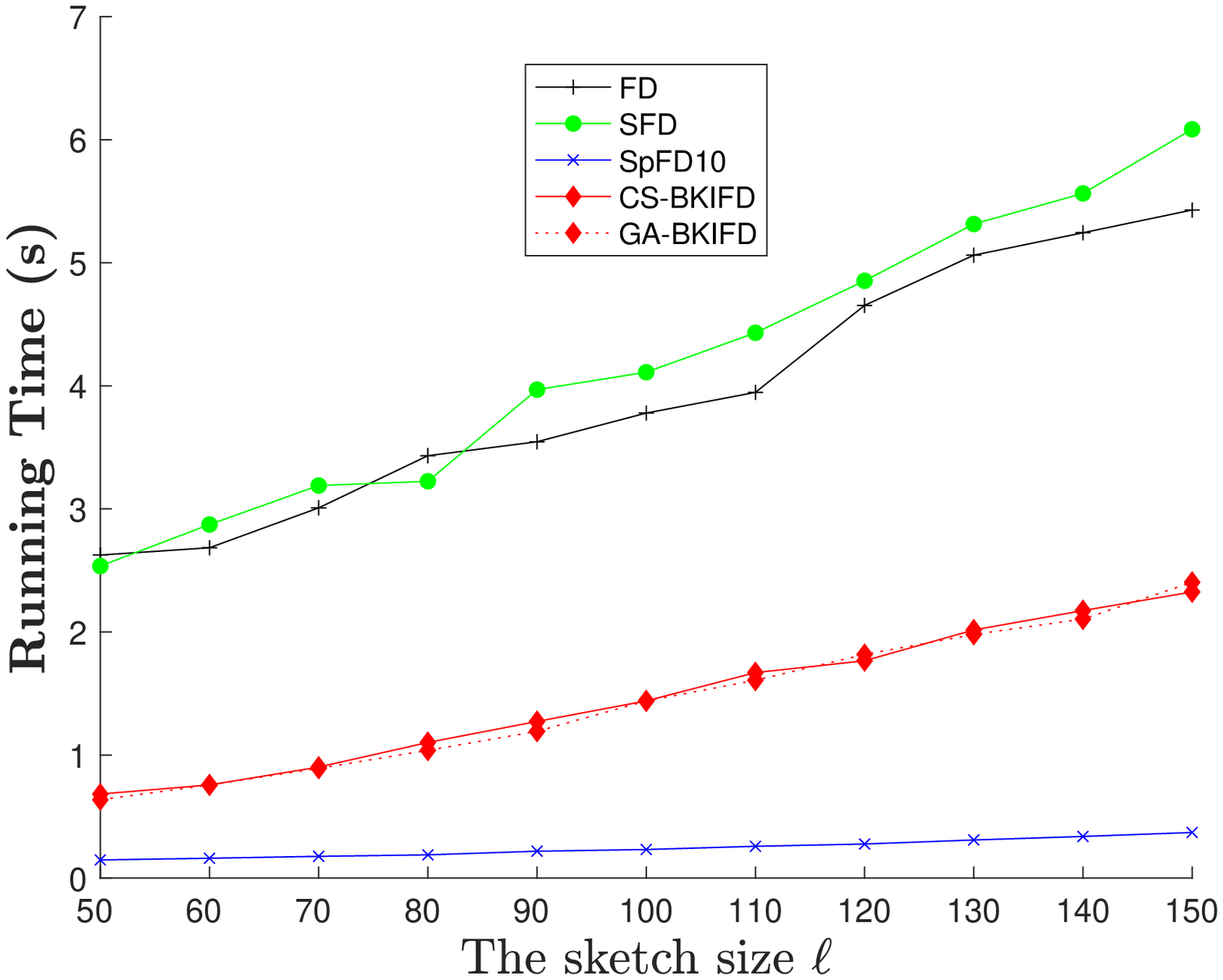}	
	}
	\caption{Comparison results on real datasets: Side0, MovieLens 10M, MovieLens 20M and Protein}
	\label{fig:real1}
\end{figure*}

\begin{figure*}[htbp]
	\centering
	\subfigure{
		\includegraphics[width=0.42\columnwidth]{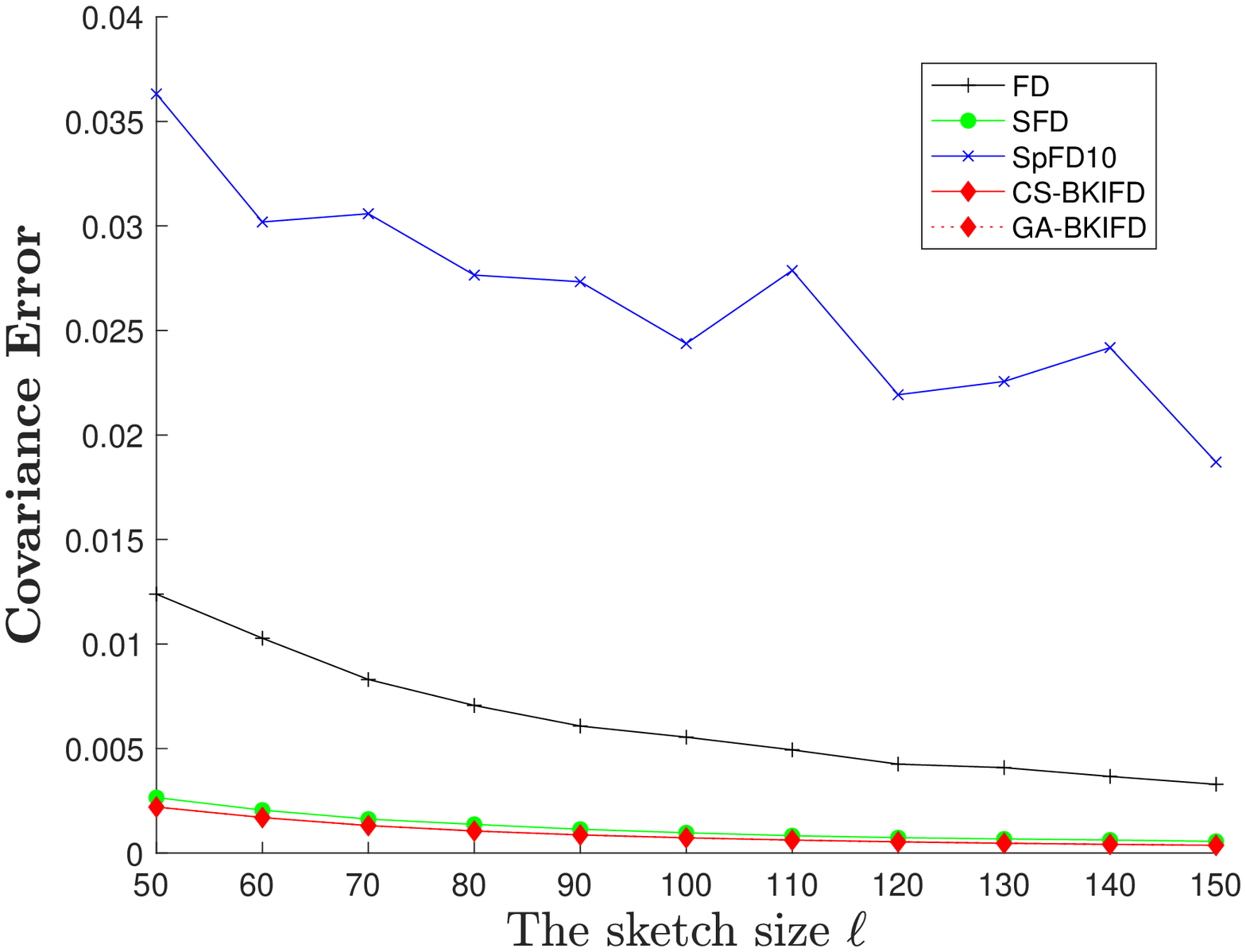}
	}
	\subfigure{
		\includegraphics[width=0.42\columnwidth]{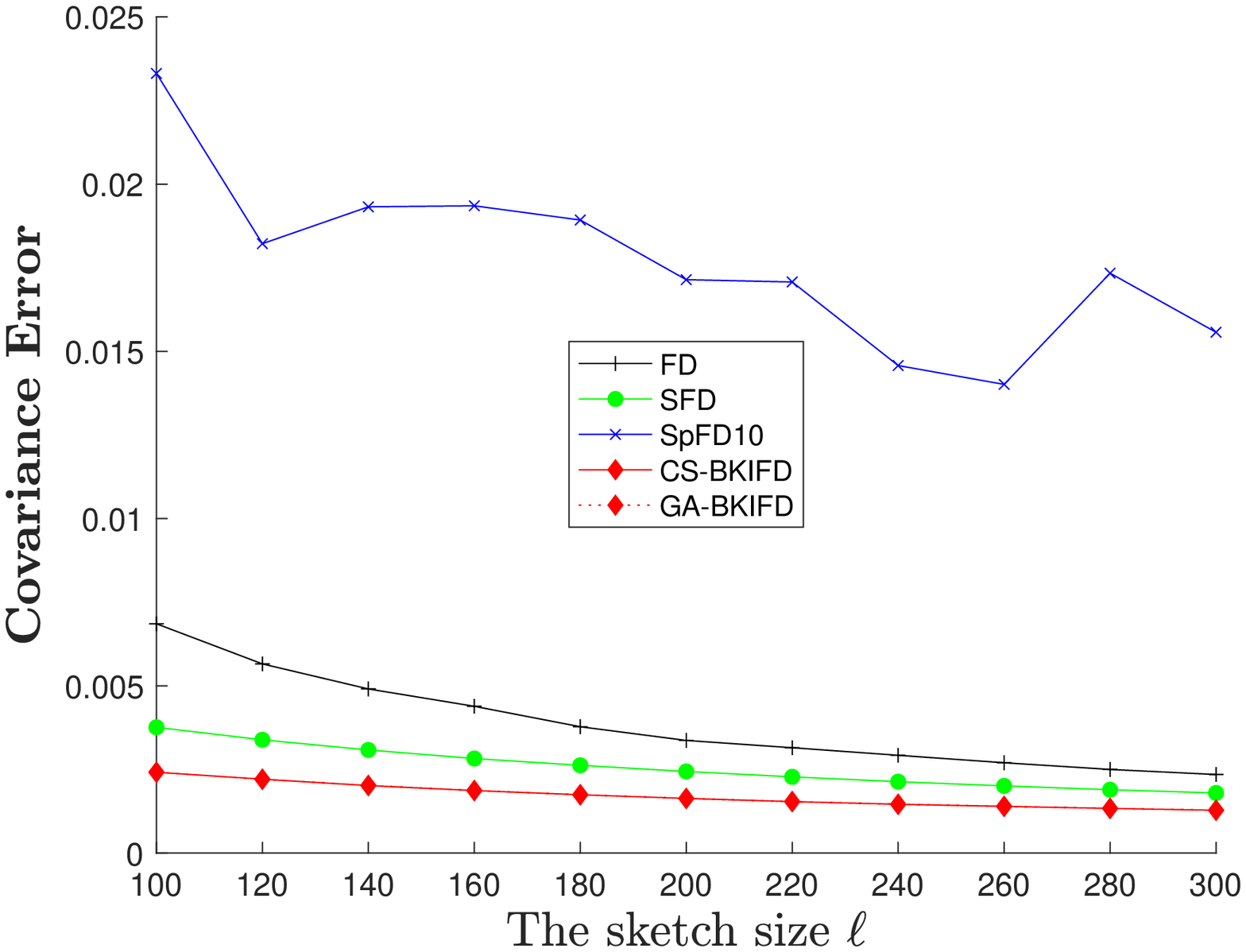}
	}
	\subfigure{
		\includegraphics[width=0.42\columnwidth]{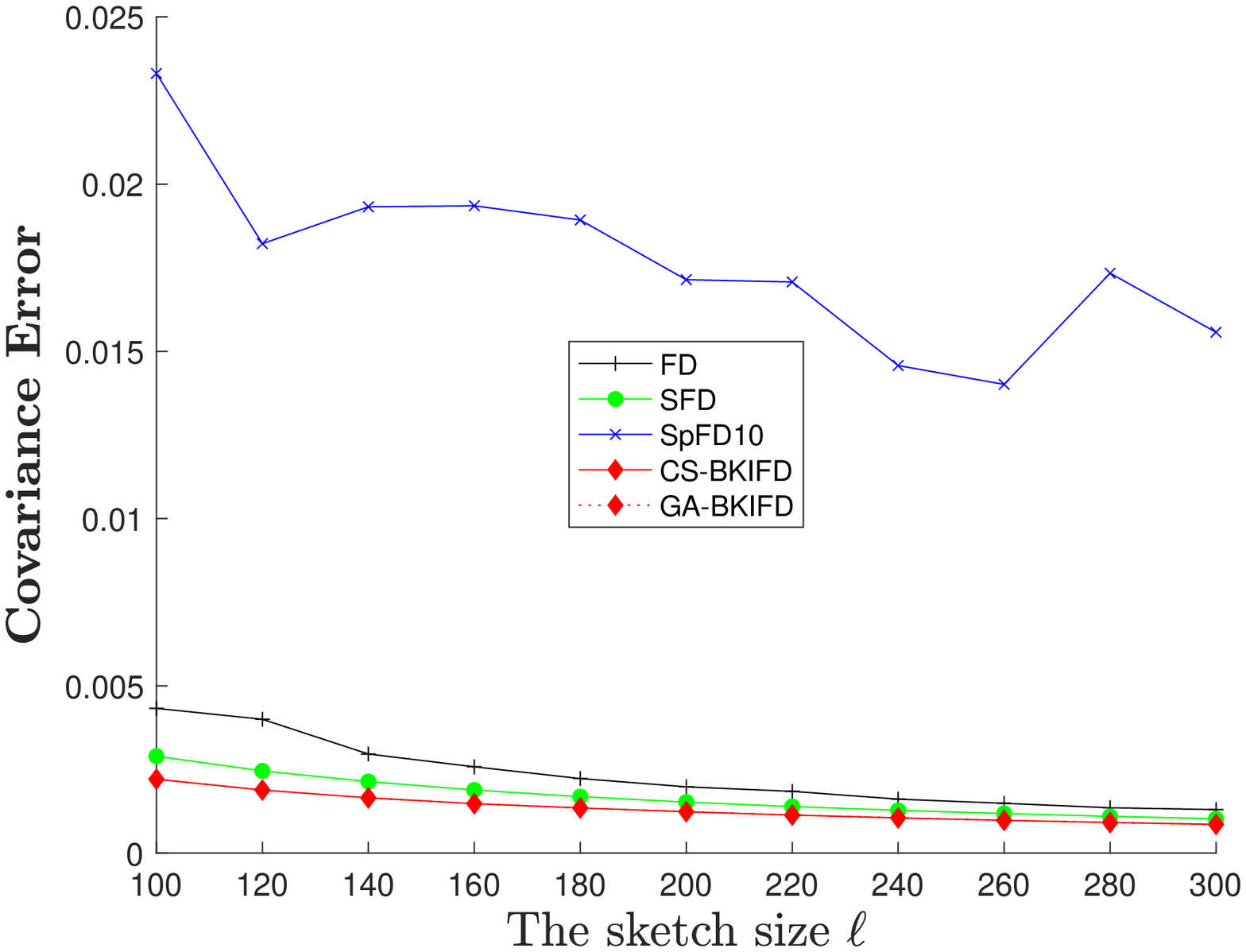}
	}
	\subfigure{
		\includegraphics[width=0.42\columnwidth]{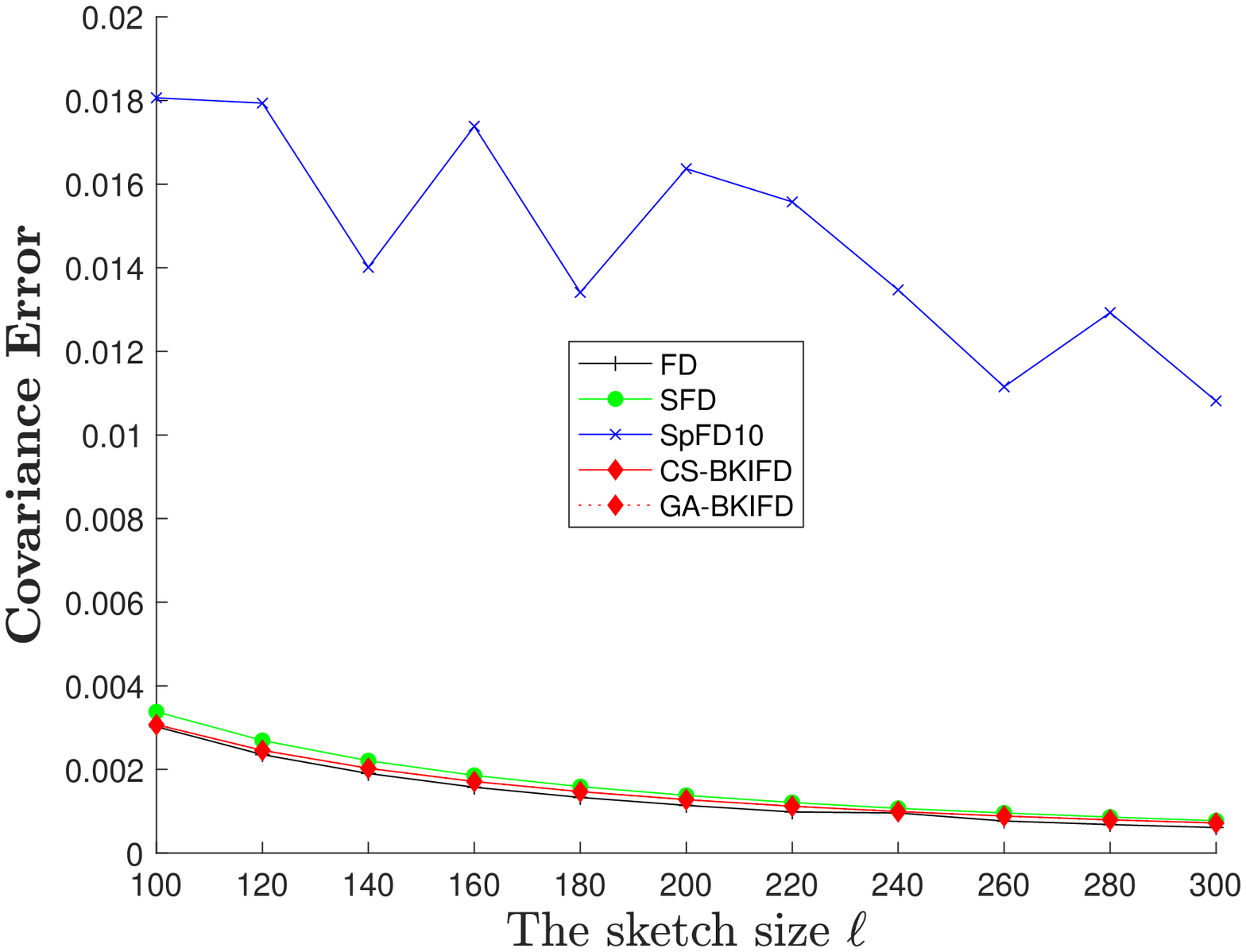}
	}
	\\
	\subfigure{
		\includegraphics[width=0.42\columnwidth]{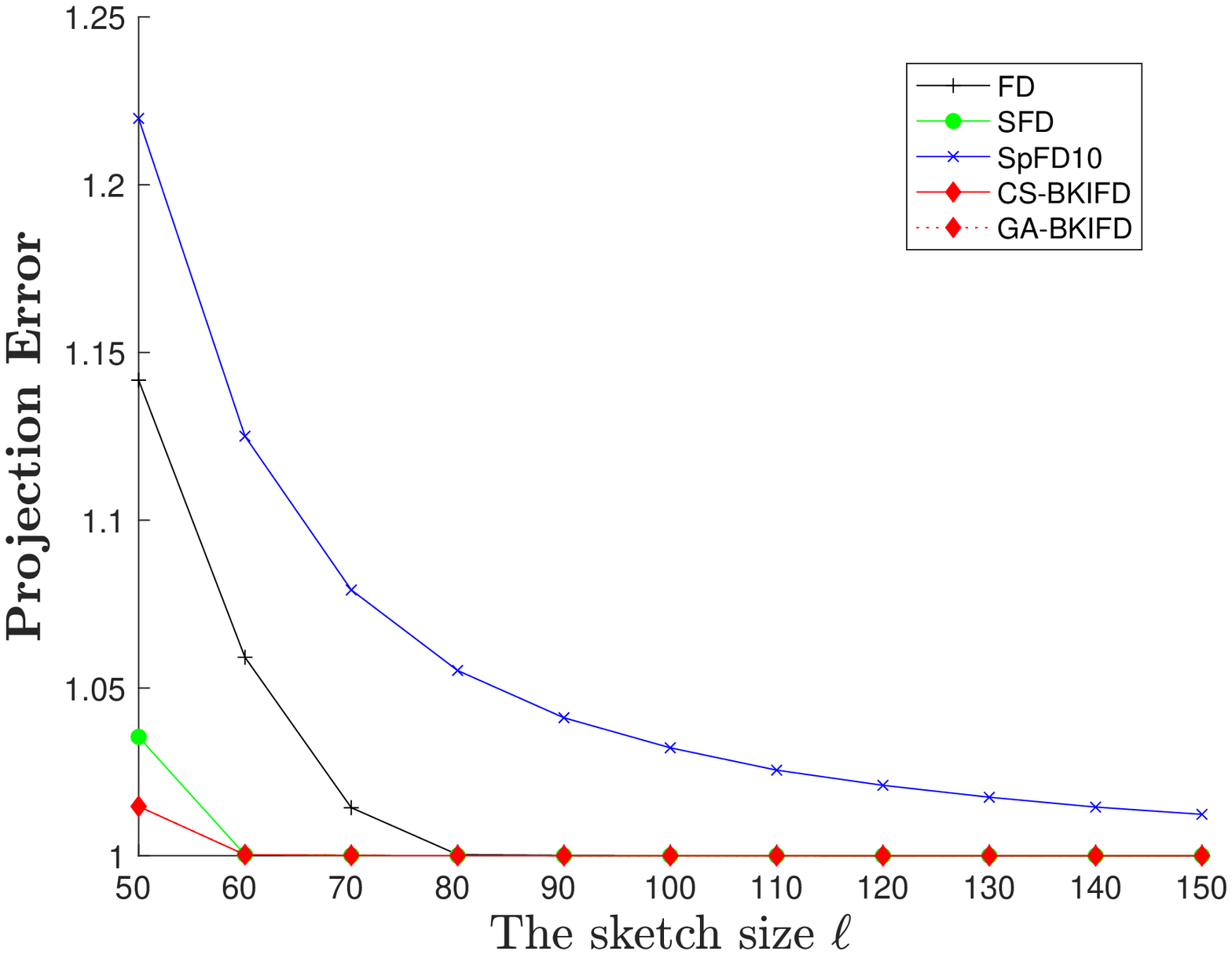}
	}
	\subfigure{
		\includegraphics[width=0.42\columnwidth]{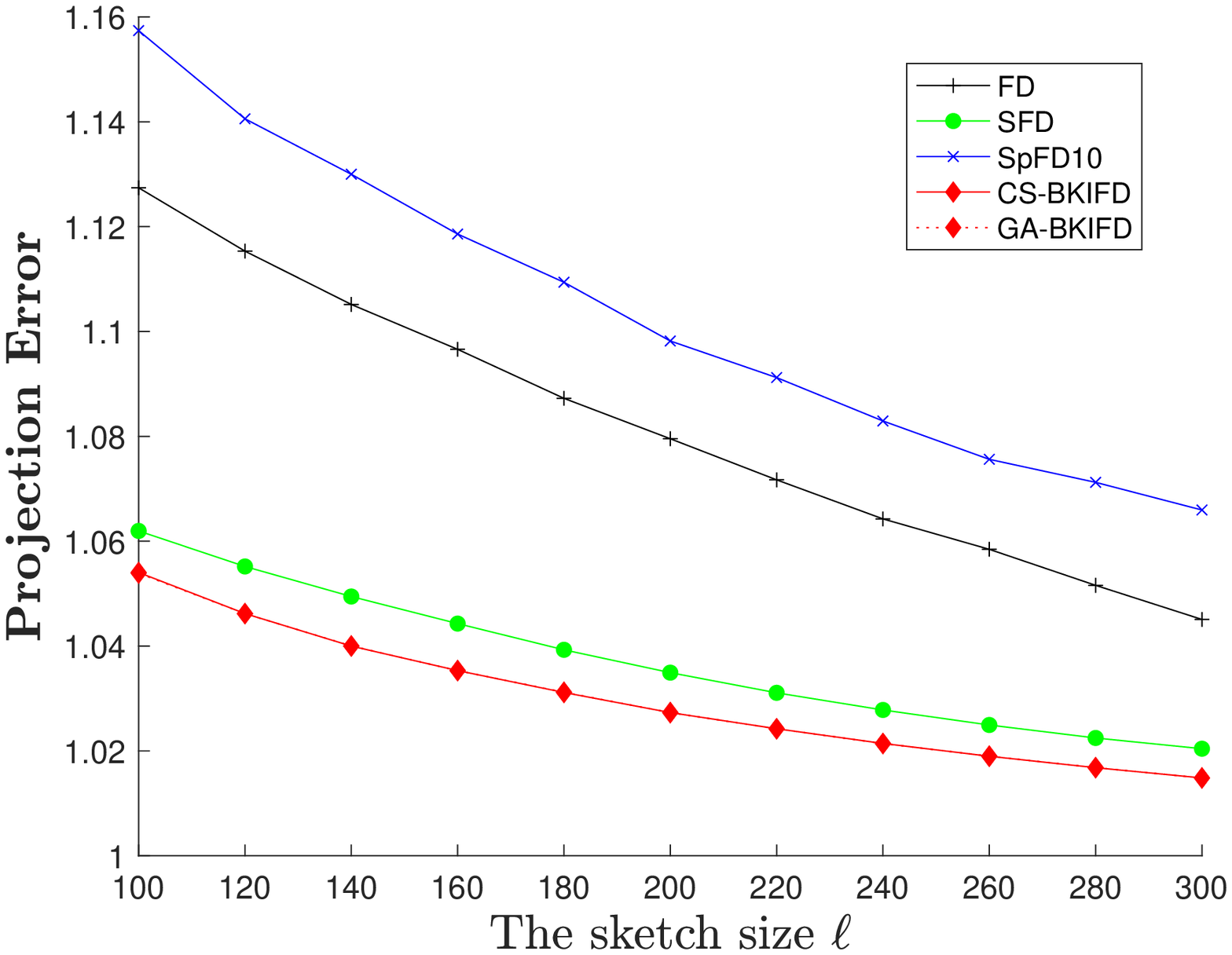}
	}
	\subfigure{
		\includegraphics[width=0.42\columnwidth]{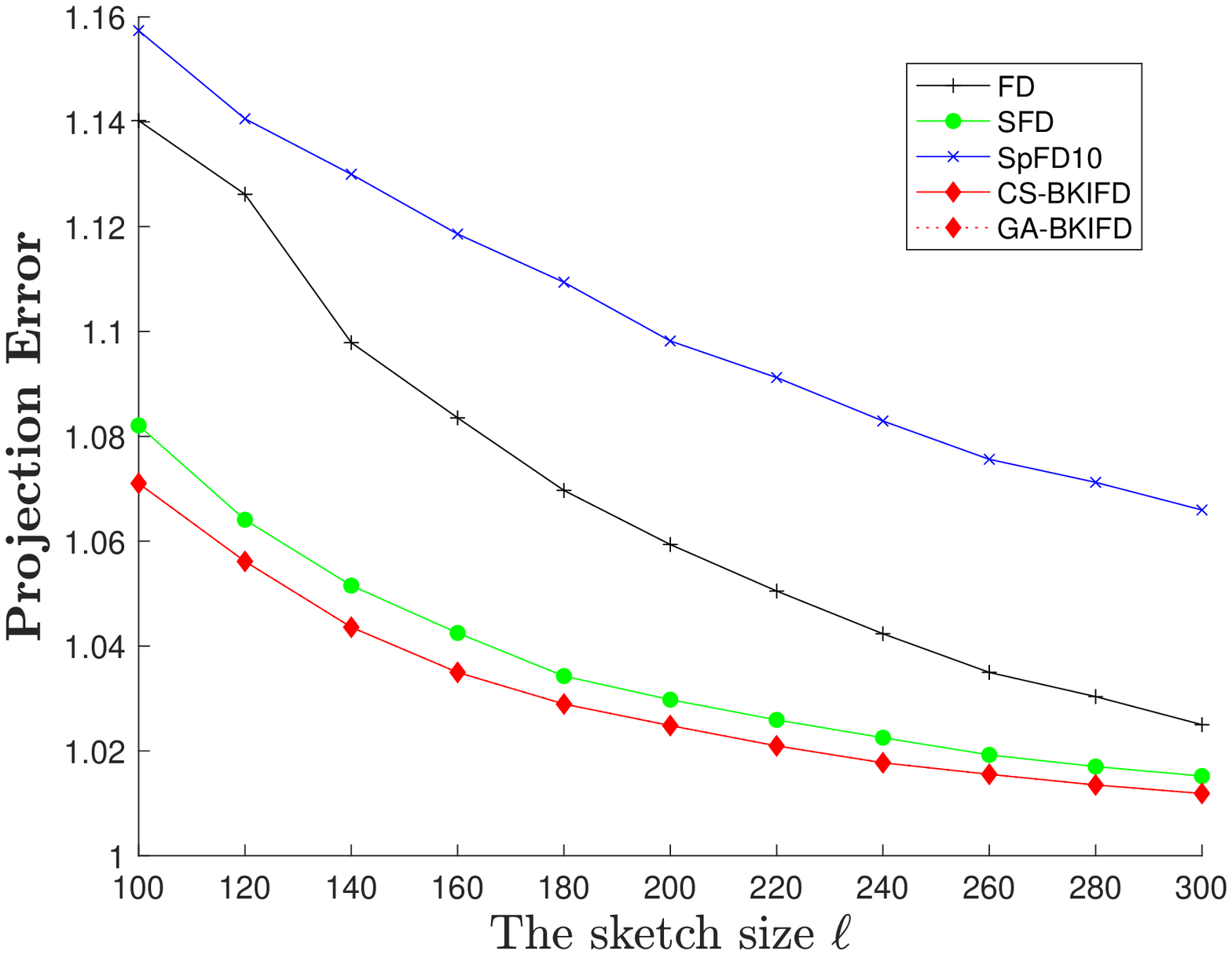}
	}
	\subfigure{
		\includegraphics[width=0.42\columnwidth]{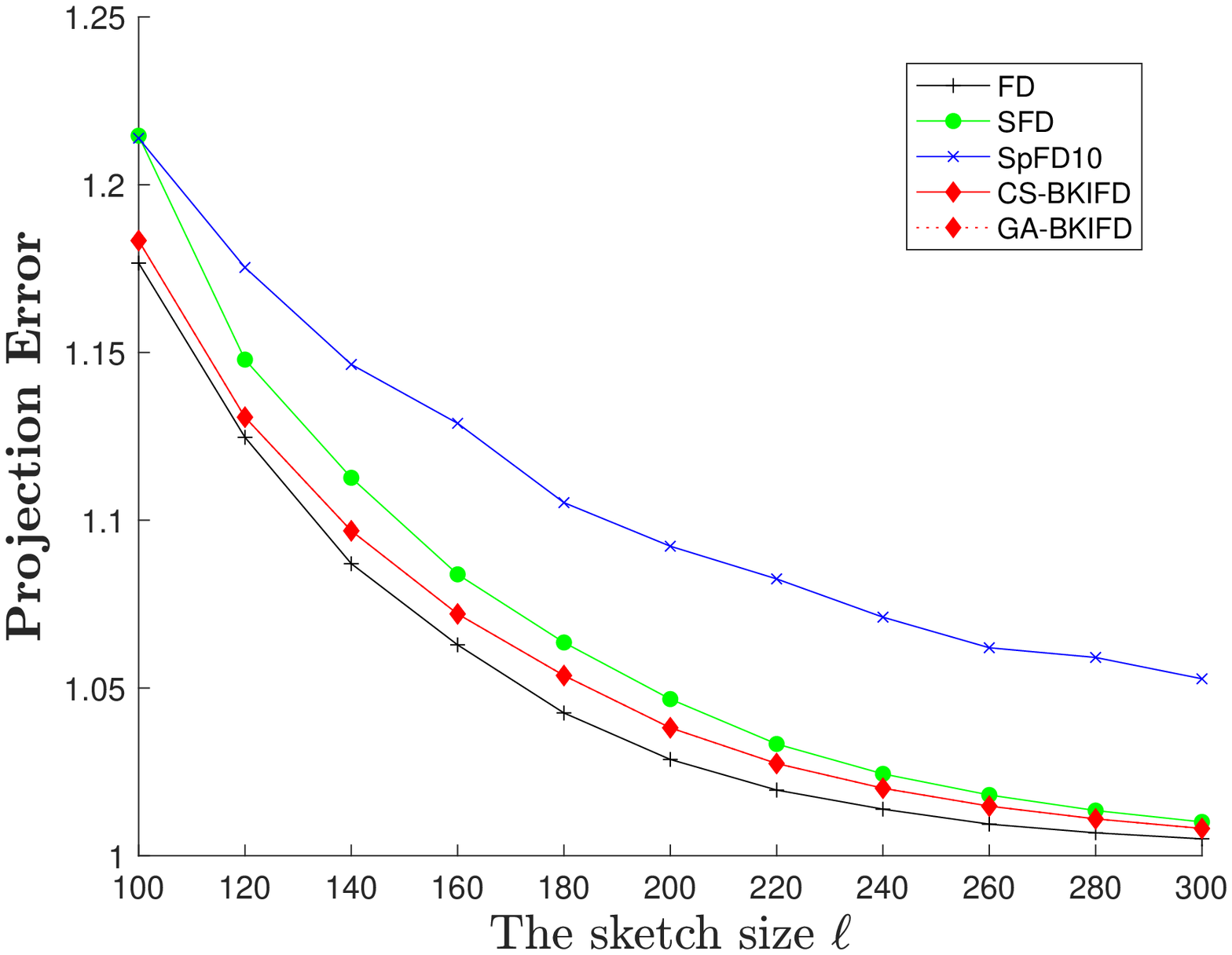}
	}
	\\
	\subfigure[\small{MNIST}]{
		\includegraphics[width=0.42\columnwidth]{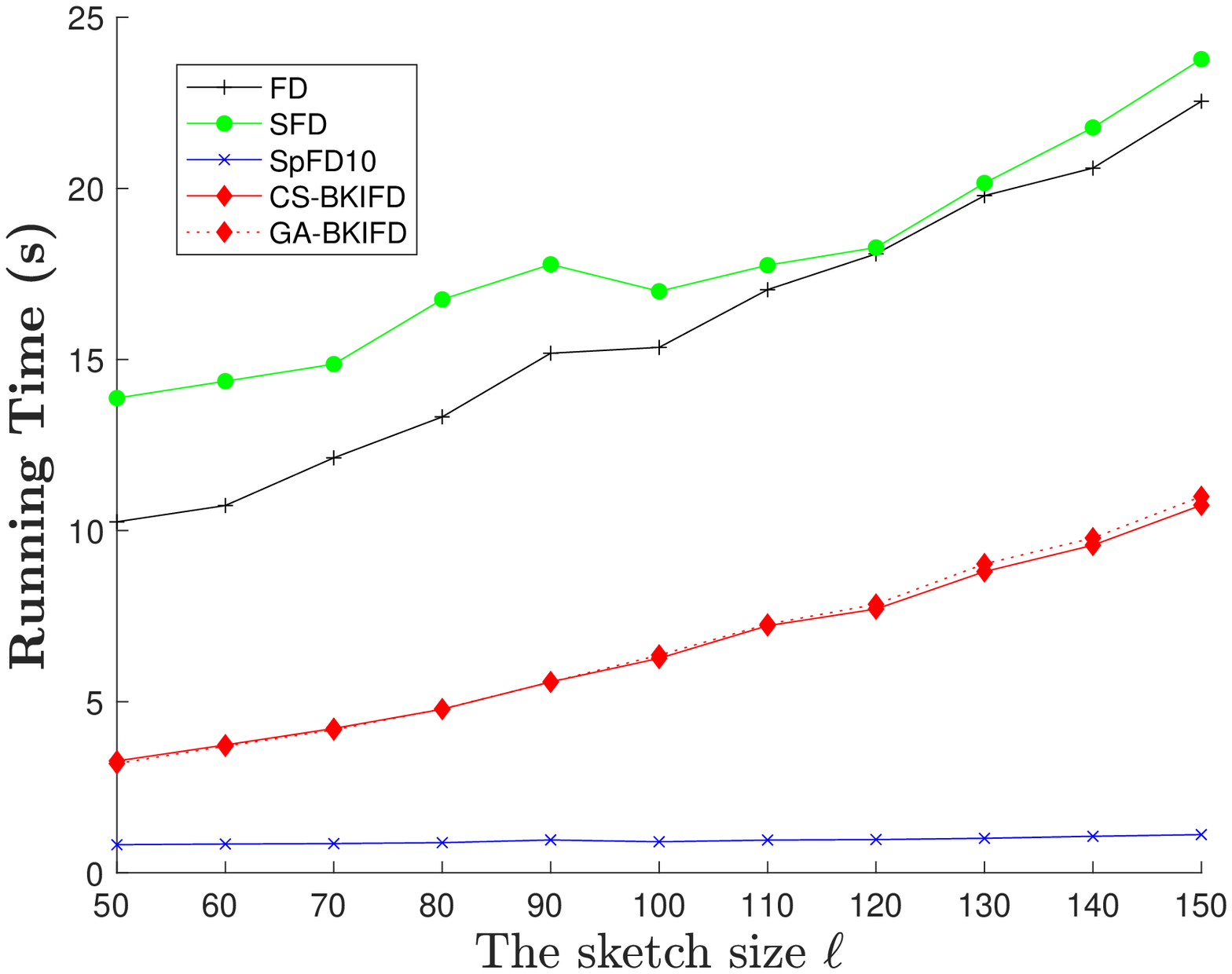}	
	}
	\subfigure[\small{rcv1-small}]{
		\includegraphics[width=0.42\columnwidth]{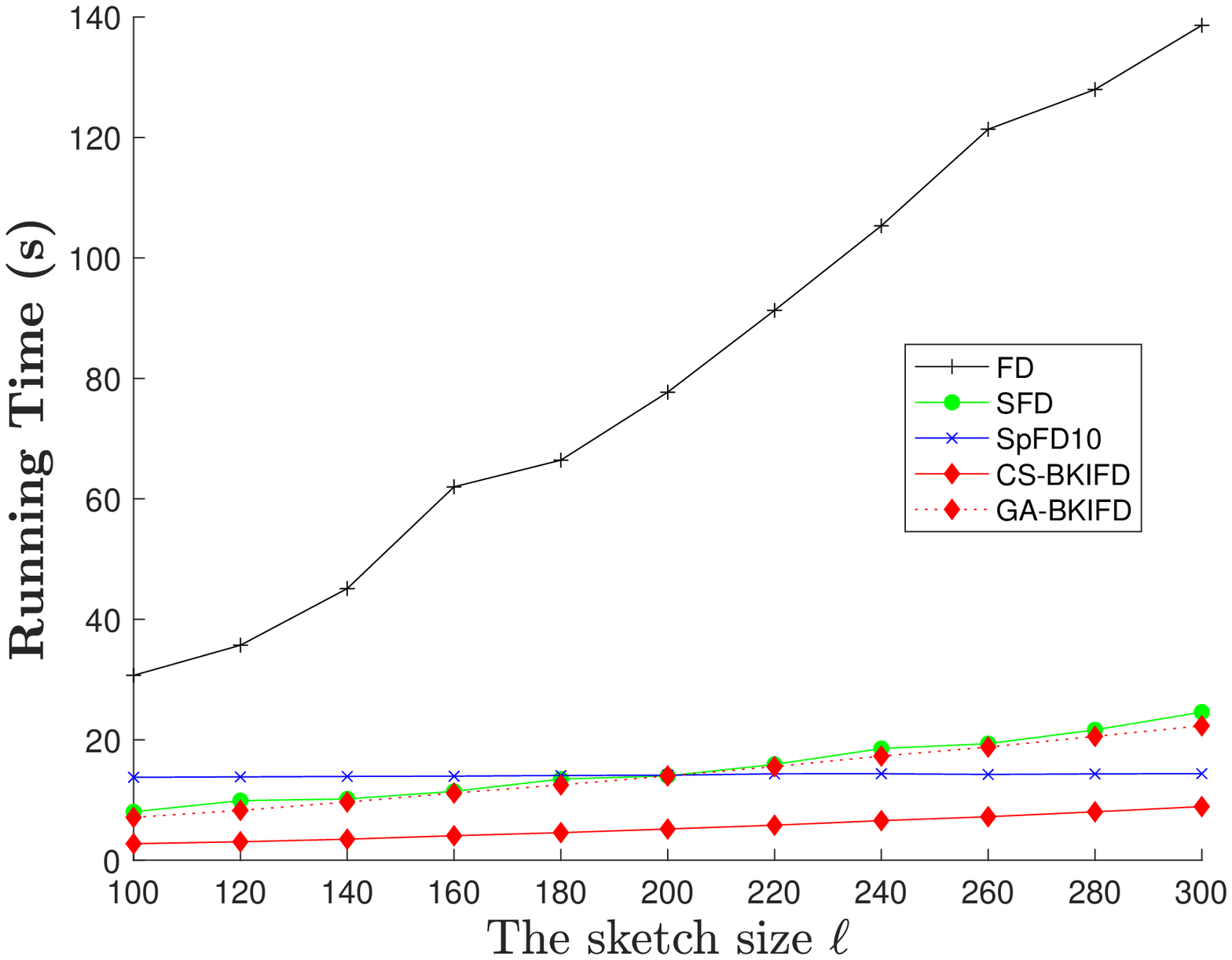}
	}	
	\subfigure[\small{Newsgroups}]{
		\includegraphics[width=0.42\columnwidth]{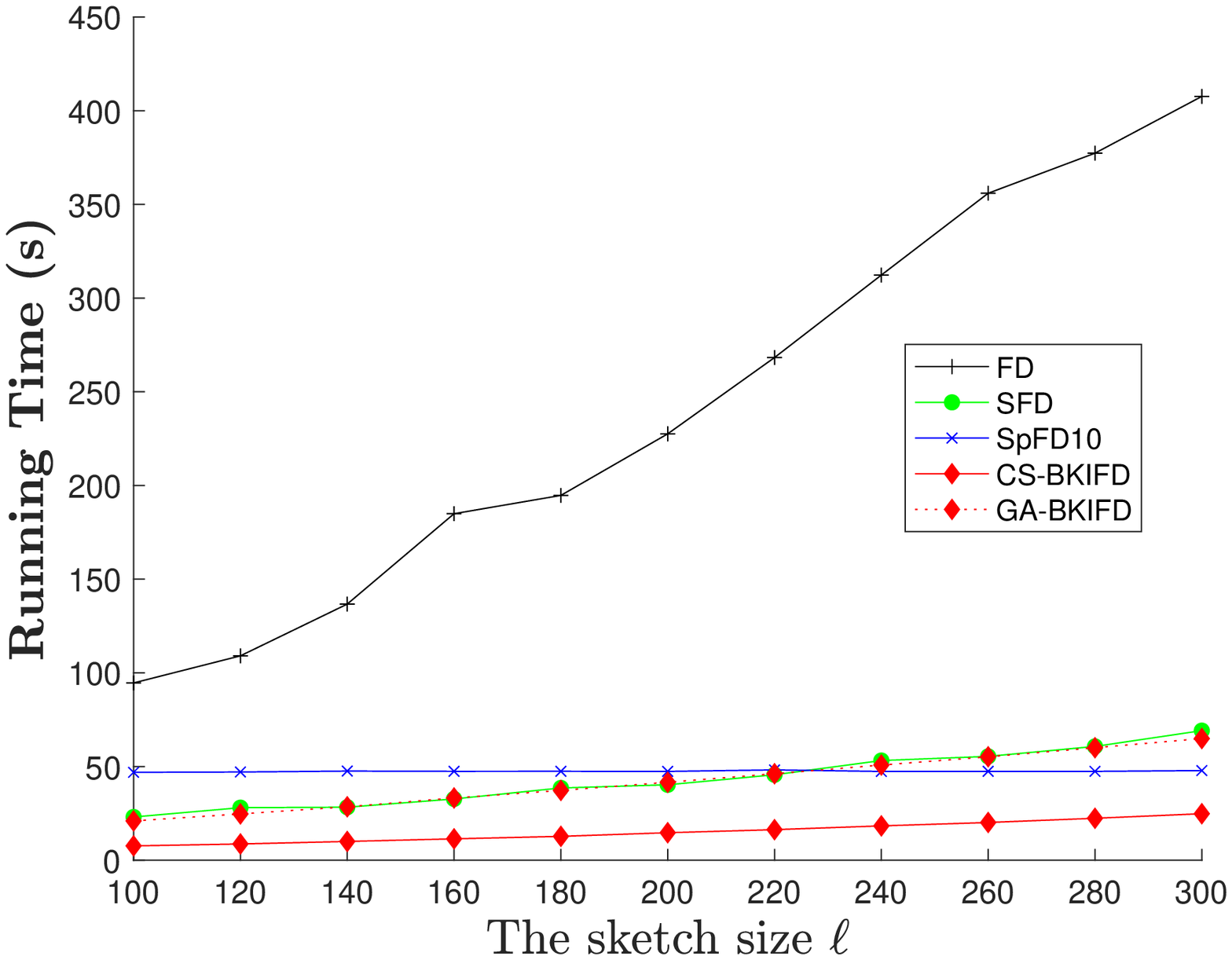}
	}
	\subfigure[\small{amazon7-small}]{
		\includegraphics[width=0.42\columnwidth]{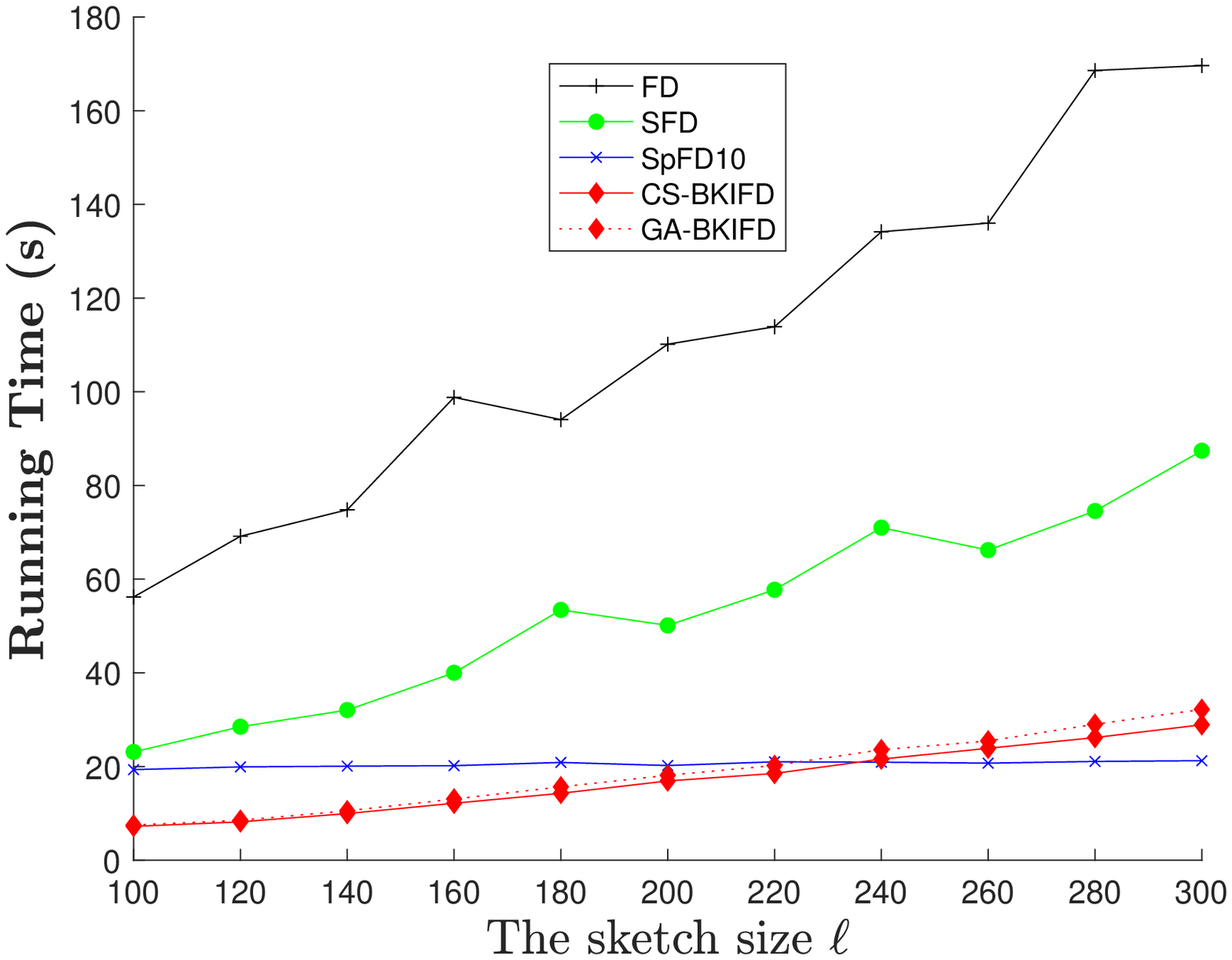}	
	}
	\caption{Comparison results on real datasets: MNIST,  rcv1-small, Newsgroups, amazon7-small}
	\label{fig:real}
\end{figure*}

\subsection{Synthetic Data Experiments}
We consider both dense and sparse synthetic data. The generation of dense data follows  the setting in \cite{ ghashami2016frequent}, that is, we generate $A = SDU + N/\zeta \in \mathbb{R}^{n \times d}$, where $S \in \mathbb{R}^{n \times k}$ is the coefficients matrix with $S_{i,j} \sim \mathcal{N}(0,1)$, $D \in \mathbb{R}^{k \times k }$ is a diagonal matrix with $D_{ii} = 1-(i-1)/k$ that gives linearly diminishing singular values, $U \in \mathbb{R}^{k \times d}$ is the row space matrix with $U U^T = I_k$, and $N \in \mathbb{R}^{n \times d}$ is a noise matrix with $N_{ij} \sim \mathcal{N}(0,1)$. The parameter $\zeta$ determines whether the noise can dominate the signal.  The generation of the sparse data is by random sampling, each row contains roughly $0.1\%$ non-zeros chosen uniformly from $[0,1]$, with the remaining entries as 0. 

In our method, we fix the batch size to the dimension $d$  for dense data and $2d$ for sparse data, and the iteration number $q = 2$. Empirically, setting $m = \ell + p$ is sufficient to estimate an accurate subspace, where $p$ is a small nonnegative integer. For other computing methods, we follow the parameter settings in their original papers. In our experiments, we consider $n = 60000, d = 5000$, $k \in \{50, 200, 500\}$ and $\zeta = 10$, and vary the sketch size $\ell$ to measure the performance. We present their average results in Fig. \ref{fig:exp}. 

Several observations can be easily obtained from Fig. \ref{fig:exp}. Firstly, all compared methods have a tendency to reduce error as the growth of sketch size $\ell$ in both projection and covariance errors.  Secondly, the proposed GA-BKIFD and CS-BKIFD algorithms obtain much lower error bounds compared with two randomized FDs as well as  FD in most cases. And SFD has a close performance as ours but has a much higher computational cost for the dense matrix cases. For FD and SpFD10 methods, they both obtain a worse estimation, especially for the sparse matrix. Thirdly, all randomized methods generally spend less running time than FD. And for the dense matrix cases, though the SpFD10 method is the fastest among all such methods, it fails to find an accurate estimation. In addition, once the input matrix is extremely sparse, by utilizing the structure of CountSketch matrix, CS-BKIFD only takes at most  one sixth of the time compared with GA-BKIFD, and has {the least} running time among all methods. All these verify the effectiveness and the efficiency of the proposed  Algorithm 4, and thus further support the performance guarantees provided by Theorems 1 and 2. 

%demonstrate the performance of our algorithm is substantially better than the theoretical guarantee in Theorem 1 and 2. One possible reason is that the techniques in the proof are still strongly dependent on the properties of FD, thus the advantage of r-BKI algorithm hasn't been exploited completely. 
%
%
%\textit{For SFD, it has a similar performance as ours in Frobenius norm error, but for the covariance error, our performance is consistently much better. Also it is not efficient, the reason behind this is that a large iteration number is required to achieve lower error bound. However, when utilizing Block Krylov Iteration method in our algorithm, only a small number of iterations is sufficient. Another advantage of our method occurs when the sketch size is small, we could still find an accurate estimation. For SpFD10, though it maintains a lowest running time varying the sketch size $\ell$, the precisions are the worst, even compared with FD. As the  sketch size grows, the gap between it and FD is becoming larger, which is obvious in covariance error setting. The inaccurate estimation comes from the directly projection of the original matrix $A$ with the CountSketch matrix. From the experiments, we could observe the covariance error could measure the performance more intrinsically compared with the Frobenius norm. The advantages of CS-BKIFD are revealed when the original data matrix is sparse enough, it only takes approximately one third of the time compared with GA-BKIFD, while maintains the nearly same performance. }

\subsection{Real Data Experiments}

In this section, we evaluate the performance by considering  ten real-world datasets: "w8a", "CIFAR-10", "sido0", "MovieLens-10M", "MovieLens-20M", "Protein", "MNIST", "rcv1-small", "Newsgroups" and "amazon7-small".  The sparsity in the datasets varies from $0.017\%$ to $99.76\%$ and the detailed information is listed in the TABLE I.

\begin{table}[!htbp]
	\centering
	\begin{threeparttable}
		\begin{tabular}{cccccc}
			\toprule  
			dataset& n& d &nnz\% &k &$\ell$\\
			\midrule  
			w8a \cite{platt1999fast} & 64700 & 300 & 3.88&20 & 20:10:120 \\
			%			{color{bule}
			\midrule
			 	CIFAR-10  \cite{krizhevsky2009learning} & 60000 & 3072 & 99.76&20&20:10:120 \\
			\midrule
			 sido0 \cite{guyon2008design} &12678 & 4932&9.84&20&20:10:120 \\
			\midrule
			 MovieLens-10M\textsuperscript{1} \cite{harper2015movielens} & 71567 & 3000&3.24&50&50:10:150  \\
			\midrule
			 MovieLens-20M\textsuperscript{1} \cite{harper2015movielens} & 138493 & 3000&3.99&50&50:10:150 \\
			\midrule
			Protein \cite{wang2002application} &  24387 & 357 & 28.2 & 50 & 50:10:120 \\
			\midrule 
			MNIST \cite{lecun1998gradient} & 70000& 784 & 19.14 & 50 & 50:10:150\\
			\midrule
			rcv1-small \cite{teng2018fast}& 47236& 3000 &0.14& 100 & 100:20:300\\
			\midrule
			Newsgroups\textsuperscript{2}  \cite{ghashami2016efficient}& 130107& 3000 & 0.12& 100 & 100:20:300\\
			\midrule
			amazon7-small \cite{teng2018fast} & 262144 & 1500 & 0.017 & 100 & 100:20:300 \\
			\bottomrule 
		\end{tabular}
		
		\begin{tablenotes}
			\item[1]  For the MovieLens datasets, the original data matrix has  10681 and 27278 columns separately for 10M and 20M. We extract the first 3000 columns.
			\item[2] For this dataset, the original data matrix has 11314 rows and 130107 columns. We extract the first 3000 columns and use the transpose.
		\end{tablenotes}
	\end{threeparttable}
	\caption{Summary of real datasets}
\end{table}	

For the  three  datasets with small feature dimension, i.e. w8a, Protein and MNIST,  we utilize a larger batch size $1000$ in our method, and for the other  datasets, we set the batch size as in the synthetic data experiments. For other competing methods, we still keep the settings of the original papers. According to the results shown in Figs. \ref{fig:exp}-\ref{fig:real}, we observe that our algorithm outperforms other ones in terms of accuracy for nearly all the  datasets. And for the last three extremely sparse datasets, our algorithm still achieves a lowest covariance error among nearly all these methods, and a comparable projection error with the best one. The performance of SpFD10 is unstable, even with a larger sketch size, the covariance error may be higher. This may because that the sparse subspace embedding in SpFD10 fails to capture the important information underling the input matrix. Additionally,  SFD attains  comparable accuracy to ours while it has a higher computational cost. All these results, together with the previous results on synthetic data, demonstrate that the proposed r-BKIFD provides great improvement over other randomized and traditional FD algorithms, both in terms of computational efficiency and accuracy.

%, especially for the large dataset. In the Frobenius norm, our method could only utilize a smaller sketch size to achieve the best low rank approximation, which could be found in the datasets w8a and MNIST. The covariance error linearly decreases for all methods except the SpFD10, which has an unstable performance in most datasets. In the sparse datasets such as rcv1-small and Newsgroup, CS-BKIFD could even cost less time than SpFD10, this is mainly  due to the sparsity of this dataset. }
%Based on the experimental results in sythetic datasets and real datasets, our method reveals great improvement compared with other randomized FD methods by maintaining accuracy as well as efficiency. We could also observe the randomized technique could deeply accelerate the FD algorithm, by integrating the Block Krylov Iteration, the accuracy is also maintained. 
\section{Conclusion}
In this paper, we proposed a novel algorithm named r-BKIFD to alleviate the inaccuracy issue in the randomized FD variants for low-rank approximation. Different from the existing ones embedding random projection technique directly, which may lead to the loss of some important information of the original matrix during the projection process, the basic  idea of r-BKIFD is to incorporate the Block Krylov Iteration technique that could capture a more accurate projection subspace into the randomized FD. In the new algorithmic framework, we consider two types of random matrix, i.e. Gaussian and CountSketch matrices.  Our rigorous theoretical analysis reveals that the proposed r-BKIFD gives a comparable error bound with traditional FD. The extensive experiments on both synthetic and real data further demonstrate that r-BKIFD outperforms traditional FD and its randomized variants in most cases in terms of computational efficiency and accuracy. 

Noting that some real-world  data is mostly of the form of multi-dimensional arrays (or say, tensors),  such as videos and hyperspectral images, thus how to extend the proposed procedure to get efficient low-rank approximation of such tensor data is the focus of our future study.

{\appendix[Proofs]
To prove Theorem \ref{thm2}, we shall  list the following auxiliary property for Gaussian random matrix, regarding to the error bound of extreme singular values.
\begin{lemma}[Corollary 5.35 in \cite{vershynin2010introduction}] \label{Corollary5-35}
	Let $A$ be an $N \times n$ matrix whose entries are independent standard normal random variables. Then for every $\epsilon \geq 0$, with probability at least $1-2 \exp \left(-\epsilon^{2} / 2\right)$ one has
	$$
	\sqrt{N}-\sqrt{n}-\epsilon \leq s_{\min }(A) \leq s_{\max }(A) \leq \sqrt{N}+\sqrt{n}+\epsilon.
	$$
\end{lemma} 

\noindent{\it Proof of Theorem \ref{thm2}.} By the triangle inequality, we have $$ \left\|A^{ T} A-B^{T} B\right\|_{2} \leq \left\|A^{ T} A-P^{T} P\right\|_{2}+\left\|P^{ T} P-B^{T} B\right\|_{2}.$$ 
We first bound $\left\|A^{ T} A-P^{T} P\right\|_{2}$. As mentioned before, the input matrix $A$ exists as a streaming fashion with each batch arriving in order, that is, $A =\left[A_{(1)} ; \cdots ; A_{(s)}\right]\in \mathbb{R}^{n \times d}$, where $A_{(i)}\in \mathbb{R}^{\frac{n}{s} \times d}$. Then we can rewrite $\left\|A^{ T} A-P^{T} P\right\|_{2}$ as
\begin{align}
\left\|A^{ T} A-P^{T} P\right\|_{2}
&=\left\|\sum_{i=1}^{s}(A_{(i)}^{ T} A_{(i)}-P_{(i)}^{T} P_{(i)})\right\|_{2}, \notag	
\end{align}
where $P =\left[P_{(1)} ; \cdots ; P_{(s)}\right]$ with each $P_{(i)}$ obtained by the BKI compression on $A_{(i)}$. 

Let $M_{(i)}=A_{(i)}^{ T} A_{(i)}-P_{(i)}^{T} P_{(i)}$. By noting that $\left\{M_{(i)}\right\} _{i=1}^{s}$ are independent random matrices, and $\mathbb{E}[M_{(i)}-\mathbb{E}[M_{(i)}]]=0$, we thus can apply the matrix Bernstein inequality previously given in Lemma \ref{Matrix-Bernstein-inequality} to bound $M_{(i)}-\mathbb{E}[M_{(i)}]$. To this end, the necessary step is to calculate  the covariance error bound $R$ and the variance parameter $\sigma^{2}$. As for $R$, combining the triangle inequality with Jensen's inequality, we have
\begin{align}
\left\|M_{(i)}-\mathbb{E}[M_{(i)}]\right\|_{2}\le&\left\|M_{(i)}\right\|_{2}+\left\|\mathbb{E}[M_{(i)}]\right\|_{2}  \notag\\
\le&\left\|M_{(i)}\right\|_{2}+\mathbb{E}\left\|M_{(i)}\right\|_{2}.  \label{tri}
\end{align}	
Due to $M_{(i)}=A_{(i)}^{ T} A_{(i)}-P_{(i)}^{T} P_{(i)}$, and $P_{(i)}=Z_{(i)}^{T}A_{(i)}$, we can reformulate $M_{(i)}$ as $A_{(i)}^{ T}\left(I-Z_{(i)} Z_{(i)}^{T}\right) A_{(i)}$. By the fact that $I-Z_{(i)} Z_{(i)}^{T}$ is a projection, i.e., $$I-Z_{(i)} Z_{(i)}^{T}=\left(I-Z_{(i)} Z_{(i)}^{T}\right)^{T}\left(I-Z_{(i)} Z_{(i)}^{T}\right),$$ we have
\begin{align}
&\left\|M_{(i)}\right\|_{2} \notag\\
=&\left\|A_{(i)}^{ T}\left(I-Z_{(i)} Z_{(i)}^{T}\right)^{T}\left(I-Z_{(i)} Z_{(i)}^{T}\right) A_{(i)}\right\|_{2}  \notag\\
= &\left\| A_{(i)}-Z _{(i)}Z_{(i)}^{T} A_{(i)}\right\|_{2}^{2} \notag\\
\leq&\left(\left\|\phi\left([\Sigma_{(i)}]_{\ell, \perp}\right)\right\|_{2}\left\|[V_{(i)}]_{\ell, \perp}^{T} X\left([V_{(i)}]_{\ell}^{T} X\right)^{\dagger}\right\|_{F}\right.\notag\\
&\left.+\left\|A_{(i)}-[A_{(i)}]_{\ell}\right\|_{2}\right)^{2}, \label{lemma1proof}	
\end{align}
where the inequality (\ref{lemma1proof}) follows from Lemma \ref{lemma1}. Since the rows of $[V_{(i)}]^{T}$ are orthonormal, the entries
of $[V_{(i)}]_{\ell}^{T} X\in \mathbb{R}^{\ell \times m}$ are independent Gaussians. We use the same probability of success for each $[V_{(i)}]_{\ell}^{T}X$, then according to Lemma \ref{Corollary5-35}, with probability at least $1-2\exp(-\varepsilon^{2}/2)$, we have
$$ \label{singular-gaussian}
\sigma_{\ell}\left([V_{(i)}]_{\ell}^{T} X\right) \ge \sqrt{m}-\sqrt{\ell}-\varepsilon, \text{ and } \left\|X\right\|_{2}\leq\sqrt{d}+\sqrt{m}+\varepsilon.
$$		
By the sub-multiplicativity property of the Frobenius norm, we then further get that 
\begin{align}
\left\|[V_{(i)}]_{\ell, \perp}^{T} X\left([V_{(i)}]_{\ell}^{T} X\right)^{\dagger}\right\|_{F} &\leq\left\|[V_{(i)}]_{\ell, \perp}^{T}\right\|_{ F} \left\|X\right\|_{2}\left\|\left([V_{(i)}]_{\ell}^{T} X\right)^{\dagger}\right\|_{2} \notag\\
&\leq \frac{\sqrt{d-\ell}(\sqrt{d}+\sqrt{m}+\varepsilon)}{\sqrt{m}-\sqrt{\ell}-\varepsilon}.\label{sub-mul-fro}
\end{align}
Combining Lemma \ref{lemma2} with inequalities (\ref{lemma1proof}) and (\ref{sub-mul-fro}) can get that
\begin{align}
&\left\|M_{(i)}\right\|_{2} \le \left\| A_{(i)}-Z _{(i)}Z_{(i)}^{T} A_{(i)}\right\|_{2}^{2}\notag\\
\le&\left(1+\frac{4 }{2^{(2 q+1) \min \{\sqrt{\gamma}, 1\}}}\frac{\sqrt{d-\ell}(\sqrt{d}+\sqrt{m}+\varepsilon)}{\sqrt{m}-\sqrt{\ell}-\varepsilon}\right)^{2}\notag\\
&\times \left\|A_{(i)}-[A_{(i)}]_{\ell}\right\|_{2}^{2} \notag\\
=&\left(1+\delta\right)\left\|A_{(i)}-[A_{(i)}]_{\ell}\right\|_{2}^{2}, \label{delta}
\end{align}
where we denote $\left(1+\frac{4}{2^{(2 q+1) \min \{\sqrt{\gamma}, 1\}}}\frac{\sqrt{d-\ell}(\sqrt{d}+\sqrt{m}+\varepsilon)}{\sqrt{m}-\sqrt{\ell}-\varepsilon}\right)^{2}$ by $1+\delta$. Denote $\sigma_{i}(\cdot)$ and $\lambda_{i}(\cdot)$ as the singular value and eigenvalue, respectively. Then we can reformulate $\left\|A_{(i)}-[A_{(i)}]_{\ell}\right\|_{2}^{2}$ as 
	\begin{align}
	\left\|A_{(i)}-[A_{(i)}]_{\ell}\right\|_{2}^{2}=\sigma_{\ell+1}^2(A_{(i)})=\lambda_{\ell+1}(A_{(i)}^TA_{(i)}).\label{transform}
	\end{align}
	According to Lemma \ref{minmax} and the fact that 
	$$
	R_{\sum_{i=1}^{s}A_{(i)}^TA_{(i)}}(x)\ge \max_{i\in[s]} R_{A_{(i)}^TA_{(i)}}(x),
	$$
	we can further upper bound (\ref{transform}) as
	\begin{align}
	&\lambda_{\ell+1}(A_{(i)}^TA_{(i)})\le	\lambda_{\ell+1}(\sum_{i=1}^{s}A_{(i)}^TA_{(i)})\notag\\
	=&\lambda_{\ell+1}(A^TA)=\sigma_{\ell+1}^2(A)=\left\|A-A_{\ell}\right\|_{2}^{2}.\notag
	\end{align}
Hence, it then follows the inequality (\ref{tri}) that 
$$
\left\|M_{(i)}-\mathbb{E}[M_{(i)}]\right\|_{2}\le 2\left(1+\delta\right)\left\|A-A_{\ell}\right\|_{2}^{2}.
$$
This completes the calculation of $R$.

Next, we shall focus on calculating $\sigma^{2}$. Due to the fact that $M_{(i)}-\mathbb{E}[M_{(i)}]$ is a symmetric matrix, we have
$\sigma^{2}=\left\|\sum_{i=1}^{s}\mathbb{E}[(M_{(i)}-\mathbb{E}[M_{(i)}])^2]\right\|_{2}$. Thus by the triangle inequality, we further have $\sigma^{2}\le\sum_{i=1}^{s}\left\|\mathbb{E}[(M_{(i)}-\mathbb{E}[M_{(i)}])^2]\right\|_{2}$. Expanding the square gives 
$$
\mathbb{E}[(M_{(i)}-\mathbb{E}[M_{(i)}])^2]=\mathbb{E}[M_{(i)}^{2}]-\mathbb{E}[M_{(i)}]^{2}\preceq\mathbb{E}[M_{(i)}^{2}].
$$
Consequently, $\left\|\mathbb{E}[(M_{(i)}-\mathbb{E}[M_{(i)}])^2]\right\|_{2} \le \left\|\mathbb{E}[M_{(i)}^{2}]\right\|_2$.

We then bound $\mathbb{E}[M_{(i)}^{2}]$ as 
\begin{align}
&\mathbb{E}[M_{(i)}^{2}] =\mathbb{E}\left[\left(A_{(i)}^{ T} A_{(i)}-P_{(i)}^{T} P_{(i)}\right)^{2}\right] \notag \\
=&\mathbb{E}\left[\left(\left(A_{(i)}-Z_{(i)} Z_{(i)}^{T}A_{(i)}\right)^{T}\left(A_{(i)}-Z_{(i)} Z_{(i)}^{T}A_{(i)}\right) \right)^{2}\right] \notag \\
\preceq&\mathbb{E}\left[\left\|A_{(i)}-Z_{(i)} Z_{(i)}^{T}A_{(i)}\right\|_{2}^{2}\right.\notag\\
&\times \left.\left(A_{(i)}-Z_{(i)} Z_{(i)}^{T}A_{(i)}\right)^{T}\left(A_{(i)}-Z_{(i)} Z_{(i)}^{T}A_{(i)}\right) \right] \label{norm-ine} \\
\preceq&\left(1+\delta\right)\left\|A-A_{\ell}\right\|_{2}^{2}\mathbb{E}\left[\left(A_{(i)}-Z_{(i)} Z_{(i)}^{T}A_{(i)}\right)^{T}\left(A_{(i)}-Z_{(i)} Z_{(i)}^{T}A_{(i)}\right) \right] \label{A-ZZA}\\
=&\left(1+\delta\right)\left\|A-A_{\ell}\right\|_{2}^{2}\mathbb{E}[M_{(i)}], \label{expection-m}
\end{align}
where the inequality (\ref{norm-ine}) follows the fact that for any $y\in \mathbb{R}^{d}$,
\begin{align}
&y^{T}\left(\left(A_{(i)}-Z_{(i)} Z_{(i)}^{T}A_{(i)}\right)^{T}\left(A_{(i)}-Z_{(i)} Z_{(i)}^{T}A_{(i)}\right) \right)^{2}y \notag\\
=&\left\|y^{T}\left(A_{(i)}-Z_{(i)} Z_{(i)}^{T}A_{(i)}\right)^{T}\left(A_{(i)}-Z_{(i)} Z_{(i)}^{T}A_{(i)}\right) \right\|_{2}^{2} \notag\\
\le&\left\|y^{T}\left(A_{(i)}-Z_{(i)} Z_{(i)}^{T}A_{(i)}\right)^{T}\right\|_{2}^{2}\left\|\left(A_{(i)}-Z_{(i)} Z_{(i)}^{T}A_{(i)}\right) \right\|_{2}^{2} \notag \\
=&\left\|\left(A_{(i)}-Z_{(i)} Z_{(i)}^{T}A_{(i)}\right) \right\|_{2}^{2}\notag\\
&\times y^{T}\left(A_{(i)}-Z_{(i)} Z_{(i)}^{T}A_{(i)}\right)^{T}\left(A_{(i)}-Z_{(i)} Z_{(i)}^{T}A_{(i)}\right) y, \notag
\end{align}
and the inequality (\ref{A-ZZA}) holds by (\ref{delta}). 
Therefore,  
$$
\sigma^{2}\le\sum_{i=1}^{s}\left\|\mathbb{E}[M_{(i)}^{2}]\right\|_{2} 
\le\sum_{i=1}^{s}\left(1+\delta\right)\left\|A-A_{\ell}\right\|_{2}^{2}\left\|\mathbb{E}[M_{(i)}] \right\|_{2}\notag. 
$$
And by Jensen's inequality, we further have
\begin{align}
&\sigma^{2}\le\sum_{i=1}^{s}\left(1+\delta\right)\left\|A-A_{\ell}\right\|_{2}^{2}\mathbb{E}\left\|M_{(i)}\right\|_{2}\notag\\
=&\sum_{i=1}^{s}\left(1+\delta\right)^2\left\|A-A_{\ell}\right\|_{2}^{4}
= s \left(1+\delta\right)^2\left\|A-A_{\ell}\right\|_{2}^{4}. \notag
\end{align}
It then follows the Lemma \ref{Matrix-Bernstein-inequality} that 
\begin{align}
\mathbb{P}\left(\left\|\sum_{i=1}^{s} \left(M_{(i)}-\mathbb{E}[M_{(i)}]\right)\right\|_{2} \geq t\right) \le 2d \exp \left(\frac{-t^{2} / 2}{\sigma^{2}+R t / 3}\right) \label{apply-bernstein}
\end{align}
for all $t \geq 0$. 
We denote the right-hand side of (\ref{apply-bernstein}) by $\eta$, then
\begin{align}
t &=\log \left(\frac{2 d}{\eta}\right)\left(\frac{R}{3}+\sqrt{\left(\frac{R}{3}\right)^{2}+\frac{2 \sigma^{2}}{\log (2 d / \eta)}}\right) \notag\\
&\leq \log \left(\frac{2 d}{\eta}\right) \frac{2 R}{3}+\sqrt{2 \sigma^{2} \log \left(\frac{2 d}{\eta}\right)}. \label{t-bound}
\end{align}
Plugging $R$ and $\sigma^{2}$ into (\ref{t-bound}) can get that
\begin{align}
t &\leq \left(\log \left(\frac{2 d}{\eta}\right) \frac{4 \left(1+\delta\right)}{3}+\sqrt{2 s \left(1+\delta\right)^2 \log \left(\frac{2 d}{\eta}\right)}\right)  \left\|A-A_{\ell}\right\|_{2}^{2}. \notag 
\end{align}
That is to say, with probability at least $1-\eta$, we have 
\begin{align}
&\left\|\sum_{i=1}^{s} \left(M_{(i)}-\mathbb{E}[M_{(i)}]\right)\right\|_{2}\notag\\
\le&\left(\log \left(\frac{2 d}{\eta}\right) \frac{4 \left(1+\delta\right)}{3}+\sqrt{2 s \left(1+\delta\right)^2 \log \left(\frac{2 d}{\eta}\right)}\right)  \left\|A-A_{\ell}\right\|_{2}^{2}.\notag
\end{align}
By triangle and Jensen's inequalities, we have
\begin{align}
\left\|\sum_{i=1}^{s} M_{(i)}\right\|_{2}
\le\sum_{i=1}^{s} \mathbb{E}\left\|M_{(i)}\right\|_{2}+
\left\|\sum_{i=1}^{s} \left(M_{(i)}-\mathbb{E}[M_{(i)}]\right)\right\|_{2}.\notag
\end{align}
Therefore, with probability at least $1-\eta-2s\exp(-\varepsilon^{2}/2)$, we have
\begin{align}
&\left\|A^{ T} A-P^{T} P\right\|_{2}=\left\|\sum_{i=1}^{s} M_{(i)}\right\|_{2}\notag\\
\le&\sum_{i=1}^{s} \mathbb{E}\left\|M_{(i)}\right\|_{2}+
\left\|\sum_{i=1}^{s} \left(M_{(i)}-\mathbb{E}[M_{(i)}]\right)\right\|_{2}\notag\\
\le&
\left(s\left(1+\delta\right)+\log \left(\frac{2 d}{\eta}\right) \frac{4 \left(1+\delta\right)}{3}+\sqrt{2 s \left(1+\delta\right)^2 \log \left(\frac{2 d}{\eta}\right)}\right)\notag\\
&\times\left\|A-A_{\ell}\right\|_{2}^{2}.\notag
\end{align}
Now we bound $\left\|P^{ T} P-B^{T} B\right\|_{2}$. Based on the property of FD stated in \cite{ghashami2016frequent}, that is, $\left\|P^{ T} P-B^{T} B\right\|_{2}\le \frac{\left\|P-P_{k}\right\|_{F}^{2}}{\ell-k}$, where $P_{k}$ is the rank-$k$ approximation of $P$. Let $M$ be the projection matrix onto the subspace spanned by the top-$k$ right singular vectors of $A$. Noting that $M$ is of rank $k$, then $\frac{\left\|P-P_{k}\right\|_{F}^{2}}{\ell-k} \le \frac{\left\|P-PM\right\|_{F}^{2}}{\ell-k}$. Thus, 
%\begin{small}
\begin{align}	
&\left\|P^{ T} P-B^{T} B\right\|_{2}\le \frac{\left\|P-PM\right\|_{F}^{2}}{\ell-k}=\sum_{i=1}^{s} \frac{\left\|P_{(i)}-P_{(i)}M\right\|_{F}^{2}}{\ell-k} \notag\\
=&\sum_{i=1}^{s} \frac{\left\|Z_{(i)}^{T}A_{(i)}-Z_{(i)}^{T}A_{(i)}M\right\|_{F}^{2}}{\ell-k}
\le\sum_{i=1}^{s} \frac{\left\|Z_{(i)}^{T}\right\|_{2}^{2}   \left\|A_{(i)}-A_{(i)}M\right\|_{F}^{2}}{\ell-k} \notag\\
=& \frac{   \left\|A-AM\right\|_{F}^{2}}{\ell-k} \label{ortho}\\
=& \frac{   \left\|A-A_{k}\right\|_{F}^{2}}{\ell-k} ,\label{def-M}
\end{align}
where (\ref{ortho}) holds because of $Z_{(i)}$ is orthonormal, and (\ref{def-M}) holds by the definition of $M$.
%\end{small}

From the aforementional analysis, we finally have
%combined with Theorem $\ref{55}$, we can find that:
%\small
%\begin{small}
$$
\begin{aligned}	
&\left\|A^{ T} A-B^{T} B\right\|_{2}\\
\leq&\left\|A^{ T} A-P^{T} P\right\|_{2}+\left\|P^{ T} P-B^{T} B\right\|_{2}\\
\le&\left(s\left(1+\delta\right)+\log \left(\frac{2 d}{\eta}\right) \frac{4 \left(1+\delta\right)}{3}+\sqrt{2 s \left(1+\delta\right)^2 \log \left(\frac{2 d}{\eta}\right)}\right)\notag\\
&\times\left\|A-A_{\ell}\right\|_{2}^{2}+\frac{   \left\|A-A_{k}\right\|_{F}^{2}}{\ell-k}.
\end{aligned}
$$
%\end{small}
This arrives to the conclusion of Theorem \ref{thm2}.   \qed
%\end{proof of theorem}

Before proving Theorem \ref{thm5}, we need to list the following auxiliary property for CountSketch matrix.
\begin{lemma}[Lemma 1 in \cite{teng2018fast}] \label{csproperty}
	Given a matrix $U \in \mathbb{R}^{n \times k}$ with orthonormal columns
	$(k \leq n) .$ For any $\varepsilon, p \in(0,1),$ let the CountSketch matrix $S \in \mathbb{R}^{m \times n}$ be defined as
	with
	$$
	m \geq \frac{k^{2}+k}{\varepsilon^{2} p}.
	$$
	Let $P$ be an $n \times n$ random permutation matrix (i.e., for each $e_{j}$ with $j \in[n],\ \mathbb{P}\left[P e_{j}=e_{i}\right]=1 / n$ for $\left.i \in[n]\right),$ then with probability at least $1-p$,
	$$
	\left\|U^{T}(S P)^{T} S P U-I\right\|_{2} \leq \varepsilon.
	$$
\end{lemma}
This lemma reveals that CountSketch is a $1 \pm \varepsilon$ $\ell_2$-subspace embedding for $U$. And it is noted that the permutation matrix $P$ can be unit matrix in special cases.

\noindent {\it Proof of Theorem 2.} The proof of Theorem \ref{thm5} is similar to the procedure of the proof of Theorem \ref{thm2} with a little bit of differences. This mainly because of the various constructions of $X$, resulting in the different error bounds of $\left\|A^{ T} A-P^{T} P\right\|_{2}$. Let the CountSketch matrix be $X =  D \Phi^{T}\in \mathbb{R}^{d \times m}$. Noting that $\sigma_{\ell}\left([V_{(i)}]_{\ell}^{T} X\right)=\sigma_{\ell}\left([V_{(i)}]_{\ell}^{T} D \Phi^{T}\right)=\sigma_{\ell}\left(\Phi D [V_{(i)}]_{\ell}\right)$, and according to Lemma \ref{csproperty}, we can get that for any $\varepsilon, p \in(0,1)$, if $m \geq \frac{\ell^{2}+\ell}{\varepsilon^{2}p}$, the inequation $\left\|[V_{(i)}]_{\ell}^{T}X X^{T} [V_{(i)}]_{\ell}-I\right\|_{2} \leq \varepsilon$ holds with probability at least $1-p$, that is,
%\small
\begin{align} \label{singular-cs}
%\begin{small}
\sigma_{\ell}\left([V_{(i)}]_{\ell}^{T} X\right) \ge \sqrt{1-\varepsilon}.
%\end{small}	
\end{align}
By the sub-multiplicativity property of Frobenius norm, we have
\begin{align} \label{F-cs}
\left\|[V_{(i)}]_{\ell, \perp}^{T} X\right\|_{ F}\le  \left\|[V_{(i)}]_{\ell, \perp}^{T}\right\|_{ F} \left\| D\right\|_{2} \left\|\Phi^{T}\right\|_{F}=\sqrt{d(d-\ell)}. 
\end{align}
It then follows by inequalities (\ref{singular-cs}) and (\ref{F-cs}) that,
%\begin{small}
\begin{align}
\left\|[V_{(i)}]_{\ell, \perp}^{T} X\left([V_{(i)}]_{\ell}^{T} X\right)^{\dagger}\right\|_{F}&\leq\left\|[V_{(i)}]_{\ell, \perp}^{T} X\right\|_{F}\left\|\left([V_{(i)}]_{\ell}^{T} X\right)^{\dagger}\right\|_{2}\notag\\
&\leq \sqrt{\frac{d(d-\ell)}{1-\varepsilon}}.\label{csfnorm}
\end{align}
%\end{small}
Next, we first derive the upper bound of $\left\|M_{(i)}\right\|_{2}$ by combining inequalities (\ref{lemma1proof}) and (\ref{csfnorm}), and then calculate $R$ and $\sigma^2$ using the same  procedure as Theorem \ref{thm2}. Thus with probability at least $1-sp-\eta$, we can get that
%\begin{small}	
\begin{align}
&\left\|A^{ T} A-P^{T} P\right\|_{2}\notag\\
\le&
\left(s\left(1+\delta\right)+\log \left(\frac{2 d}{\eta}\right) \frac{4 \left(1+\delta\right)}{3}+\sqrt{2 s \left(1+\delta\right)^2 \log \left(\frac{2 d}{\eta}\right)}\right)\notag\\
&\times\left\|A-A_{\ell}\right\|_{2}^{2},\notag
\end{align}
%\end{small}
where $1+\delta=\left(1+\frac{4 }{2^{(2 q+1) \min \{\sqrt{\gamma}, 1\}}}\sqrt{\frac{d(d-\ell)}{1-\varepsilon}}\right)^{2}$.
Based on the above analysis, we can easily derive that
%\begin{small}
$$
\begin{aligned}	
&\left\|A^{ T} A-B^{T} B\right\|_{2}\\
\leq&\left\|A^{ T} A-P^{T} P\right\|_{2}+\left\|P^{ T} P-B^{T} B\right\|_{2}\\
%&\leq\left\|A^{ T} A-P^{T} P\right\|_{F}+\left\|P^{ T} P-B^{T} B\right\|_{2}\\
\le&\left(s\left(1+\delta\right)+\log \left(\frac{2 d}{\eta}\right) \frac{4 \left(1+\delta\right)}{3}+\sqrt{2 s \left(1+\delta\right)^2 \log \left(\frac{2 d}{\eta}\right)}\right)\notag\\
&\times\left\|A-A_{\ell}\right\|_{2}^{2}+\frac{   \left\|A-A_{k}\right\|_{F}^{2}}{\ell-k}.\notag
\end{aligned}
$$
%\end{small}
With this, the proof of Theorem \ref{thm5} is complete.     \qed
%\end{proof of theorem}
}

\bibliographystyle{IEEEtran}
\bibliography{refs}

% or


%\appendices
%\section{Proof of Theorem 1}


%\section*{Acknowledgment}

\ifCLASSOPTIONcaptionsoff
\newpage
\fi

\end{document}